\documentclass[superscriptaddress,twocolumn]{revtex4}

\RequirePackage[normalem]{ulem} 
\RequirePackage{color}\definecolor{RED}{rgb}{1,0,0}\definecolor{BLUE}{rgb}{0,0,1} 

\usepackage{url}
\usepackage{hyperref}
\usepackage{graphicx,epsfig,verbatim,enumerate}
\usepackage{amssymb,amsmath}
\usepackage{csquotes}
\usepackage{ifthen}
\newboolean{twocolswitch}

\usepackage{morefloats}

\usepackage{xcolor}
\usepackage{color}
\usepackage[normalem]{ulem}

\definecolor{lightgrey}{rgb}{0.5,0.5,0.5}

\usepackage{rotating}
\usepackage{pdflscape}
\usepackage{longtable}
\usepackage{adjustbox}
\usepackage{array}

\newcommand{\sindex}[1]{}
\newcommand{\nindex}[1]{}

\newcommand{\www}[1]{\url{#1}}

\newcommand{\PreserveBackslash}[1]{\let\temp=\\#1\let\\=\temp}
\newcommand{\PBS}[1]{\let\temp=\\#1\let\\=\temp}

\newcommand{\nbooks}{1,327}

\newcommand{\plainlatexonly}[1]{}

\begin{document}

\title{
The emotional arcs of stories are dominated by six basic shapes
}

\author{Andrew J. Reagan}
\affiliation{Department of Mathematics \& Statistics, Vermont Complex Systems Center, Computational Story Lab, \& the Vermont Advanced Computing Core, The University of Vermont, Burlington, VT 05401}
\author{Lewis Mitchell}
\affiliation{School of Mathematical Sciences, The University of Adelaide, SA 5005 Australia}
\author{Dilan Kiley}
\affiliation{Department of Mathematics \& Statistics, Vermont Complex Systems Center, Computational Story Lab, \& the Vermont Advanced Computing Core, The University of Vermont, Burlington, VT 05401}
\author{Christopher M. Danforth}
\affiliation{Department of Mathematics \& Statistics, Vermont Complex Systems Center, Computational Story Lab, \& the Vermont Advanced Computing Core, The University of Vermont, Burlington, VT 05401}
\author{Peter Sheridan Dodds}
\affiliation{Department of Mathematics \& Statistics, Vermont Complex Systems Center, Computational Story Lab, \& the Vermont Advanced Computing Core, The University of Vermont, Burlington, VT 05401}

\date{\today}

\begin{abstract}
Advances in computing power, natural language processing, and digitization of text now make it possible to study a culture's evolution through its texts using a ``big data'' lens.
Our ability to communicate relies in part upon a shared emotional experience, with stories often following distinct emotional trajectories and forming patterns that are meaningful to us.
Here, by classifying the emotional arcs for a filtered subset of \nbooks~stories from Project Gutenberg's fiction collection, we find a set of six core emotional arcs which form the essential building blocks of complex emotional trajectories.
We strengthen our findings by separately applying Matrix decomposition, supervised learning, and unsupervised learning.
For each of these six core emotional arcs, we examine the closest characteristic stories in publication today and find that particular emotional arcs enjoy greater success, as measured by downloads.
\end{abstract}

\maketitle

\section{Introduction}

The power of stories to transfer information and define our own existence has been shown time and again \cite{pratchett2003a,campbell2008a,gottschall2013a,cave2013a,dodds2013homo}.
We are fundamentally driven to find and tell stories, likened to \textit{Pan Narrans} or \textit{Homo Narrativus}.
Stories are encoded in art, language, and even in the mathematics of physics:
We use equations to represent both simple and complicated functions that describe our observations of the real world.
In science, we formalize the ideas that best fit our experience with principles such as Occam's Razor:
The simplest story is the one we should trust.
We tend to prefer stories that fit into the molds which are familiar,
and reject narratives that do not align with our experience \cite{nickerson1998a}.

We seek to better understand stories that are captured and shared in written form,
a medium that since inception has radically changed how information flows \cite{gleick2011a}.
Without evolved cues from tone, facial expression, or body language,
written stories are forced to capture the entire transfer of experience on a page.
An often integral part of a written story is the emotional experience that is evoked in the reader.
Here, we use a simple, robust sentiment analysis tool to extract the reader-perceived emotional content of written stories as they unfold on the page.

We objectively test aspects of the theories of folkloristics \cite{propp1968morphology,macdonal1982storytellers},
specifically the commonality of core stories within societal boundaries \cite{cave2013a,silva2016a}.
A major component of folkloristics is the study of society and culture through literary analysis.
This is sometimes referred to as \textit{narratology},
which at its core is ``a series of events, real or fictional, presented to the reader or the listener'' \cite{min2016a}.
In our present treatment,
we consider the plot as the ``backbone'' of events that occur in a chronological sequence (more detail on previous theories of plot are in Appendix~\ref{sec:plots}).
While the plot captures the mechanics of a narrative and the structure encodes their delivery,
in the present work we examine the emotional arc that is invoked through the words used.
The emotional arc of a story does not give us direct information about the plot or the intended meaning of the story,
but rather exists as part of the whole narrative
(e.g., an emotional arc showing a fall in sentiment throughout a story may arise from very different plot and structure combinations).
This distinction between the emotional arc and the plot of a story is one point of misunderstanding in other work that has drawn criticism from the digital humanities community \cite{jockers2016novel}.
Through the identification of motifs \cite{dundes1997motif},
narrative theories \cite{dolby2008literary} allow us to analyze, interpret, describe, and compare stories across cultures and regions of the world \cite{uther2011a}.
We show that automated extraction of emotional arcs is not only possibly,
but can test previous theories and provide new insights with the potential to quantify unobserved trends as the field transitions from data-scarce to data-rich \cite{kirschenbaum2007remaking,moretti2013}.

The rejected master's thesis of Kurt Vonnegut---which he personally considered his greatest contribution---defines the \textit{emotional arc} of a story on the ``Beginning--End'' and ``Ill Fortune--Great Fortune'' axes \cite{vonnegut1981}.
Vonnegut finds a remarkable similarity between Cinderella and the origin story of Christianity in the Old Testament (see Fig.~\ref{fig:vonnegut} in Appendix~\ref{sec:extras}),
leading us to search for all such groupings.
In a recorded lecture available on YouTube \cite{vonnegut1995},
Vonnegut asserted:
\begin{displayquote}
  ``There is no reason why the simple shapes of stories can't be fed into computers, they are beautiful shapes.''
\end{displayquote}

For our analysis,
we apply three independent tools: Matrix decomposition by Singular Value Decomposition (SVD),
supervised learning by agglomerative (hierarchical) clustering with Ward's method,
and unsupervised learning by a Self Organizing Map (SOM, a type of neural network).
Each tool encompasses different strengths: the SVD finds the underlying basis of all of the emotional arcs,
the clustering classifies the emotional arcs into distinct groups,
and the SOM generates arcs from noise which are similar to those in our corpus using a stochastic process.
It is only by considering the results of each tool in support of each other that we are able to confirm our findings.

We proceed as follows.
We first introduce our methods in Section~\ref{sec:methods},
we then discuss the combined results of each method in Section~\ref{sec:results},
and we present our conclusions in Section~\ref{sec:conclusion}.
A graphical outline of the methodology and results can be found as Fig.~\ref{fig:infographic} in Appendix~\ref{sec:extras}.

\section{Methods}
\label{sec:methods}

\subsection{Emotional arc construction}
\begin{figure}[tbp!]
  \centering
  \includegraphics[width=0.48\textwidth]{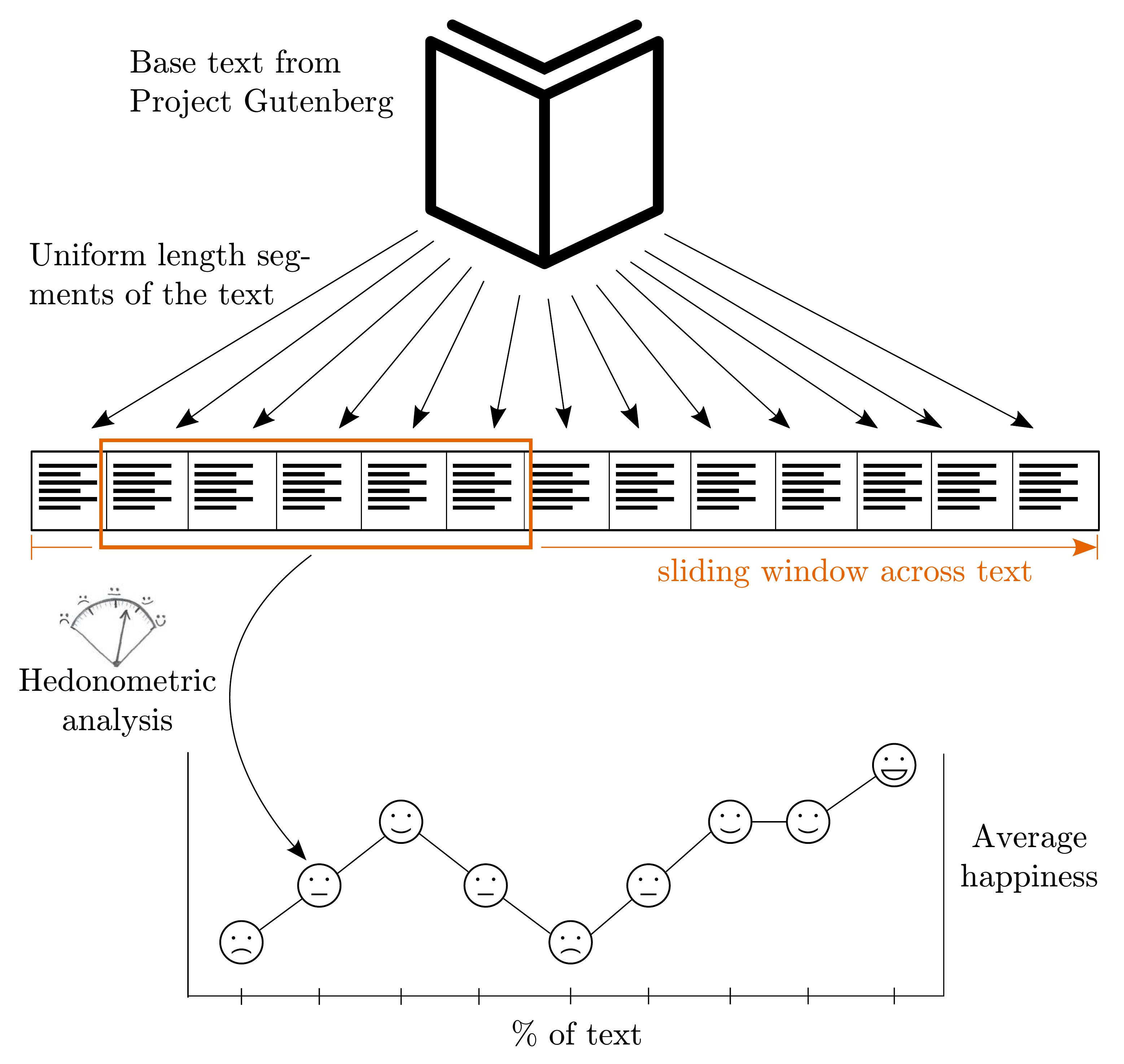}
  \caption[]{
    Schematic of how we compute emotional arcs.
    The indicated uniform length segments (gap between samples) taken from the text form the sample with fixed window size set at $N_w = 10,000$ words.
    The segment length is thus $N_s = (N -(N_w+1))/n$ for $N$ the  length of the book in words,
    and $n$ the number of points in the time series.
    Sliding this fixed size window through the book,
    we generate $n$ sentiment scores with the Hedonometer,
    which comprise the emotional arc \cite{dodds2011a}.
  }
  \label{fig:timeseries-schematic}
\end{figure}

\begin{figure*}[tbp!]
  \centering
  \includegraphics[width=0.97\textwidth]{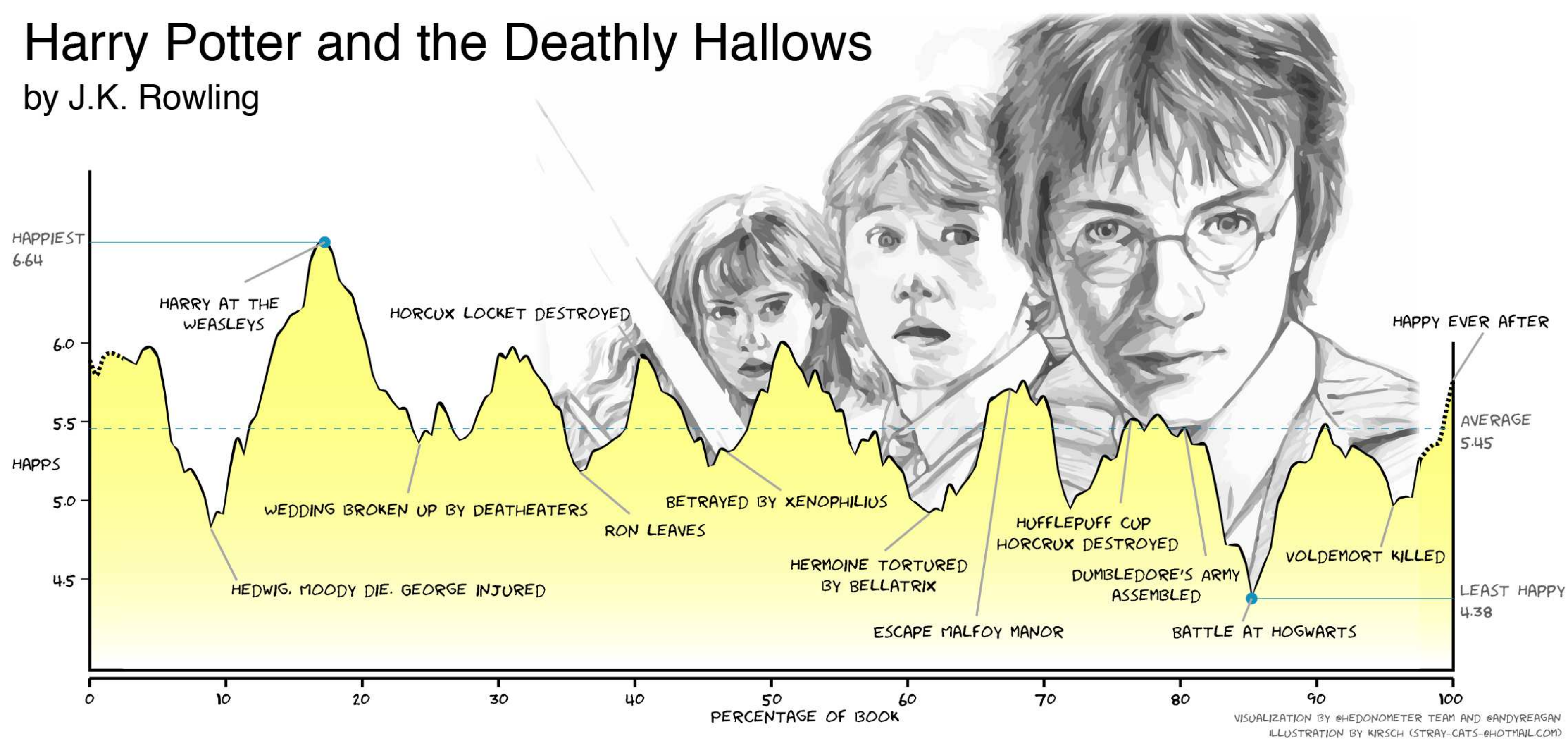}
  \caption[]{
    Annotated emotional arc of \textit{Harry Potter and the Deathly Hallows},
    by J.K. Rowling,
    inspired by the illustration made by Medaris for The Why Files \cite{tenenbaum2015a}.
    The entire seven book series can be classified as a ``Kill the monster'' plot \cite{booker2006a},
    while the many sub plots and connections between them complicate the emotional arc of each individual book: this plot could not be readily inferred from the emotional arc alone.
    The emotional arc shown here,
    captures the major highs and lows of the story,
    and should be familiar to any reader well acquainted with Harry Potter. 
    Our method does not pick up emotional moments discussed briefly,
    perhaps in one paragraph or sentence (e.g., the first kiss of Harry and Ginny).    
    We provide interactive visualizations of all Project Gutenberg books at \url{http://hedonometer.org/books/v3/1/} and a selection of classic and popular books at \url{http://hedonometer.org/books/v1/}.
  }
  \label{fig:harry-potter}
\end{figure*}

To generate emotional arcs,
we analyze the sentiment of 10,000 word windows,
which we slide through the text (see Fig.~\ref{fig:timeseries-schematic}).
We rate the emotional content of each window using our Hedonometer with the labMT dataset,
chosen for lexical coverage and its ability to generate meaningful word shift graphs,
specifically using 10,000 words as a minimum necessary to generate meaningful sentiment scores \cite{reagan2016a,ribeiro2016sentibench}.
We emphasize that dictionary-based methods for sentiment analysis usually perform worse than random on individual sentences \cite{reagan2016a,ribeiro2016sentibench},
and although this issue can be resolved by using a rolling average of sentences scores,
it begets a basic misunderstanding of similar efforts \cite{jockers2016novel}.
In Fig.~\ref{fig:harry-potter},
we show the emotional arc of \textit{Harry Potter and the Deathly Hallows},
the final book in the popular Harry Potter series by J.K. Rowling.
While the plot of the book is nested and complicated,
the emotional arc associated with each sub-narrative is clearly visible.
We analyze the emotional arcs corresponding to complete books,
and to limit the conflation of multiple core emotional arcs, we restrict our analysis to shorter books by selecting a maximum number of words when building our filter.
Further details of the emotional arc construction can be found in Appendix~\ref{sec:construction}.

\subsection{Project Gutenberg Corpus}

For a suitable corpus we draw on the open access Project Gutenberg data set \cite{gutenberg}.
We apply rough filters to the collection (roughly 50,000 books) in an attempt to obtain a set of books that represent English works of fiction.
We start by selecting for only English books,
with total words between 20,000 and 100,000,
with more than 40 downloads from the Project Gutenberg website,
and with Library of Congress Class corresponding to English fiction\footnotemark.
\footnotetext{The specific classes have labels PN, PR, PS, and PZ.}
To ensure that the 40-download limit is not influencing the results here,
we further test each method for 10, 20, 40, and 80 download thresholds,
in each case confirming the 40 download findings to be qualitatively unchanged.
Next, we remove books with any word in the title from a list of keywords (e.g., ``poems'' and ``collection'',
full list in Appendix~\ref{sec:construction}).
From within this set of books,
we remove the front and back matter of each book using regular expression pattern matches that match on 98.9\% of the books included.
Two slices of the data for download count and the total word count are shown in Appendix~\ref{sec:construction} Fig.~\ref{fig:length-distribution}.\ 
We provide a list of the book ID's which are included for download in the Online Appendices at \url{http://compstorylab.org/share/papers/reagan2016b/},
the books are listed in Table~\ref{tbl:allbooks} in Appendix~\ref{sec:lists},
and we attempt to provide the Project Gutenberg ID when we mention a book by title herein.
Given the Project Gutenberg ID $n$,
the raw ebook is available online from Project Gutenberg at \url{http://www.gutenberg.org/ebooks/n},
e.g., \textit{Alice's Adventures in Wonderland} by Lewis Carroll,
has ID 11 and is available at \url{http://www.gutenberg.org/ebooks/11}.
We also provide an online, interactive version of the emotional arc for each book indexed by the ID,
e.g., \textit{Alice's Adventures in Wonderland} is available at \url{http://hedonometer.org/books/v3/11/}.

\subsection{Principal Component Analysis (SVD)}

We use the standard linear algebra technique Singular Value Decomposition (SVD) to find a decomposition of stories onto an orthogonal basis of emotional arcs.
Starting with the sentiment time series for each book $b_i$ as row $i$ in the matrix $A$,
we apply the SVD to find
\begin{align}
  A  &= U \Sigma V^{T}
  = W V^{T}
  ,
  \label{eq:SVD}
\end{align}
where $U$ contains the projection of each sentiment time series onto each of the right singular vectors (rows of $V^{T}$,
eigenvectors of $A^TA$),
which have singular values given along the diagonal of $\Sigma$,
with $W = U \Sigma$.
Different intuitive interpretations of the matrices $U, \Sigma,$ and $V^T$ are useful in the various domains in which the SVD is applied; here,
we focus on right singular vectors as an orthonormal basis for the sentiment time series in the rows of $A$,
which we will refer to as the \textit{modes}.
We combine $\Sigma$ and $U$ into the single coefficient matrix $W$ for clarity and convenience,
such that $W$ now represents the mode coefficients.

\subsection{Hierarchical Clustering}
\label{sec:clustering}

We use Ward's method to generate a hierarchical clustering of stories,
which proceeds by minimizing variance between clusters of books \cite{ward1963hierarchical}.
We use the mean-centered books and the distance function
\begin{equation}
  D(b_i, b_j )
  =
  l^{-1}
  \sum ^l _{t=1} |b_i(t) - b_j (t)|
  .
  \label{eq:distance}
\end{equation}
for $t$ indexing the window in books $b_i, b_j$ to generate the distance matrix.

\subsection{Self Organizing Map (SOM)}
\label{sec:SOM}

We implement a Self Organized Map (SOM),
an unsupervised machine learning method (a type of neural network)
to cluster emotional arcs \cite{kohonen1990self}.
The SOM works by finding the most similar emotional arc in a random collection of arcs.
We use an 8x8 SOM (for 64 nodes,
roughly 5\% of the number of books),
connected on a square grid,
training according to the original procedure (with winner take all, and scaling functions across both distance and magnitude).
We take the neighborhood influence function at iteration $i$ as
\begin{equation}
  \text{Nbd}_{k}(i) = \left [ j \in \mathcal{N} ~|~ D(k,j) < \sqrt{N} \, \cdot (i+1)^\alpha \right ]
\end{equation}
for a node $k$ in the set of nodes $\mathcal{N}$,
with distance function $D$ given above and total number of nodes $N$.
For results shown here we take $\alpha = -0.15$.
We implement the learning adaptation function at training iteration $i$ as
$f(i) = (i+1)^{\beta}$,
again with $\beta = -0.15$, a standard value for the training hyper-parameters.

\section{Results}
\label{sec:results}

We obtain a collection of \nbooks~books that are mostly,
but not all,
fictional stories by using metadata from Project Gutenberg to construct a rough filter.
We find broad support for the following six emotional arcs:
\begin{itemize}
\item ``Rags to riches'' (rise).
\item ``Tragedy'', or ``Riches to rags'' (fall).
\item ``Man in a hole'' (fall-rise).
\item ``Icarus'' (rise-fall).
\item ``Cinderella'' (rise-fall-rise).
\item ``Oedipus'' (fall-rise-fall).
\end{itemize}

\noindent Importantly,
we obtain these same six emotional arcs from all possible arcs by observing them as the result of three methods:
As modes from a matrix decomposition by SVD,
as clusters in a hierarchical clustering using Ward's algorithm,
and as clusters using unsupervised machine learning.
We examine each of the results in this section.

\subsection{Principal Component Analysis (SVD)}

\begin{figure*}[tbp!]
  \centering
  \includegraphics[width=0.9\textwidth]{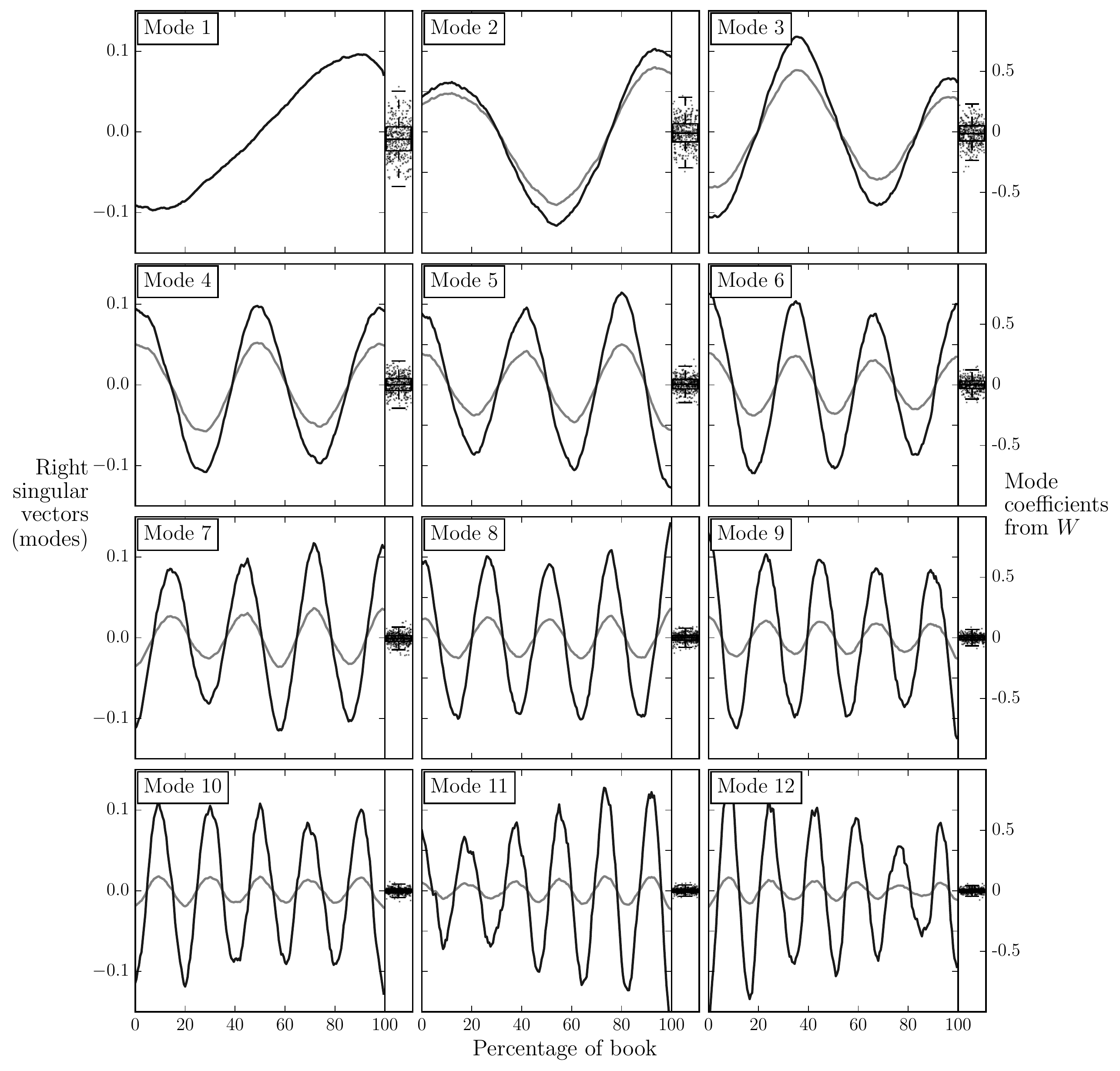}
  \caption[]{
    Top 12 modes from the Singular Value Decomposition of \nbooks~Project Gutenberg books.
    We show in a lighter color modes weighted by their corresponding singular value,
    where we have scaled the matrix $\Sigma$ such that the first entry is 1 for comparison (for reference, the largest singular value is 34.5).
    The mode coefficients normalized for each book are shown in the right panel accompanying each mode, in the range -1 to 1, with the ``Tukey'' box plot.
  }
  \label{fig:SVD-12-comp}
\end{figure*}

In Fig.~\ref{fig:SVD-12-comp} we show the leading 12 modes in both the weighted (dark) and un-weighted (lighter) representation.
In total, the first 12 modes explain 80\% and 94\% of the variance from the mean centered and raw time series, respectively.
The modes are from mean-centered emotional arcs,
such that the first SVD mode need not extract the average from the labMT scores nor the positivity bias present in language \cite{dodds2015a}.
The coefficients for each mode within a single emotional arc are both positive and negative,
so we need to consider both the modes and their negation.
We can immediately recognize the familiar shapes of core emotional arcs in the first four modes,
and compositions of these emotional arcs in modes 5 and 6.
We observe ``Rags to riches'' (mode 1, positive),
``Tragedy'' or ``Riches to rags'' (mode 1, negative),
Vonnegut's ``Man in a hole'' (mode 2, positive),
``Icarus'' (mode 2, negative),
``Cinderella'' (mode 3, positive),
``Oedipus'' (mode 3, negative).
We choose to include modes 7--12 only for completeness,
as these high frequency modes have little contribution to variance and do not align with core emotional arc archetypes from other methods (more below).

We emphasize that by definition of the SVD,
the mode coefficients in $W$ can be either positive and negative,
such that the modes themselves explain variance with both the positive and negative version.
In the right panels of each mode in Fig.~\ref{fig:SVD-12-comp} we project the \nbooks~stories onto each of first six modes and show the resulting coefficients.
While none are far from 0 (as would be expected),
mode 1 has a mean slightly above 0 and both modes 3 and 4 have means slightly below 0.
To sort the books by their coefficient for each mode,
we normalize the coefficients within each book in the rows of $W$ to sum to 1,
accounting for books with higher total energy,
and these are the coefficients shown in the right panels of each mode in Fig.~\ref{fig:SVD-12-comp}.
In Appendix~\ref{sec:SVD-supp},
we provide supporting, intuitive details of the SVD method,
as well as example emotional arc reconstruction using the modes (see Figs.~\ref{fig:SVD-USV}--\ref{fig:SVD-reconstruction}).
As expected, less than 10 modes are enough to reconstruct the emotional arc to a degree of accuracy visible to the eye.

\begin{figure*}[tbp!]
  \centering
  \includegraphics[width=.98\textwidth]{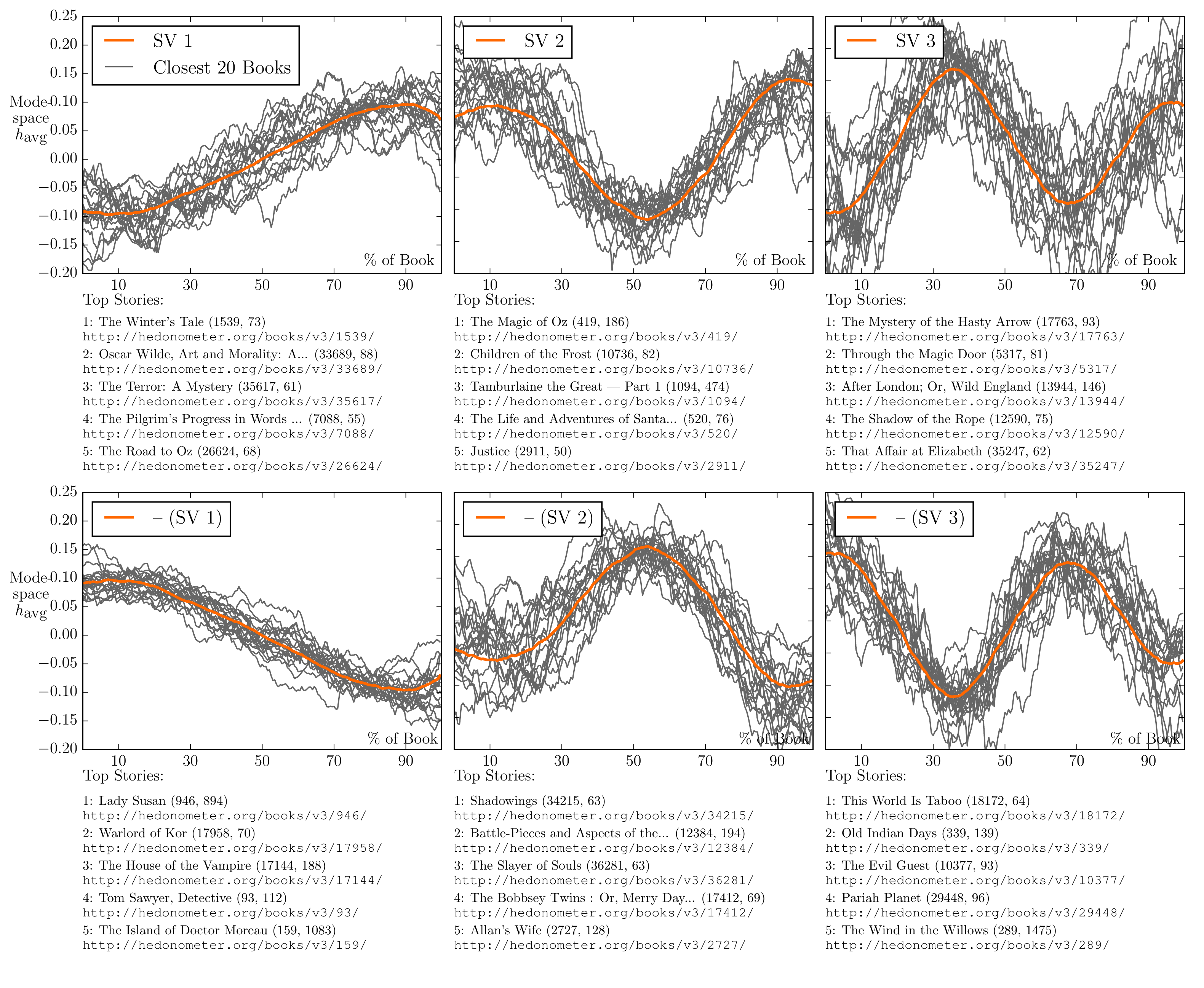}
  \caption[]{
    First 3 SVD modes and their negation with the closest stories to each.
    To locate the emotional arcs on the same scale as the modes,
    we show the modes directly from the rows of $V^T$ and weight the emotional arcs by the inverse of their coefficient in $W$ for the particular mode.
    The closest stories shown for each mode are those stories with emotional arcs which have the greatest coefficient in $W$.
    In parentheses for each story is the Project Gutenberg ID and the number of downloads from the Project Gutenberg website, respectively.
    Links below each story point to an interactive visualization on \url{http://hedonometer.org} which enables detailed exploration of the emotional arc for the story.
  }
  \label{fig:SVD-1-3-labelled}
\end{figure*}

We show labeled examples of the emotional arcs closest to the top 6 modes in Figs.~\ref{fig:SVD-1-3-labelled} and~\ref{fig:SVD-4-6-labelled}.
We present both the positive and negative modes,
and the stories closest to each by sorting on the coefficient for that mode.
For the positive stories, we sort in ascending order, and vice versa.
Mode 1, which encompasses both the ``Rags to riches'' and ``Tragedy'' emotional arcs, captures 30\% of the variance of the entire space.
We examine the closest stories to both sides of modes 1--3,
and direct the reader to Fig.~\ref{fig:SVD-4-6-labelled} for more details on the higher order modes.
The two stories that have the most support from the ``Rags to riches'' mode are \textit{The Winter's Tale} (1539) and \textit{Oscar Wilde, Art and Morality: A Defence of ``The Picture of Dorian Gray''} (33689).
Among the most categorical tragedies we find \textit{Lady Susan} (946) and \textit{Warlord of Kor} (17958).
Number 8 in the sorted list of tragedies is perhaps the most famous tragedy: \textit{Romeo and Juliet} by William Shakespeare.
Mode 2 is the ``Man in a hole'' emotional arc,
and we find the stories which most closely follow this path to be \textit{The Magic of Oz} (419) and \textit{Children of the Frost} (10736).
The negation of mode 2 most closely resembles the emotional arc of the ``Icarus'' narrative.
For this emotional arc,
the most characteristic stories are \textit{Shadowings} (34215) and \textit{Battle-Pieces and Aspects of the War} (12384).
Mode 3 is the ``Cinderella'' emotional arc,
and includes \textit{Mystery of the Hasty Arrow} (17763) and \textit{Through the Magic Dorr} (5317).
The negation of Mode 3, which we refer to as ``Oedipus'', is found most characteristically in \textit{This World is Taboo} (18172), \textit{Old Indian Days} (339), and \textit{The Evil Guest} (10377).
We also note that the spread of the stories from their core mode increases strongly for the higher modes.

\subsection{Hierarchical Clustering}
\label{sec:clustering}

\begin{figure*}[tbp!]
  \centering
  \includegraphics[width=0.98\textwidth]{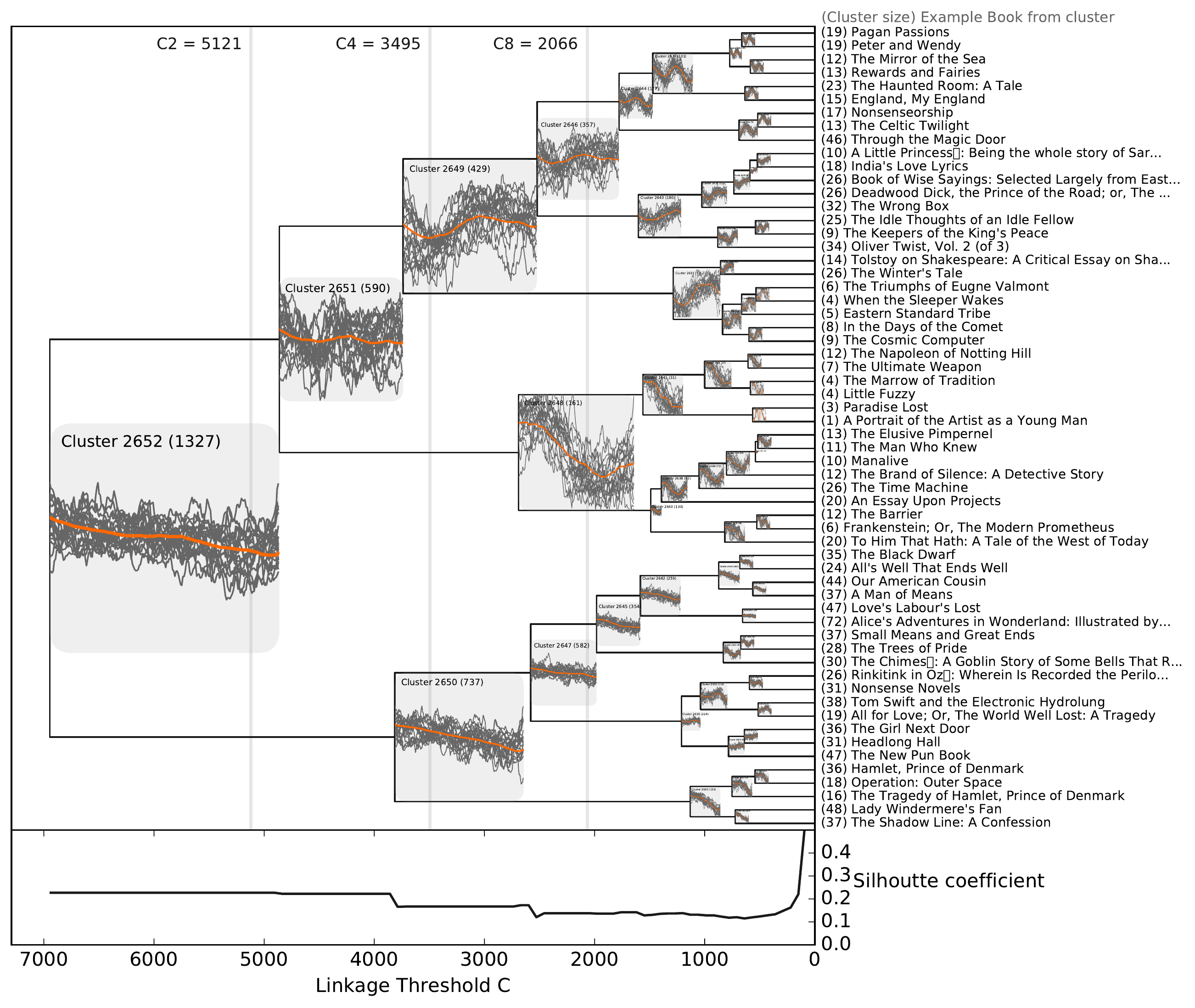}
  \caption[]{
    Dendrogram from the hierarchical clustering procedure using Ward's minimum variance method.
    For each cluster,
    a selection of the 20 most central books to a fully-connected network of books are shown along with the average of the emotional arc for all books in the cluster,
    along with the cluster ID and number of books in each cluster (shown in parenthesis).
    The cluster ID is given by numbering the clusters in order of linkage starting at 0, with each individual book representing a cluster of size 1 such that the final cluster (all books) has the ID $2(N-1)$ for the $N=\nbooks$ books.
    At the bottom,
    we show the average Silhouette value for all books,
    with higher value representing a more appropriate number of clusters.
    For each of the 60 leaf nodes (right side) we show the number of books within the cluster and the most central book to that cluster's book network.
  }
  \label{fig:ward-small}
\end{figure*}

We show a dendrogram of the 60 clusters with highest linkage cost in Fig.~\ref{fig:ward-small}.
The average silhouette coefficient is shown on the bottom of Fig.~\ref{fig:ward-small},
and the distributions of silhouette values within each cluster (see Figs.~\ref{fig:clustering-2-5-clusters}--\ref{fig:clustering-6-9-clusters}) can be used to analyze the appropriate number of clusters \cite{rousseeuw1987silhouettes}.
A characteristic book from each cluster is shown on the leaf nodes by sorting the books within each cluster by the total distance to other books in the cluster (e.g.,
considering each intra-cluster collection as a fully connected weighted network,
we take the most central node),
and in parenthesis the number of books in that cluster.
In other words,
we label each cluster by considering the network centrality of the fully connected cluster with edges weighted by the distance between stories.
By cutting the dendrogram in Fig.~\ref{fig:ward-small} at various linkage costs we are able to extract clusters of the desired granularity.
For the cuts labeled C2, C4, and C8, we show these clusters in Figs.~\ref{fig:ward-cluster-2},~\ref{fig:ward-cluster-4}, and~\ref{fig:ward-cluster-8}.
We find the first four of our final six arcs appearing among the eight most different clusters (Fig.~\ref{fig:ward-cluster-8}).

The clustering method groups stories with a ``Man in a hole'' emotional arc for a range of different variances,
separate from the other arcs,
in total these clusters (Panel A, E, and I of Fig.~\ref{fig:ward-cluster-9}) account for 30\% of the Gutenberg corpus.
The remainder of the stories have emotional arcs that are clustered among the ``Tragedy'' arc (32\%),
``Rags to riches'' arc (5\%),
and the ``Oedipus'' arc (31\%).
A more detailed analysis of the results from hierarchical clustering can be found in Appendix~\ref{sec:clustering-supp},
and this result generally agrees with other attempts that use only hierarchical clustering \cite{jockers2016novel}.

\subsection{Self Organizing Map (SOM)}
\label{sec:SOM}

Finally, we apply Kohonen's Self-Organizing Map (SOM) and find core arcs from unsupervised machine learning on the emotional arcs.
On the two dimensional component plane, the prescribed network topology, we find seven spatially coherent groups, with five emotional arcs.
These spatial groups are comprised of stories with core emotional arcs of differing variance.

\begin{figure*}[tbp!]
  \centering
  \includegraphics[width=0.98\textwidth]{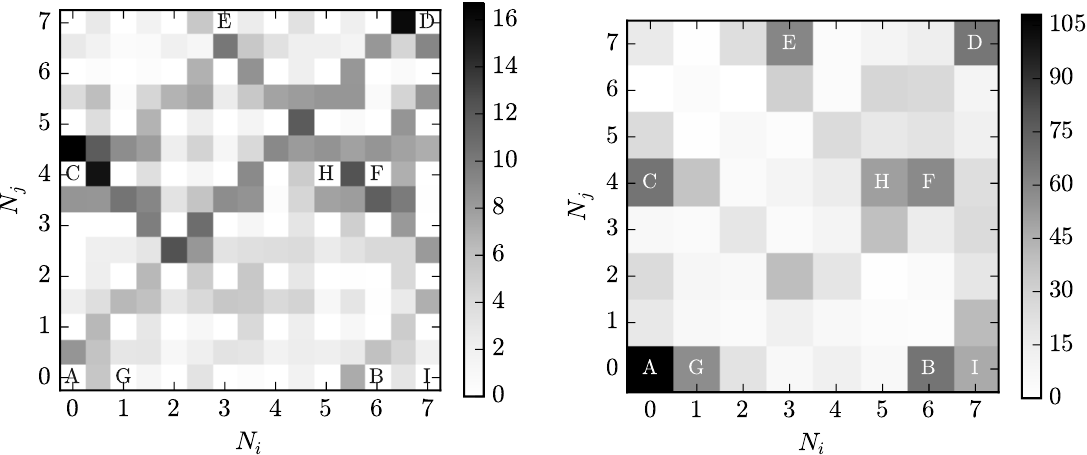}
  \caption[]{
    Results of the SOM applied to Project Gutenberg books.
    Left panel: Nodes on the 2D SOM grid are shaded by the number of stories for which they are the winner.
    Right panel: The B-Matrix shows that there are clear clusters of stories in the 2D space imposed by the SOM network.
  }
  \label{fig:SOM-matrices}
\end{figure*}

In Fig.~\ref{fig:SOM-matrices} we see both the B-Matrix to demonstrate the strength of spatial clustering and a heat-map showing where we find the winning nodes.
The A--I labels refer to the individual nodes shown in Fig.~\ref{fig:SOM-stories},
and we observe seven spatial groups in the both panels of Fig.~\ref{fig:SOM-matrices}: (1) A and G, (2) B and I, (3) C, (4) D, (5) E, and (6) H, and (7) F.
These spatial clusters reinforce the visible similarity of the winning node arcs, given that nodes H and F are close spatially but separated by the B-Matrix and contain very distinct arcs.
We show the winning node emotional arcs and the arcs of books for which they are the winners in Fig.~\ref{fig:SOM-stories}.
The legend shows the node ID, numbers the cluster by size, and in parentheses indicates the size of the cluster on that individual node.
In Panels A and G we see varying strengths of the ``Man in a hole'' emotional arc.
In Panels B and I, the second largest individual cluster consists of the ``Rags to riches'' arcs.
In Panel C, and in Panel F, we find the ``Oedipus'' emotional arc,
with a more pronounced positive start and decline in Panel C.
In Panel D we see the ``Icarus'' arc, and in Panel E and Panel H we see the ``Tragedy'' arc.
Each of these top stories are all readily identifiable,
yet again demonstrating the universality of these story types.

\subsection{Null comparison}

There are many possible emotional arcs in the space that we consider.
To demonstrate that these specific arcs are uniquely compelling as stories written by and for \textit{homo narrativus},
we consider the true emotional arcs in relation to their most suitable comparison: the book with randomly shuffled words (``word salad'') and the resulting text from a 2-gram Markov model trained on the individual book itself (``nonsense'').
We chose to compare to ``word salad'' and ``nonsense'' versions as they are more representative of a null model: written stories that are without coherent plot or structure to generate a coherent emotional arc,
which is not true of a stochastic process (e.g., a random walk model or noise).
Examples of the emotional arc and null emotional arcs for a single book are shown in Fig.~\ref{fig:salad-1},
with 10 ``word salad'' and ``nonsense'' versions.
Sampled text using each method is given in Appendix~\ref{sec:construction}.
We re-run each method on the English fiction Gutenberg Corpus with the null versions of each book and verify that the emotional arcs of real stories are not simply an artifact.
The singular value spectrum from the SVD is flatter,
with higher-frequency modes appearing more quickly,
and in total representing 45\% of the total variance present in real stories (see Figs.~\ref{fig:SVD-12-comp-salad} and~\ref{fig:SVD-spectrum-comparison}).
Hierarchical clustering generates less distinct clusters with considerably lower linkage cost (final linkage cost 1400 vs 7000) for the emotional arcs from nonsense books,
and the winning node vectors on a self-organizing map lack coherent structure (see Figs.~\ref{fig:ward-small-salad} and~\ref{fig:SOM-stories-salad} in Appendix~\ref{sec:shuffled}).

\subsection{The Success of Stories}

\begin{figure*}[tbp!]
  \centering
  \includegraphics[width=0.9\textwidth]{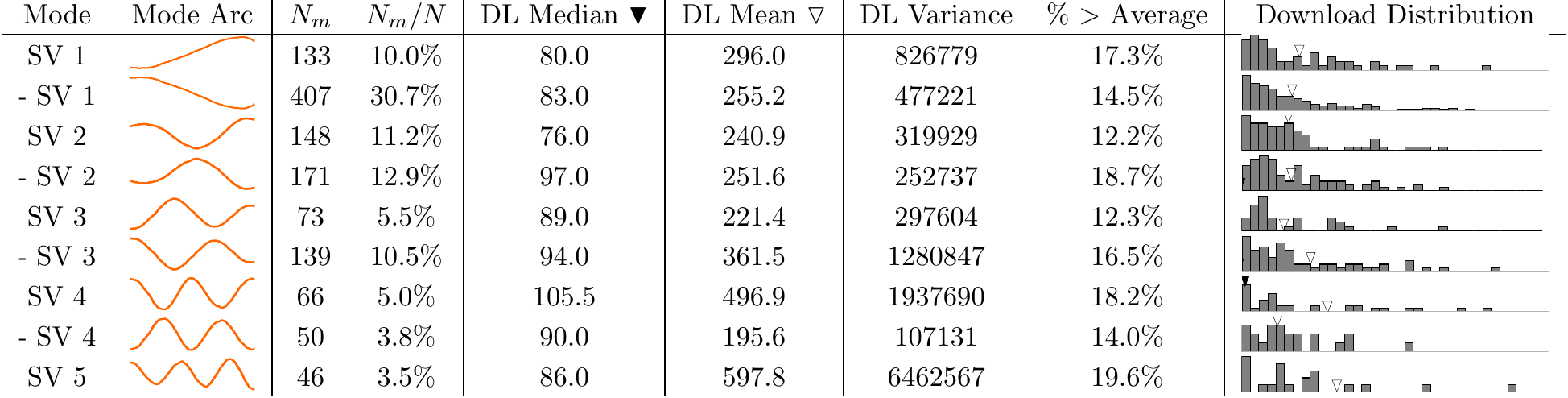}
  \caption[]{
    Download statistics for stories whose SVD Modes comprise more than 2.5\% of books,
    for $N$ the total number of books and $N_m$ the number corresponding to the particular mode.
    Modes \textit{SV 3} through \textit{-SV 4} (both polarities of modes 3 and 4) exhibit a higher average number of downloads and more variance than the others.
    Mode arcs are rows of $V^{T}$ and the download distribution is show in $\log_{10}$ space from 20 to 30,000 downloads.
  }
  \label{fig:download-table-2point5}
\end{figure*}

To examine how the emotional trajectory impacts success,
in Fig.~\ref{fig:download-table-2point5} we examine the downloads for all of the books that are most similar to each SVD mode (for additional modes,
see Fig.~\ref{fig:download-table-point5} in Appendix~\ref{sec:extras}).
We find that the first four modes,
which contain the greatest total number of books,
are not the most popular.
Along with the negative of mode 2, both polarities of modes 3 and 4 have markedly higher median downloads,
while we discount the importance of the mean with the high variance.
The success of the stories underlying these emotional arcs suggests that the emotional experience of readers strongly affects how stories are shared.
We find ``Icarus'' (-SV 2),
``Oedipus'' (-SV 3), and
two sequential ``Man in a hole'' arcs (SV 4),
are the three most successful emotional arcs.
These results are influenced by individual books within each mode which have high numbers of downloads,
and we refer the reader to the download-sorted tables for each mode in Appendix~\ref{sec:SVD-supp}.

\section{Conclusion}
\label{sec:conclusion}

Using three distinct methods,
we have demonstrated that there is strong support for six core emotional arcs.
Our methodology brings to bear a cross section of data science tools with a knowledge of the potential issues that each present.
We have also shown that consideration of the emotional arc for a given story is important for the success of that story.
Of course, downloads are only a rough proxy for success, and this work may provide an outline for more detailed analysis of the factors that impact meaningful measures of success, i.e., sales or cultural influence.

Our approach could be applied in the opposite direction: namely by beginning with the emotional arc and aiding in the generation of compelling stories \cite{li2013story}.
Understanding the emotional arcs of stories may be useful to aid in constructing arguments \cite{bex2010persuasive} and teaching common sense to artificial intelligence systems \cite{riedl2015using}.

Extensions of our analysis that use a more curated selection of full-text fiction can answer more detailed questions about which stories are the most popular throughout time,
and across regions \cite{silva2016a}.
Automatic extraction of character networks would allow a more detailed analysis of plot structure for the Project Gutenberg corpus used here \cite{bost2016a,prado2016a,min2016a}.
Bridging the gap between the full text stories \cite{nenkova2012survey} and systems that analyze plot sequences will allow such systems to undertake studies of this scale \cite{winston2011strong}.
Place could also be used to consider separate character networks through time,
and to help build an analog to Randall Munroe's Movie Narrative Charts \cite{munroe657}.

We are producing data at an ever increasing rate,
including rich sources of stories written to entertain and share knowledge,
from books to television series to news.
Of profound scientific interest will be the degree to which we can eventually understand the full landscape of human stories,
and data driven approaches will play a crucial role.

PSD and CMD acknowledge support from NSF Big Data Grant \#1447634.

\bibliographystyle{unsrtabbrv}

\clearpage
\pagebreak

\newwrite\tempfile
\immediate\openout\tempfile=startsupp.txt
\immediate\write\tempfile{\thepage}
\immediate\closeout\tempfile

\setcounter{page}{1}
\renewcommand{\thepage}{S\arabic{page}}
\renewcommand{\thefigure}{S\arabic{figure}}
\renewcommand{\thetable}{S\arabic{table}}
\setcounter{figure}{0}
\setcounter{table}{0}

\onecolumngrid
\appendix

\section{Plot theories}
\label{sec:plots}

We again emphasize that our method of mining emotional arcs from novels does not measure the popular notion of ``plot''.
To make the distinction even clearer, using terms often employed in narratology we consider the common notion of ``plot'' to be \emph{fabula} whereas the emotional arc is an attempt to measure the emotional trajectory of the \emph{syuzhet}, what could be commonly referred to as the ``structure'' \cite{cobley2005narratology}.
For example, the difference between Booker's \emph{Overcoming the monster} and \emph{Rags to riches} may very well have a similar emotional arc, while being distinct plots.
Nevertheless, we include our research on the different types of plots that have been enumerated.

There have been various hand-coded attempts to enumerate and classify the core types of stories from their plots, including models that generalize broad categories and detailed classification systems.
We consider a range of these theories in turn while noting that plot similarities do not necessitate a concordance of emotional arcs.
\begin{itemize}
\item Three plots: In his 1959 book, Foster-Harris contends that there are three basic patterns of plot (extending from the one central pattern of conflict): the happy ending, the unhappy ending, and the tragedy \cite{harris1959basic}.
In these three versions, the outcome of the story hinges on the nature and fortune of a central character: virtuous, selfish, or struck by fate, respectively.
\item Seven plots: Often espoused as early as elementary school in the United States, we have the notion that plots revolve around the conflict of an individual with either (1) him or herself, (2) nature, (3) another individual, (4) the environment, (5) technology, (6) the supernatural, or (7) a higher power \cite{abbott2008cambridge}.
\item Seven plots: Representing over three decades of work, Christopher Booker's \textit{The Seven Basic Plots: Why we tell stories} describes in great detail seven narrative structures:  \cite{booker2006a}
\begin{itemize}
\item Overcoming the monster (e.g., \textit{Beowulf}).
\item Rags to riches (e.g., \textit{Cinderella}).
\item The quest (e.g., \textit{King Solomon’s Mines}).
\item Voyage and return (e.g., \textit{The Time Machine}).
\item Comedy (e.g., \textit{A Midsummer Night's Dream}).
\item Tragedy (e.g., \textit{Anna Karenina}).
\item Rebirth (e.g., \textit{Beauty and the Beast}).
\end{itemize}
In addition to these seven, Booker contends that the unhappy ending of all but the tragedy are also possible.
\item Twenty plots: In \textit{20 Master Plots}, Ronald Tobias proposes plots that include ``quest'', ``underdog'', ``metamorphosis'', ``ascension'', and ``descension'' \cite{tobias1993master}.
\item Thirty-six plots: In a translation by Lucille Ray, Georges Polti attempts to reconstruct the 36 plots that he posits Gozzi originally enumerated \cite{polti1921thirty}.
These are quite specific and include ``rivalry of kinsmen'', ``all sacrificed for passion'', both involuntary and voluntary ``crimes of love'' (with many more on this theme), ``pursuit'', and ``falling prey to cruelty of misfortune''.
\end{itemize}

\clearpage
\pagebreak
\section{Additional Figures}
\label{sec:extras}

Here we include additional supporting information.

\begin{figure}[!htb]
 \centering
  \includegraphics[width=0.44\textwidth]{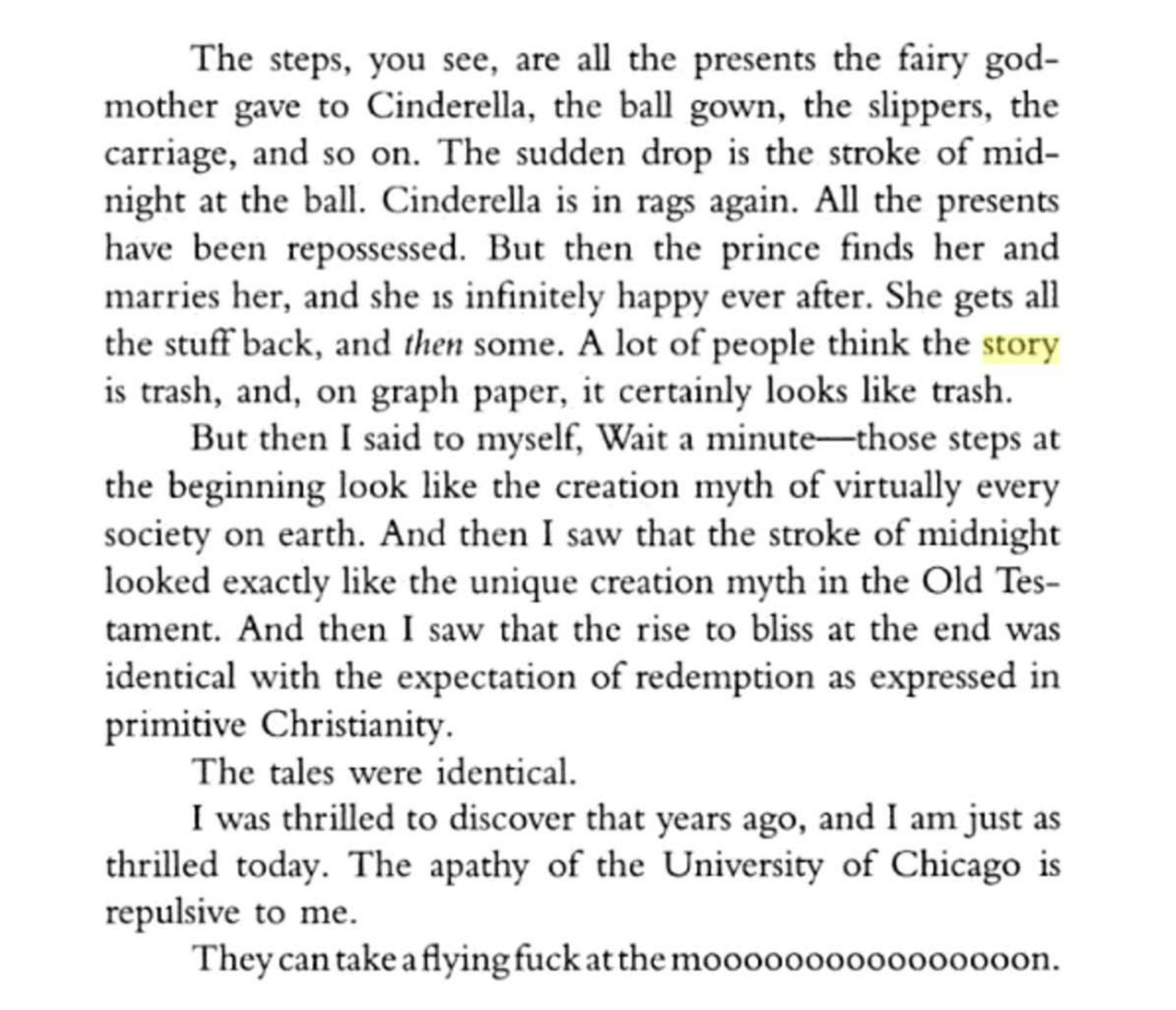}
  \caption{Kurt Vonnegut writes in his autobiography \textit{Palm Sunday} on the similarity of certain story shapes \cite{vonnegut1981}.
  The exposition of this particular similarity would place both of these stories in our grouping of ``Rags to Riches'' emotional arcs.}
  \label{fig:vonnegut}
\end{figure}

\begin{figure}[!htb]
  \centering
  \includegraphics[width=0.9\textwidth]{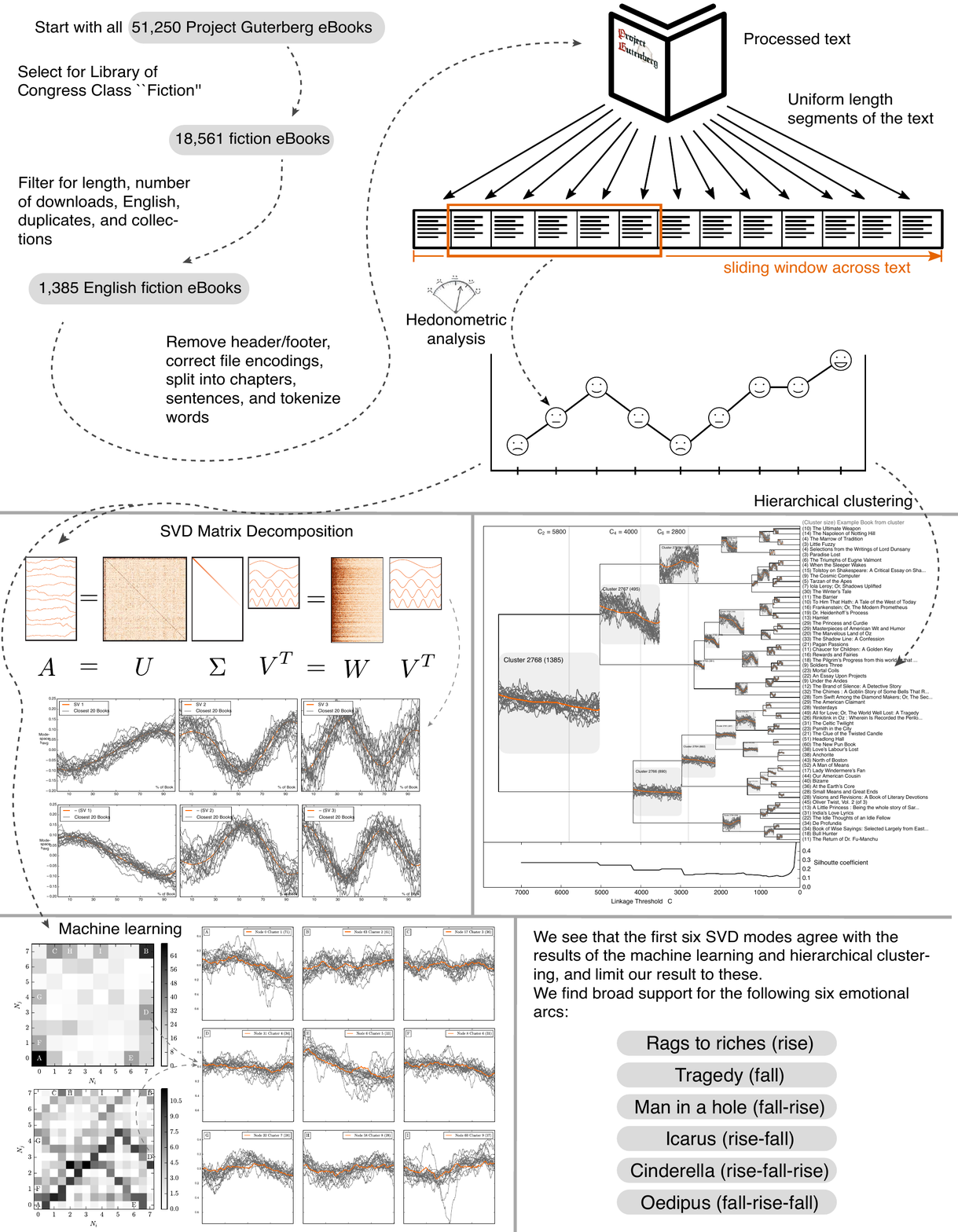}
  \caption[]{
    Schematic (infographic) of the workflow for the entire paper.
  }
  \label{fig:infographic}
\end{figure}

\begin{figure}[!htb]
  \centering
  \includegraphics[width=0.9\textwidth]{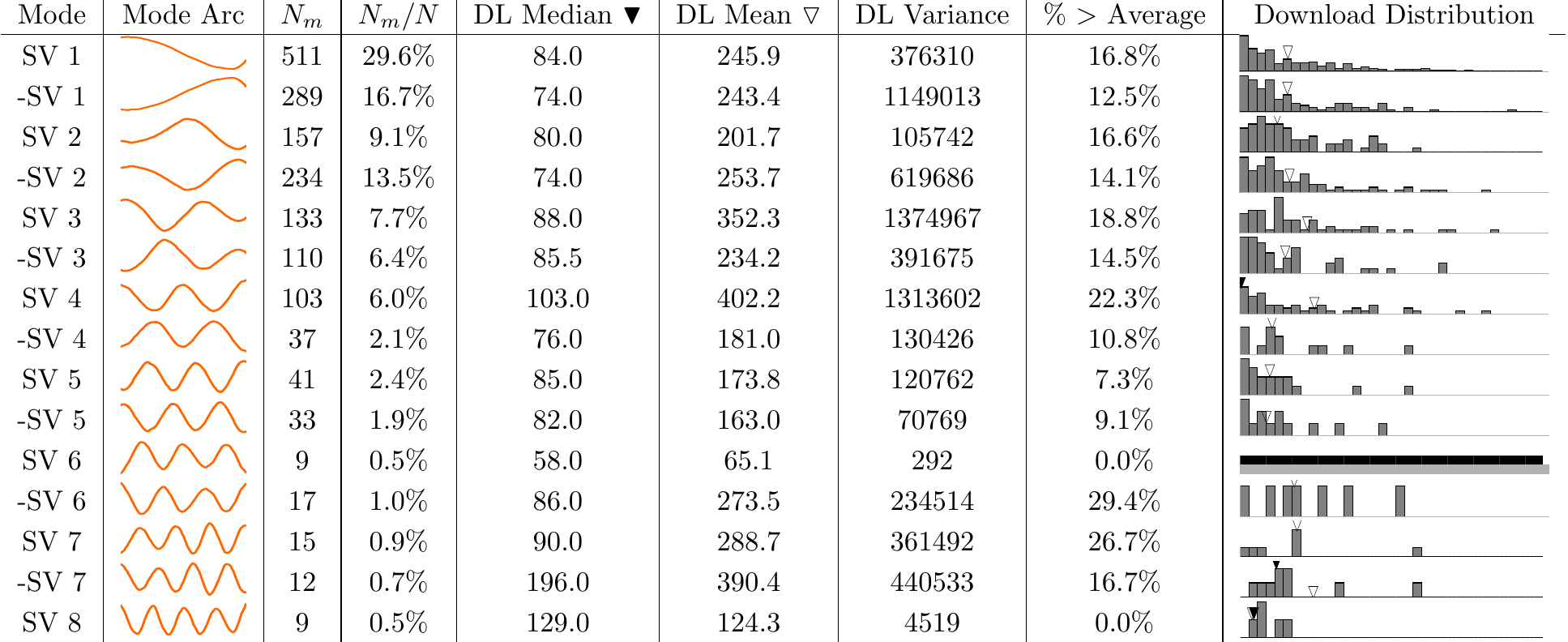}
  \caption[]{
    Download statistics for SVD Modes with more than 0.5\% of books.
  }
  \label{fig:download-table-point5}
\end{figure}

\clearpage
\pagebreak
\section{Emotional Arc Construction}
\label{sec:construction}

To generate emotional arcs, we consider many different approaches with the goal of generating time series that meaningfully reflect the narrative sentiment.
In general, we proceed as described in Fig. \ref{fig:timeseries-schematic} and consider a method of breaking up the text as having three (interdependent) parameter choices for a sliding window:
\begin{enumerate}
\item Length of the desired sample text.
\item Breakpoint between samples.
\item Overlap of each sample.
\end{enumerate}
These methods vary between rating individual words with no overlap to rating the entire text.
To make our choice, we consider competing two objectives of time series generation: meaningfulness of sentiment scores and increased temporal resolution of time series.
For the most accurate sentiment scores, we can use the entire book.
The highest temporal resolution is possible with a sliding window of length 1, generating time series that have potentially as many data points as words in the book.

Since our goal is not only the generation of time series, but the comparison of time series across texts, we consider the additional objective of consistency.
We seek time series which are consistent both in the accuracy of the time series, as well as consistent in the length of the resulting time series.
Again these goals are orthogonal, and we note that our choice here can be tuned to test the sensitivity.

We normalize the length of emotional arcs for books of different length (while using a fixed window size) by varying the amount that the window needs to move.
To make a time series of length $l$ from a book with $N$ words, we fix the sample length at $k$ and set the overlap of samples to $$(N-k-1)/l$$ words.
This guarantees that we have temporal resolution $l$ and sample length $k$ for any $N > k + l$.
We do not consider books with $N \leq k + l$ words.

To generate a sentiment score as in Fig. \ref{fig:timeseries-schematic}, we use a dictionary based approach for transparency and understanding of sentiment.
We select the LabMT dictionary for robust performance over many corpora and best coverage of word usage.
In particular, we determine a sample $T$'s average happiness using the equation:
\begin{equation} 
  h_\textnormal{avg}(T) = 
  \frac{
    \sum^{N}_{i=1} 
    h_{\textnormal{avg}}(w_i)
    \cdot 
    f_i (T)
  }
  {
    \sum^{N}_{i=1} 
    f_i (T)
  }
  = 
  \sum^{N}_{i=1} 
  h_\textnormal{avg} (w_i)
  \cdot 
  p_i (T),
  \label{eq:havg} 
\end{equation}
where we denote each of the $N$ words in a given dictionary as $w_i$,
word sentiment scores as $h_\textnormal{avg}(w_i)$, 
word frequency as $f_i(T)$,
and normalized frequency of $w_i$ in $T$ as
$
p_i (T) 
=
f_i (T) 
/ 
\sum^{N}_{i=1} 
f_i (T)
$.

We note here that, in general, for each emotional arc we subtract the mean before computing the distance or clustering.

\subsection{Null emotional arc construction}
\label{sec:null-construction}

In our first analysis, we generated the null set of emotional arc time series by randomly shuffling the words in each book.
Other variations on generating this null set include sampling from a phrase-level parse of the book with a Markov process, using continuous space random walks directly, or shuffling on sentences.
Even more sophisticated approaches could utilize Recursive Neural Nets (RNNs), for examples see \url{http://karpathy.github.io/2015/05/21/rnn-effectiveness/}.
For a realistic comparison to meaningful stories, we generate ``nonsense'' using a Markov chain model (MCM) trained on 2-grams from each book.
To contrast these approaches, consider the beginning of one null version of \textit{Alice's Adventures in Wonderland} using randomly shuffled words (the ``word salad''):
\begin{quote}
\textit{the  but little --but all the , with I flowers that small a what the he could queer ran it near , and altogether remain A with somebody , gardeners the thought your I the , door head she me hardly of is were said the - them she Alice I But one you nice large use walked what anything 's and It many I , the , execution , she by of came I witness , turned she upon suddenly took While , if I , hear --well goose mouth , do replied the of play would SAID seem , of business shrink 's she flower if--if the Hare the so Alice of , a - very the hear reason to whispered BEE it thought by I large not , your dream on Herald SOUP she I some to her the all of of of guess Perhaps tell to the answer I-- now nibbled , must folded , going himself taught centre wo Northumbria-- hanging can you a went a said ! said all cats do before conversation had of jury baby be Run will again three that herself to ! the Alice ! While the girl neighbour very growing they want , across the , whiting round little , with of a indeed went the on tell offended only forgotten to to , tell n't wo see the hardly Turtle What , on , into three him we ten appear you and at I and which thought makes eyes I it and the looked Is n't baby disappeared , an goes of you all talking ; herself that she in bleeds THAT No in - and-butter how I wash , went a 'll way--never kind at with As Dodo , fear officers been off the opening it , said to , removed at said , went muchness--you for and time court what very will to among Queen Turtle things I so Pigeon herself lie me the naturedly the changed never HER , missed but hurry The March--just the said been   beak-- of , the now whole , Dodo  |}
\end{quote}
and the null version using a 2-gram MCM (the ``nonsense'' version):
\begin{quote}
\textit{But then , thought Alice to herself , after all --SAID I COULD NOT SWIM-- you ca n't go , said the Dormouse began in a minute. And how odd the directions will look. It was the Rabbit 's voice along--'Catch him , I should be like , said the Mouse in the newspapers , at the top of it. The question is , said the Caterpillar. I 'd better ask HER about it. The Queen 's absence , and yet it was n't very civil of you , sooner or later. While she was considering in her life , and that 's a fact. Alice kept her waiting. I ca n't get out of the fact. As for pulling me out of the evening , beautiful Soup. This was such a rule at processions and besides , that finished the first witness , said Alice , and went stamping about , reminding her very much at first but she stopped hastily , for the rest were quite silent for a baby altogether Alice did n't think , said the Queen , who was sitting on a little worried. Sure , it 'll never go THERE again said Alice , who had been to her in such a nice little dog near our house I should say With what porpoise. You do n't seem to put everything upon Bill. And the muscular strength , which remained some time in silence at last she spread out her hand in hand , in chains , with the dream of Wonderland of long ago anything had happened. --as far out to be nothing but the great wonder is , said Alice , with their hands and feet at the flowers and the Queen say only yesterday you deserved to be two people. Here the Dormouse said-- the Hatter , and , after all it might happen any minute , while the Mock Turtle nine the next witness was the Cat again , to be seen--everything seemed to be sure but I shall be a very long silence after this , as it 's coming down. In THAT direction , the Duchess said to Alice a good deal on where you want to go.  Wow wow wow. She 'll get me executed , as the Dormouse go on with the bread and-butter. So they could n't guess of what work it would be like , said the King sharply Do you take me for his housemaid , she pictured to herself , after all. Yes , but it was quite silent for a rabbit. She waited for a minute , nurse. Begin at the house before she had tired herself out with trying , the Queen put on your shoes and stockings for you said the Dodo. How CAN I have n't opened it yet , before Alice could see it trot away quietly into the roof of the Mock Turtle , suddenly dropping his voice , What HAVE you been doing here. It was high time to begin with , the Gryphon added Come , there 's no pleasing them. Alice remained looking thoughtfully at the other , saying to herself , whenever I eat or drink anything so I should think you 'd like it , said the Caterpillar. Ugh said the King.}
\end{quote}

\subsection{Further Gutenberg Processing}

Here we provide the details of the processing applied to the Gutenberg corpus.
In the manuscript, we stated the following:
\begin{quote}
  We start by selecting for only English books, with total words between 20,000 and 100,000, with more than 20 downloads from the Project Gutenberg website, and with Library of Congress Class PN, PR, PS, or PZ.
  Next, we remove books with any word in the title from a list of keywords (e.g. ``poems'' and ``collection'').
  From within this set of books, we remove the front and backmatter of each book using regular expression pattern matches.
\end{quote}
The full list of keywords which we used to filter the titles are the following: ``stories'', ``collection'', ``poems'', ``complete'', ``essays'', ``fables'', ``tales'', ``papers'', ``poetry'', ``verses'', ``ballads'', ``sketches'', ``vol.'', ``vols.'', ``works'', ``volume'', and ``other''.
A list of of LoC Classes is given at \url{https://www.loc.gov/catdir/cpso/lcco/}.

To remove the front matter, we first detect the end of the front matter by matching for either \verb|START OF THIS PROJECT GUTENBERG EBOOK| in the line or \verb|START OF THE PROJECT GUTENBERG EBOOK|.
If neither of these work, we look for a line that contains both \verb|END| and \verb|SMALL PRINT| in the line, in the first half of the text.

To remove the back matter, we check for three different endings, in order.
First, similar to the front matter we check, here without being sensitive for case, for \verb|END OF THIS PROJECT GUTENBERG EBOOK| or \verb|END OF THE PROJECT GUTENBERG EBOOK| or \verb|END OF PROJECT GUTENBERG|.
Next, we check the last 25\% of the book, case insensitive, for the words \verb|END| and \verb|PROJECT GUTENBERG|.
Finally, we check the last 10\% of the book for the words, case sensitively, \verb|THE END|.

Together, these filters each remove text from the beginning and end of 98.9\% of ebooks.
The first pass in each case works for 78.9\% of cases.
On average, this removes less than 1\% of the beginning lines, and 3-4\% of the ending lines.

\begin{figure}[tbp!]
  \centering
  \includegraphics[width=0.96\textwidth]{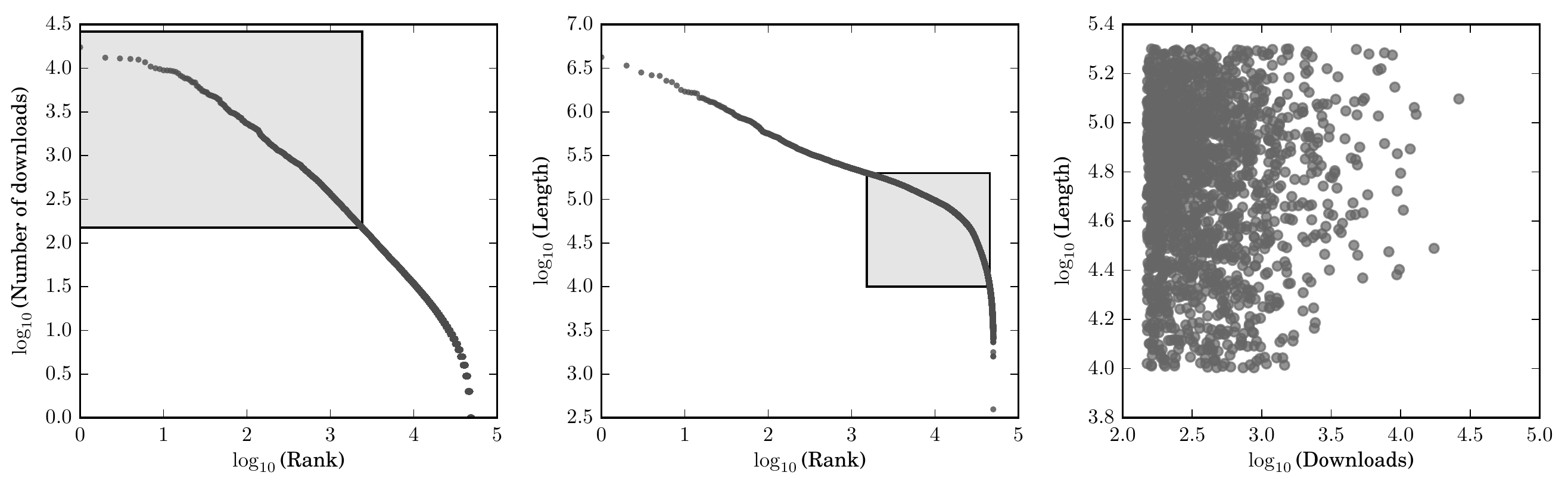}
  \caption[]{
    Rank-frequency distributions of book downloads and length in the Gutenberg corpus: (A) downloads, (B) book length in words, and (C) both downloads and length.
    We filter by both number of downloads and book length to select for fiction books, with the filters shown as gray boxes in Panels A and B.
    In Panel C, we plot each of 1,748 books selected by their download count and length, shown in download-length space.
  }
  \label{fig:length-distribution}
\end{figure}

\clearpage
\pagebreak
\section{Book list}
\label{sec:lists}

We include a list of all books used in this study with more than 40 downloads from Project Gutenberg, such that we list those from all of the experiments with 40 and 80 download thresholds in the following Table.
We do not include the full list of books with more than 10 downloads for brevity, as it is more than 90 pages long (this list is 22 pages).

\begin{longtable}{l | p{6.5cm} | l | l | l | l | l | l}
\caption{All Project Gutenberg eBooks considered in this study, sorted by downloads.}\\
ID & Title & DL's & Mode & $N_w$ & $\textrm{Uniq}(N_w)$ & $h_{\textrm{avg}}^{b_i}$ & LoCC\\
\hline
\endhead
\href{http://www.gutenberg.org/ebooks/11}{11} & Alice's Adventures in Wonderland & 17,366 & SV5 & 27,386 & 3,973 & 5.99 & PR,PZ\\
\href{http://www.gutenberg.org/ebooks/84}{84} & Frankenstein; Or, The Modern Pro... & 11,699 & -SV3 & 77,680 & 9,691 & 5.89 & PR\\
\href{http://www.gutenberg.org/ebooks/74}{74} & The Adventures of Tom Sawyer & 9,454 & SV4 & 73,265 & 11,094 & 5.76 & PZ,PS\\
\href{http://www.gutenberg.org/ebooks/844}{844} & The Importance of Being Earnest:... & 9,373 & SV1 & 23,161 & 3,612 & 6.15 & PR\\
\href{http://www.gutenberg.org/ebooks/174}{174} & The Picture of Dorian Gray & 7,652 & -SV1 & 84,591 & 9,588 & 5.96 & PR\\
\href{http://www.gutenberg.org/ebooks/16}{16} & Peter Pan & 5,789 & SV4 & 48,189 & 6,747 & 5.89 & PZ,PR\\
\href{http://www.gutenberg.org/ebooks/16328}{16328} & Beowulf
: An Anglo-Saxon Epic Poem & 5,359 & -SV1 & 44,949 & 9,351 & 5.64 & PR\\
\href{http://www.gutenberg.org/ebooks/42}{42} & The Strange Case of Dr. Jekyll a... & 4,908 & -SV1 & 26,085 & 5,300 & 5.63 & PR\\
\end{longtable}

\clearpage
\pagebreak
\section{Principal Component Analysis (SVD)}
\label{sec:SVD-supp}

In this section we provide a (1) more in-depth, intuitive explanation of the method and (2) more results from the SVD analysis.

\begin{figure*}[!htb]
  \centering
  \includegraphics[width=0.71\textwidth]{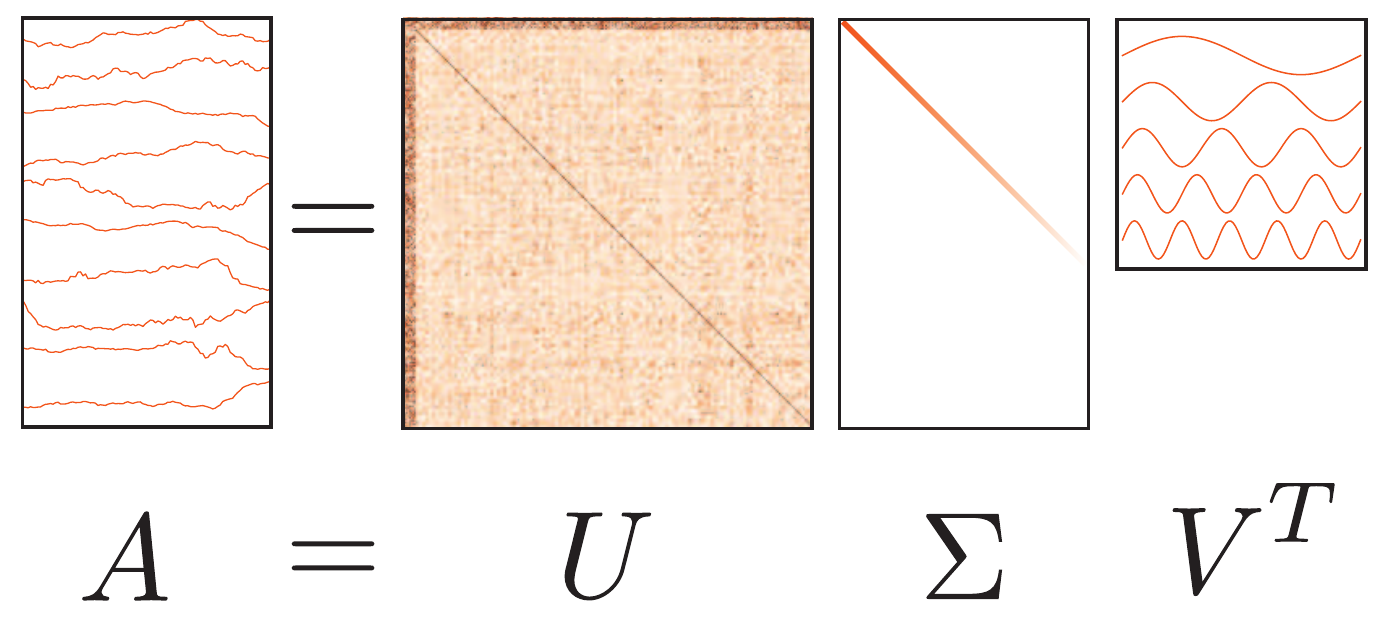}
  \caption[]{
    Schematic of the Singular Value Decomposition applied to emotional arcs of Project Gutenberg books.
    Shown in $A$ are 10 randomly chosen emotional arcs, in $U$ a ``spy'' of the matrix, in $\Sigma$ the decreasing singular values, and in $V^T$ sinusoidal modes.
    We emphasize that this representation is purely for intuition, as only $U$ is a image of the actual matrix, and $A$ has only 10 of the \nbooks~books.
  }
  \label{fig:SVD-USV}
\end{figure*}

In an effort to develop a better intuition for the results of the principal component analysis by way of SVD, we plot Eq.~\ref{eq:SVD} along with representations of the matrices in Fig.~\ref{fig:SVD-USV}.

Further, we considered in Eq.~\ref{eq:SVD} the mode coefficient in the matrix $W$, and in Fig~\ref{fig:SVD-WV} we plot the second line of the equation with $W$:

\begin{figure*}[!htb]
  \centering
  \includegraphics[width=0.51\textwidth]{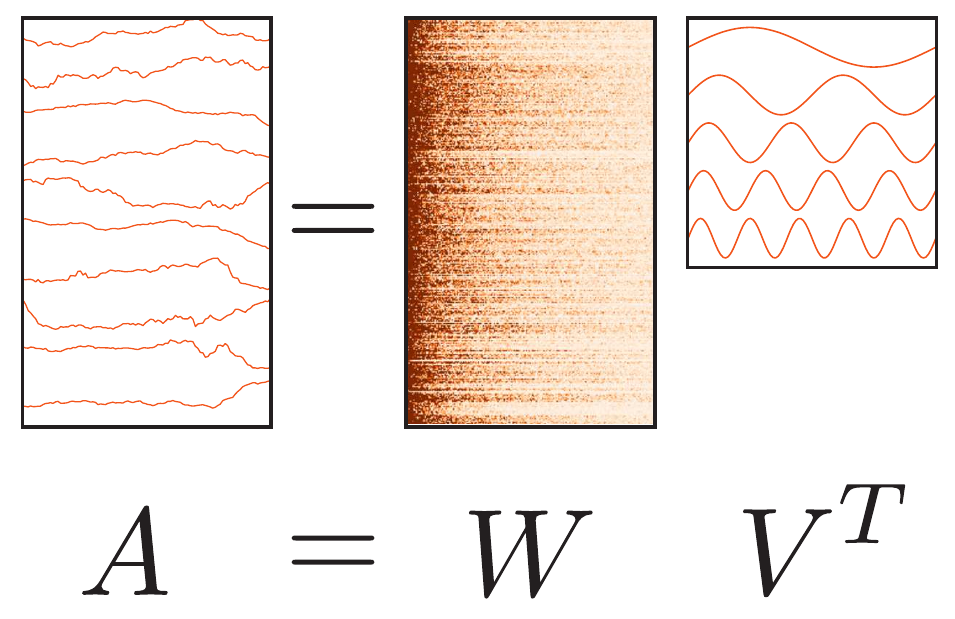}
  \caption[]{
    Schematic of the Singular Value Decomposition applied to emotional arcs of Project Gutenberg books, with $W=U\Sigma$ containing the mode coefficients.
    Again shown in $A$ are 10 randomly chosen emotional arcs, in $W$ a ``spy'' of the matrix used in the analysis, and in $V^T$ representative sinusoidal modes.
  }
  \label{fig:SVD-WV}
\end{figure*}

With $A$ written as $W\cdot V^T$, the coefficients for each mode (row of $V^T$) for a book $i$ are given as the rows of $W$.
To reconstruct the emotional arc of book $i$, using mode $j$ from $V^T$, we simply multiply $W[i,j] \cdot V^T[j,:]$. Shown below in Fig.~\ref{fig:SVD-reconstruction}, we built the emotional arc for an example story using only the first mode through the first 12 modes.

\begin{figure*}[!htb]
  \centering
  \includegraphics[width=0.98\textwidth]{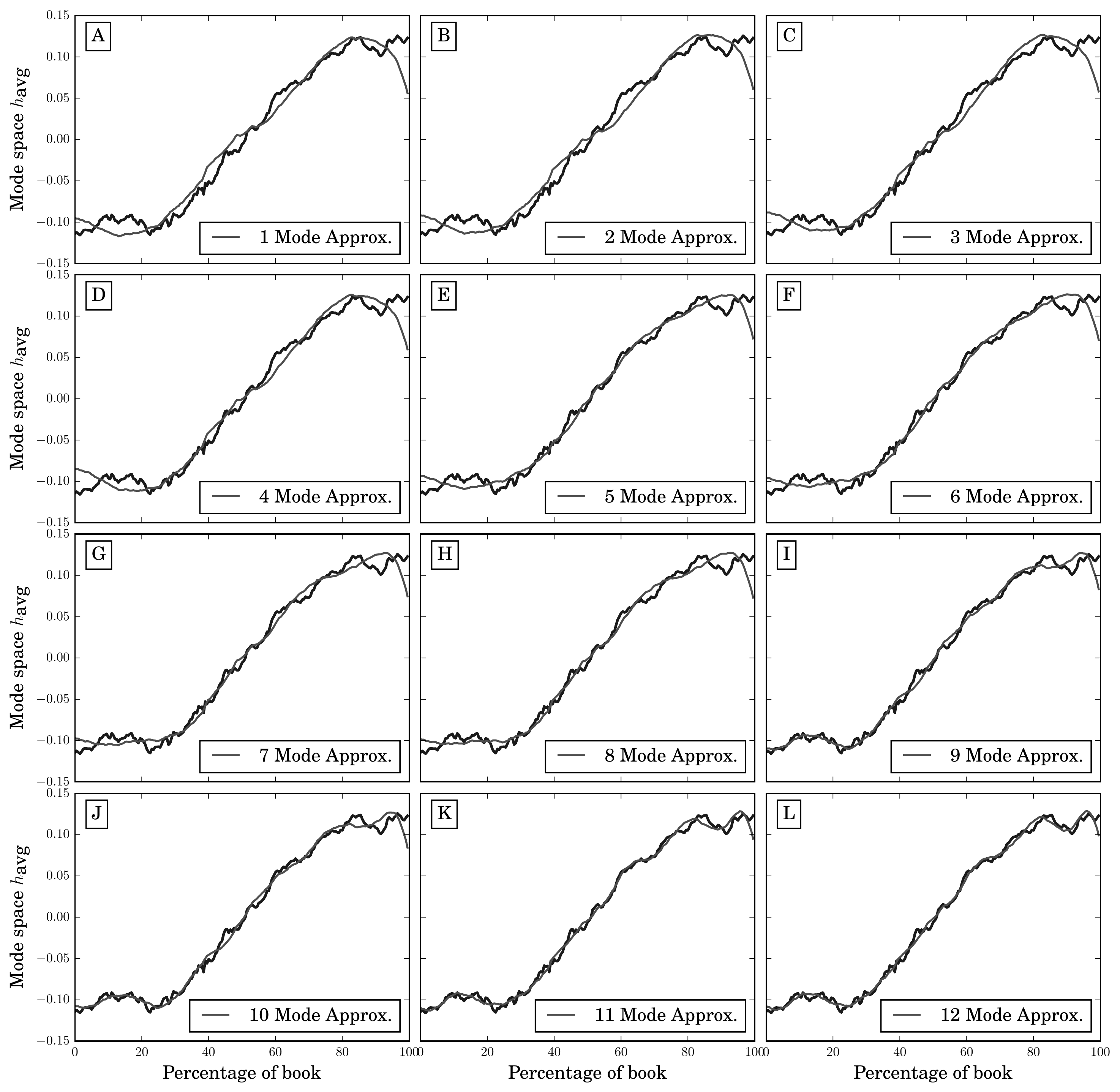}
  \caption[]{
    Reconstruction of the emotional arc from \textit{Alice's Adventures Under Ground}, by Lewis Carroll.
    The addition of more modes from the SVD more closely reconstructs the detailed emotional arc.
    This book is well represented by the first mode alone, with only minor corrections from modes 2-11, as we should expect for a book whose emotional arc so closely resembles the ``Rags to Riches'' arc.
  }
  \label{fig:SVD-reconstruction}
\end{figure*}

\clearpage
\pagebreak
\subsection{Additional details for 40 download threshold}

\begin{figure*}[!htb]
  \centering
  \includegraphics[width=0.98\textwidth]{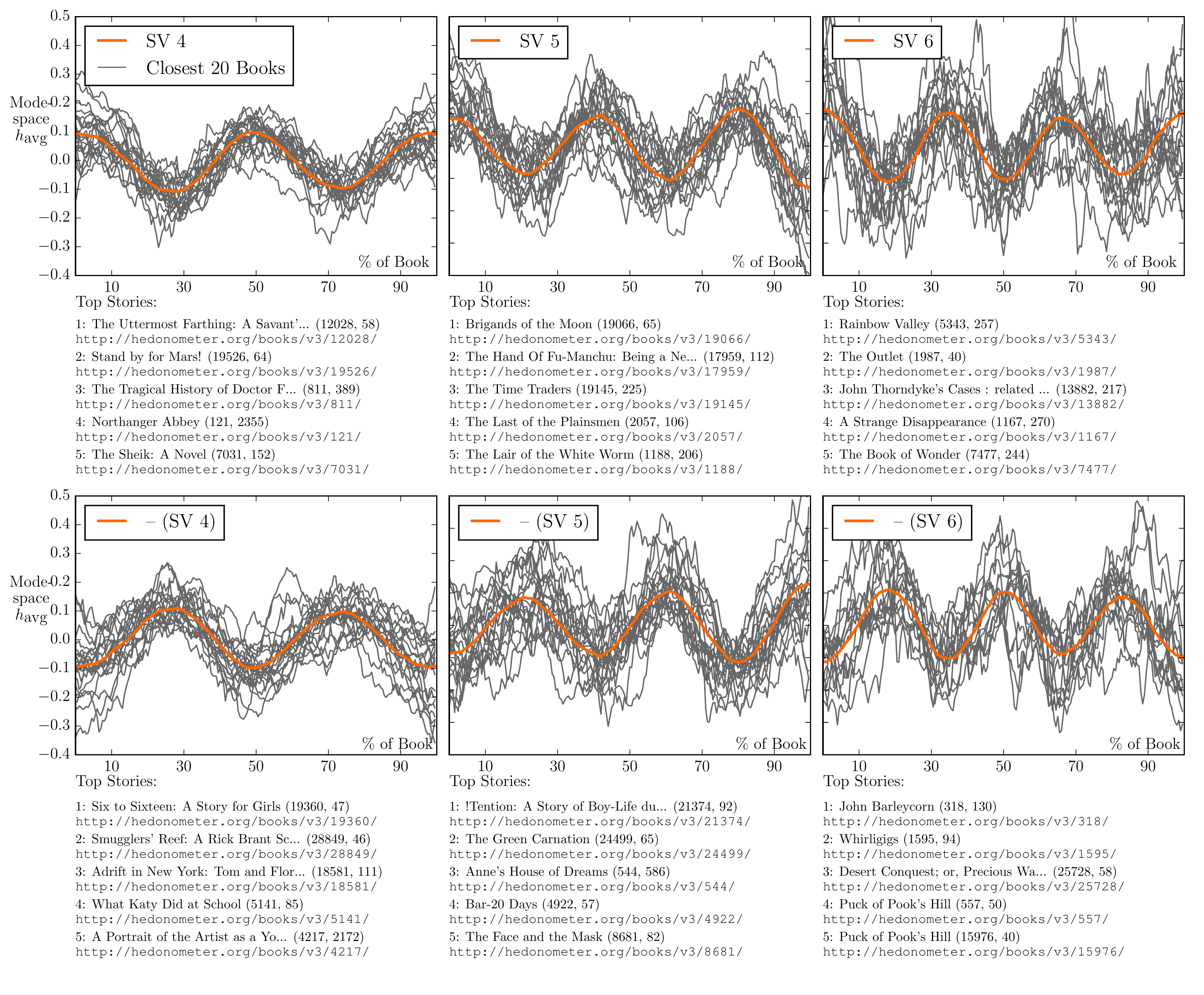}
  \caption[]{
    SVD modes 4--6 (and their negation) with closest stories.
    Again, to show the emotional arcs on the same scale as the modes, we show the modes directly from the rows of $V^T$ and weight the emotional arcs by the inverse of their coefficient in $W$ for the particular mode.
    Shown in parenthesis for each story is the Project Gutenberg ID and the number of downloads from the Project Gutenberg website, respectively.
    Links below each story point to an interactive visualization on \url{http://hedonometer.org} which enables detailed exploration of the emotional arc for the story.
  }
  \label{fig:SVD-4-6-labelled}
\end{figure*}

First, we consider modes 4--6 and their closest stories in Fig. \ref{fig:SVD-4-6-labelled}.

Next, we provide a full list of the books closest to each mode in the analysis, both sorted by downloads and support from the mode.
\textbf{Note: we are unable to provide a figure accompanying each arc in the following table due to size restrictions on the arXiv.}

\begin{longtable}{l | l | l | l}
\caption{Stories which are are represented by core emotional arc 1, sorted by the variance explained in their emotional arc by core emotional arc 1.}\\
ID & Title & DL's & max($W$[i,:])\\
\hline
\endhead
\href{http://www.gutenberg.org/ebooks/1539}{1539} & The Winter's Tale & 73 & 0.5217\\
\href{http://www.gutenberg.org/ebooks/33689}{33689} & Oscar Wilde, Art and Morality: A... & 88 & 0.3804\\
\href{http://www.gutenberg.org/ebooks/35617}{35617} & The Terror: A Mystery & 61 & 0.3734\\
\href{http://www.gutenberg.org/ebooks/7088}{7088} & The Pilgrim's Progress in Words ... & 55 & 0.3522\\
\href{http://www.gutenberg.org/ebooks/26624}{26624} & The Road to Oz & 68 & 0.3412\\
\href{http://www.gutenberg.org/ebooks/21130}{21130} & Book of Wise Sayings: Selected La... & 86 & 0.3361\\
\href{http://www.gutenberg.org/ebooks/2248}{2248} & The Winter's Tale & 137 & 0.3355\\
\href{http://www.gutenberg.org/ebooks/485}{485} & The Road to Oz & 178 & 0.3320\\
\end{longtable}
\begin{longtable}{l | l | l | l}
\caption{Top 10 stories which are are represented by core emotional arc 1, sorted by downloads.}\\
ID & Title & DL's & max(W[i,:])\\
\hline
\endhead
\href{http://www.gutenberg.org/ebooks/844}{844} & The Importance of Being Earnest:... & 9,373 & 0.0963  \\
\href{http://www.gutenberg.org/ebooks/236}{236} & The Jungle Book & 3,478 & 0.1728  \\
\href{http://www.gutenberg.org/ebooks/215}{215} & The Call of the Wild & 2,439 & 0.2046  \\
\href{http://www.gutenberg.org/ebooks/779}{779} & The Tragical History of Doctor F... & 2,133 & 0.1133  \\
\href{http://www.gutenberg.org/ebooks/35997}{35997} & The Jungle Book & 1,370 & 0.1623  \\
\href{http://www.gutenberg.org/ebooks/885}{885} & An Ideal Husband & 1,303 & 0.2557  \\
\href{http://www.gutenberg.org/ebooks/78}{78} & Tarzan of the Apes & 1,272 & 0.1341  \\
\href{http://www.gutenberg.org/ebooks/500}{500} & The Adventures of Pinocchio & 863 & 0.2108  \\
\href{http://www.gutenberg.org/ebooks/269}{269} & Beasts and Super-Beasts & 804 & 0.2271  \\
\end{longtable}
\begin{longtable}{l | l | l | l}
\caption{Stories which are are represented by core emotional arc 2, sorted by the variance explained in their emotional arc by core emotional arc 2.}\\
ID & Title & DL's & max($W$[i,:])\\
\hline
\endhead
\href{http://www.gutenberg.org/ebooks/946}{946} & Lady Susan & 894 & 0.5006 \\
\href{http://www.gutenberg.org/ebooks/17958}{17958} & Warlord of Kor & 70 & 0.4664 \\
\href{http://www.gutenberg.org/ebooks/17144}{17144} & The House of the Vampire & 188 & 0.4657 \\
\href{http://www.gutenberg.org/ebooks/93}{93} & Tom Sawyer, Detective & 112 & 0.4566 \\
\href{http://www.gutenberg.org/ebooks/159}{159} & The Island of Doctor Moreau & 1,083 & 0.4538 \\
\href{http://www.gutenberg.org/ebooks/32208}{32208} & The Star Lord & 58 & 0.4504 \\
\href{http://www.gutenberg.org/ebooks/790}{790} & Lady Windermere's Fan & 485 & 0.4257 \\
\href{http://www.gutenberg.org/ebooks/1777}{1777} & Romeo and Juliet & 186 & 0.4013 \\
\end{longtable}
\begin{longtable}{l | l | l | l}
\caption{Top 10 stories which are are represented by core emotional arc 2, sorted by downloads.}\\
ID & Title & DL's & max($W$[i,:])\\
\hline
\endhead
\href{http://www.gutenberg.org/ebooks/174}{174} & The Picture of Dorian Gray & 7,652 & 0.1893 \\
\href{http://www.gutenberg.org/ebooks/16328}{16328} & Beowulf
: An Anglo-Saxon Epic Poem & 5,359 & 0.1584 \\
\href{http://www.gutenberg.org/ebooks/42}{42} & The Strange Case of Dr. Jekyll a... & 4,908 & 0.2379 \\
\href{http://www.gutenberg.org/ebooks/526}{526} & Heart of Darkness & 4,362 & 0.2597 \\
\href{http://www.gutenberg.org/ebooks/35}{35} & The Time Machine & 3,732 & 0.3199 \\
\href{http://www.gutenberg.org/ebooks/3825}{3825} & Pygmalion & 3,580 & 0.2446 \\
\href{http://www.gutenberg.org/ebooks/2852}{2852} & The Hound of the Baskervilles & 3,358 & 0.1951 \\
\href{http://www.gutenberg.org/ebooks/219}{219} & Heart of Darkness & 3,243 & 0.2184 \\
\href{http://www.gutenberg.org/ebooks/28520}{28520} & Forbidden Fruit: Luscious and exc... & 2,716 & 0.2028 \\
\end{longtable}
\begin{longtable}{l | l | l | l}
\caption{Stories which are are represented by core emotional arc 3, sorted by the variance explained in their emotional arc by core emotional arc 3.}\\
ID & Title & DL's & max($W$[i,:])\\
\hline
\endhead
\href{http://www.gutenberg.org/ebooks/419}{419} & The Magic of Oz & 186 & 0.3150\\
\href{http://www.gutenberg.org/ebooks/10736}{10736} & Children of the Frost & 82 & 0.3083\\
\href{http://www.gutenberg.org/ebooks/1094}{1094} & Tamburlaine the Great — Part 1 & 474 & 0.3081\\
\href{http://www.gutenberg.org/ebooks/520}{520} & The Life and Adventures of Santa... & 76 & 0.3041\\
\href{http://www.gutenberg.org/ebooks/2911}{2911} & Justice & 50 & 0.2989\\
\href{http://www.gutenberg.org/ebooks/18768}{18768} & The Sky Is Falling & 113 & 0.2978\\
\href{http://www.gutenberg.org/ebooks/29774}{29774} & A Yankee Flier Over Berlin & 42 & 0.2864\\
\href{http://www.gutenberg.org/ebooks/19726}{19726} & The Door Through Space & 201 & 0.2826\\
\end{longtable}
\begin{longtable}{l | l | l | l}
\caption{Top 10 stories which are are represented by core emotional arc 3, sorted by downloads.}\\
ID & Title & DL's & max(W[i,:])\\
\hline
\endhead
\href{http://www.gutenberg.org/ebooks/2814}{2814} & Dubliners & 4,742 & 0.2583  \\
\href{http://www.gutenberg.org/ebooks/55}{55} & The Wonderful Wizard of Oz & 3,035 & 0.1770  \\
\href{http://www.gutenberg.org/ebooks/12}{12} & Through the Looking-Glass & 2,892 & 0.0742  \\
\href{http://www.gutenberg.org/ebooks/834}{834} & The Memoirs of Sherlock Holmes & 2,164 & 0.1476  \\
\href{http://www.gutenberg.org/ebooks/1837}{1837} & The Prince and the Pauper & 1,389 & 0.1604  \\
\href{http://www.gutenberg.org/ebooks/447}{447} & Maggie: A Girl of the Streets & 1,295 & 0.1616  \\
\href{http://www.gutenberg.org/ebooks/113}{113} & The Secret Garden & 1,153 & 0.1416  \\
\href{http://www.gutenberg.org/ebooks/421}{421} & Kidnapped & 1,132 & 0.1274  \\
\href{http://www.gutenberg.org/ebooks/5230}{5230} & The Invisible Man: A Grotesque R... & 1,011 & 0.1250  \\
\end{longtable}
\begin{longtable}{l | l | l | l}
\caption{Stories which are are represented by core emotional arc 4, sorted by the variance explained in their emotional arc by core emotional arc 4.}\\
ID & Title & DL's & max($W$[i,:])\\
\hline
\endhead
\href{http://www.gutenberg.org/ebooks/34215}{34215} & Shadowings & 63 & 0.4141 \\
\href{http://www.gutenberg.org/ebooks/12384}{12384} & Battle-Pieces and Aspects of the... & 194 & 0.3546 \\
\href{http://www.gutenberg.org/ebooks/36281}{36281} & The Slayer of Souls & 63 & 0.3283 \\
\href{http://www.gutenberg.org/ebooks/17412}{17412} & The Bobbsey Twins
: Or, Merry Day... & 69 & 0.3259 \\
\href{http://www.gutenberg.org/ebooks/2727}{2727} & Allan's Wife & 128 & 0.3124 \\
\href{http://www.gutenberg.org/ebooks/27761}{27761} & Hamlet, Prince of Denmark & 301 & 0.3057 \\
\href{http://www.gutenberg.org/ebooks/363}{363} & The Oakdale Affair & 91 & 0.2968 \\
\href{http://www.gutenberg.org/ebooks/17854}{17854} & The Sport of the Gods & 61 & 0.2855 \\
\end{longtable}
\begin{longtable}{l | l | l | l}
\caption{Top 10 stories which are are represented by core emotional arc 4, sorted by downloads.}\\
ID & Title & DL's & max($W$[i,:])\\
\hline
\endhead
\href{http://www.gutenberg.org/ebooks/46}{46} & A Christmas Carol in Prose; Bein... & 4,602 & 0.1917 \\
\href{http://www.gutenberg.org/ebooks/20}{20} & Paradise Lost & 2,522 & 0.1475 \\
\href{http://www.gutenberg.org/ebooks/62}{62} & A Princess of Mars & 2,515 & 0.1032 \\
\href{http://www.gutenberg.org/ebooks/308}{308} & Three Men in a Boat & 2,059 & 0.0815 \\
\href{http://www.gutenberg.org/ebooks/5131}{5131} & Childe Harold's Pilgrimage & 1,481 & 0.1987 \\
\href{http://www.gutenberg.org/ebooks/972}{972} & The Devil's Dictionary & 1,257 & 0.0936 \\
\href{http://www.gutenberg.org/ebooks/3289}{3289} & The Valley of Fear & 1,228 & 0.1845 \\
\href{http://www.gutenberg.org/ebooks/131}{131} & The Pilgrim's Progress from this... & 1,126 & 0.1888 \\
\href{http://www.gutenberg.org/ebooks/28522}{28522} & Laura Middleton; Her Brother and... & 1,097 & 0.1851 \\
\end{longtable}
\begin{longtable}{l | l | l | l}
\caption{Stories which are are represented by core emotional arc 5, sorted by the variance explained in their emotional arc by core emotional arc 5.}\\
ID & Title & DL's & max($W$[i,:])\\
\hline
\endhead
\href{http://www.gutenberg.org/ebooks/17763}{17763} & The Mystery of the Hasty Arrow & 93 & 0.2301\\
\href{http://www.gutenberg.org/ebooks/5317}{5317} & Through the Magic Door & 81 & 0.2299\\
\href{http://www.gutenberg.org/ebooks/13944}{13944} & After London; Or, Wild England & 146 & 0.2259\\
\href{http://www.gutenberg.org/ebooks/12590}{12590} & The Shadow of the Rope & 75 & 0.2202\\
\href{http://www.gutenberg.org/ebooks/35247}{35247} & That Affair at Elizabeth & 62 & 0.2179\\
\href{http://www.gutenberg.org/ebooks/39827}{39827} & The Conduct of Life & 176 & 0.2041\\
\href{http://www.gutenberg.org/ebooks/295}{295} & The Early Short Fiction of Edith... & 49 & 0.2012\\
\href{http://www.gutenberg.org/ebooks/10459}{10459} & The Celtic Twilight & 146 & 0.1974\\
\end{longtable}
\begin{longtable}{l | l | l | l}
\caption{Top 10 stories which are are represented by core emotional arc 5, sorted by downloads.}\\
ID & Title & DL's & max(W[i,:])\\
\hline
\endhead
\href{http://www.gutenberg.org/ebooks/120}{120} & Treasure Island & 4,402 & 0.1620  \\
\href{http://www.gutenberg.org/ebooks/696}{696} & The Castle of Otranto & 1,663 & 0.1465  \\
\href{http://www.gutenberg.org/ebooks/2166}{2166} & King Solomon's Mines & 788 & 0.1652  \\
\href{http://www.gutenberg.org/ebooks/1280}{1280} & Spoon River Anthology & 671 & 0.1297  \\
\href{http://www.gutenberg.org/ebooks/1212}{1212} & Love and Freindship [sic] & 611 & 0.1657  \\
\href{http://www.gutenberg.org/ebooks/708}{708} & The Princess and the Goblin & 579 & 0.0996  \\
\href{http://www.gutenberg.org/ebooks/1164}{1164} & The Iron Heel & 506 & 0.1395  \\
\href{http://www.gutenberg.org/ebooks/2243}{2243} & The Merchant of Venice & 506 & 0.0787  \\
\href{http://www.gutenberg.org/ebooks/394}{394} & Cranford & 285 & 0.1029  \\
\end{longtable}
\begin{longtable}{l | l | l | l}
\caption{Stories which are are represented by core emotional arc 6, sorted by the variance explained in their emotional arc by core emotional arc 6.}\\
ID & Title & DL's & max($W$[i,:])\\
\hline
\endhead
\href{http://www.gutenberg.org/ebooks/18172}{18172} & This World Is Taboo & 64 & 0.3281 \\
\href{http://www.gutenberg.org/ebooks/339}{339} & Old Indian Days & 139 & 0.3148 \\
\href{http://www.gutenberg.org/ebooks/10377}{10377} & The Evil Guest & 93 & 0.2868 \\
\href{http://www.gutenberg.org/ebooks/29448}{29448} & Pariah Planet & 96 & 0.2864 \\
\href{http://www.gutenberg.org/ebooks/289}{289} & The Wind in the Willows & 1,475 & 0.2750 \\
\href{http://www.gutenberg.org/ebooks/20727}{20727} & The Cosmic Computer & 221 & 0.2716 \\
\href{http://www.gutenberg.org/ebooks/27805}{27805} & The Wind in the Willows & 543 & 0.2702 \\
\href{http://www.gutenberg.org/ebooks/27991}{27991} & The Blue Bird for Children: The W... & 111 & 0.2677 \\
\end{longtable}
\begin{longtable}{l | l | l | l}
\caption{Top 10 stories which are are represented by core emotional arc 6, sorted by downloads.}\\
ID & Title & DL's & max($W$[i,:])\\
\hline
\endhead
\href{http://www.gutenberg.org/ebooks/84}{84} & Frankenstein; Or, The Modern Pro... & 11,699 & 0.1602 \\
\href{http://www.gutenberg.org/ebooks/244}{244} & A Study in Scarlet & 4,535 & 0.1955 \\
\href{http://www.gutenberg.org/ebooks/863}{863} & The Mysterious Affair at Styles & 3,112 & 0.1937 \\
\href{http://www.gutenberg.org/ebooks/36}{36} & The War of the Worlds & 2,496 & 0.1060 \\
\href{http://www.gutenberg.org/ebooks/2097}{2097} & The Sign of the Four & 2,283 & 0.1199 \\
\href{http://www.gutenberg.org/ebooks/209}{209} & The Turn of the Screw & 2,175 & 0.2251 \\
\href{http://www.gutenberg.org/ebooks/289}{289} & The Wind in the Willows & 1,475 & 0.2750 \\
\href{http://www.gutenberg.org/ebooks/10007}{10007} & Carmilla & 1,416 & 0.1333 \\
\href{http://www.gutenberg.org/ebooks/73}{73} & The Red Badge of Courage: An Epi... & 1,163 & 0.1494 \\
\end{longtable}
\begin{longtable}{l | l | l | l}
\caption{Stories which are are represented by core emotional arc 7, sorted by the variance explained in their emotional arc by core emotional arc 7.}\\
ID & Title & DL's & max($W$[i,:])\\
\hline
\endhead
\href{http://www.gutenberg.org/ebooks/12028}{12028} & The Uttermost Farthing: A Savant'... & 58 & 0.2612\\
\href{http://www.gutenberg.org/ebooks/19526}{19526} & Stand by for Mars! & 64 & 0.2425\\
\href{http://www.gutenberg.org/ebooks/811}{811} & The Tragical History of Doctor F... & 389 & 0.2245\\
\href{http://www.gutenberg.org/ebooks/121}{121} & Northanger Abbey & 2,355 & 0.2196\\
\href{http://www.gutenberg.org/ebooks/7031}{7031} & The Sheik: A Novel & 152 & 0.2071\\
\href{http://www.gutenberg.org/ebooks/24035}{24035} & The Pirates of Ersatz & 113 & 0.2017\\
\href{http://www.gutenberg.org/ebooks/2429}{2429} & Lost Face & 190 & 0.1975\\
\href{http://www.gutenberg.org/ebooks/74}{74} & The Adventures of Tom Sawyer & 9,454 & 0.1906\\
\end{longtable}
\begin{longtable}{l | l | l | l}
\caption{Top 10 stories which are are represented by core emotional arc 7, sorted by downloads.}\\
ID & Title & DL's & max(W[i,:])\\
\hline
\endhead
\href{http://www.gutenberg.org/ebooks/74}{74} & The Adventures of Tom Sawyer & 9,454 & 0.1906  \\
\href{http://www.gutenberg.org/ebooks/16}{16} & Peter Pan & 5,789 & 0.1197  \\
\href{http://www.gutenberg.org/ebooks/4517}{4517} & Ethan Frome & 2,895 & 0.1220  \\
\href{http://www.gutenberg.org/ebooks/121}{121} & Northanger Abbey & 2,355 & 0.2196  \\
\href{http://www.gutenberg.org/ebooks/8492}{8492} & The King in Yellow & 1,504 & 0.1354  \\
\href{http://www.gutenberg.org/ebooks/910}{910} & White Fang & 1,178 & 0.1223  \\
\href{http://www.gutenberg.org/ebooks/242}{242} & My Antonia & 847 & 0.1024  \\
\href{http://www.gutenberg.org/ebooks/146}{146} & A Little Princess
: Being the who... & 825 & 0.1189  \\
\href{http://www.gutenberg.org/ebooks/47}{47} & Anne of Avonlea & 803 & 0.1000  \\
\end{longtable}
\begin{longtable}{l | l | l | l}
\caption{Stories which are are represented by core emotional arc 8, sorted by the variance explained in their emotional arc by core emotional arc 8.}\\
ID & Title & DL's & max($W$[i,:])\\
\hline
\endhead
\href{http://www.gutenberg.org/ebooks/19360}{19360} & Six to Sixteen: A Story for Girls & 47 & 0.2415 \\
\href{http://www.gutenberg.org/ebooks/28849}{28849} & Smugglers' Reef: A Rick Brant Sc... & 46 & 0.2320 \\
\href{http://www.gutenberg.org/ebooks/18581}{18581} & Adrift in New York: Tom and Flor... & 111 & 0.2182 \\
\href{http://www.gutenberg.org/ebooks/5141}{5141} & What Katy Did at School & 85 & 0.2065 \\
\href{http://www.gutenberg.org/ebooks/4217}{4217} & A Portrait of the Artist as a Yo... & 2,172 & 0.2039 \\
\href{http://www.gutenberg.org/ebooks/31308}{31308} & Orientations & 76 & 0.1924 \\
\href{http://www.gutenberg.org/ebooks/35117}{35117} & Lord Tony's Wife: An Adventure o... & 66 & 0.1748 \\
\href{http://www.gutenberg.org/ebooks/4715}{4715} & An African Millionaire: Episodes... & 163 & 0.1714 \\
\end{longtable}
\begin{longtable}{l | l | l | l}
\caption{Top 10 stories which are are represented by core emotional arc 8, sorted by downloads.}\\
ID & Title & DL's & max($W$[i,:])\\
\hline
\endhead
\href{http://www.gutenberg.org/ebooks/4217}{4217} & A Portrait of the Artist as a Yo... & 2,172 & 0.2039 \\
\href{http://www.gutenberg.org/ebooks/204}{204} & The Innocence of Father Brown & 800 & 0.1357 \\
\href{http://www.gutenberg.org/ebooks/26}{26} & Paradise Lost & 730 & 0.1565 \\
\href{http://www.gutenberg.org/ebooks/64}{64} & The Gods of Mars & 628 & 0.1481 \\
\href{http://www.gutenberg.org/ebooks/5670}{5670} & Jacob's Room & 403 & 0.1013 \\
\href{http://www.gutenberg.org/ebooks/4352}{4352} & Laughter: An Essay on the Meanin... & 365 & 0.1251 \\
\href{http://www.gutenberg.org/ebooks/225}{225} & At the Back of the North Wind & 288 & 0.0889 \\
\href{http://www.gutenberg.org/ebooks/517}{517} & The Emerald City of Oz & 266 & 0.1097 \\
\href{http://www.gutenberg.org/ebooks/2688}{2688} & The Clue of the Twisted Candle & 237 & 0.1402 \\
\end{longtable}
\begin{longtable}{l | l | l | l}
\caption{Stories which are are represented by core emotional arc 9, sorted by the variance explained in their emotional arc by core emotional arc 9.}\\
ID & Title & DL's & max($W$[i,:])\\
\hline
\endhead
\href{http://www.gutenberg.org/ebooks/19066}{19066} & Brigands of the Moon & 65 & 0.2113\\
\href{http://www.gutenberg.org/ebooks/17959}{17959} & The Hand Of Fu-Manchu: Being a Ne... & 112 & 0.1861\\
\href{http://www.gutenberg.org/ebooks/19145}{19145} & The Time Traders & 225 & 0.1801\\
\href{http://www.gutenberg.org/ebooks/2057}{2057} & The Last of the Plainsmen & 106 & 0.1681\\
\href{http://www.gutenberg.org/ebooks/1188}{1188} & The Lair of the White Worm & 206 & 0.1632\\
\href{http://www.gutenberg.org/ebooks/546}{546} & Under the Andes & 87 & 0.1544\\
\href{http://www.gutenberg.org/ebooks/18668}{18668} & In Search of the Unknown & 159 & 0.1489\\
\href{http://www.gutenberg.org/ebooks/6955}{6955} & The Prince and Betty & 85 & 0.1488\\
\end{longtable}
\begin{longtable}{l | l | l | l}
\caption{Top 10 stories which are are represented by core emotional arc 9, sorted by downloads.}\\
ID & Title & DL's & max(W[i,:])\\
\hline
\endhead
\href{http://www.gutenberg.org/ebooks/11}{11} & Alice's Adventures in Wonderland & 17,366 & 0.0988  \\
\href{http://www.gutenberg.org/ebooks/33}{33} & The Scarlet Letter & 3,045 & 0.1195  \\
\href{http://www.gutenberg.org/ebooks/28885}{28885} & Alice's Adventures in Wonderland... & 1,051 & 0.0862  \\
\href{http://www.gutenberg.org/ebooks/271}{271} & Black Beauty & 780 & 0.0870  \\
\href{http://www.gutenberg.org/ebooks/2775}{2775} & The Good Soldier & 426 & 0.1144  \\
\href{http://www.gutenberg.org/ebooks/1872}{1872} & The Red House Mystery & 422 & 0.1421  \\
\href{http://www.gutenberg.org/ebooks/25344}{25344} & The Scarlet Letter & 386 & 0.1266  \\
\href{http://www.gutenberg.org/ebooks/1051}{1051} & Sartor Resartus: The Life and Op... & 347 & 0.1074  \\
\href{http://www.gutenberg.org/ebooks/3328}{3328} & Man and Superman: A Comedy and a... & 312 & 0.1181  \\
\end{longtable}
\begin{longtable}{l | l | l | l}
\caption{Stories which are are represented by core emotional arc 10, sorted by the variance explained in their emotional arc by core emotional arc 10.}\\
ID & Title & DL's & max($W$[i,:])\\
\hline
\endhead
\href{http://www.gutenberg.org/ebooks/21374}{21374} & !Tention: A Story of Boy-Life du... & 92 & 0.2195 \\
\href{http://www.gutenberg.org/ebooks/24499}{24499} & The Green Carnation & 65 & 0.1986 \\
\href{http://www.gutenberg.org/ebooks/544}{544} & Anne's House of Dreams & 586 & 0.1676 \\
\href{http://www.gutenberg.org/ebooks/4922}{4922} & Bar-20 Days & 57 & 0.1648 \\
\href{http://www.gutenberg.org/ebooks/8681}{8681} & The Face and the Mask & 82 & 0.1528 \\
\href{http://www.gutenberg.org/ebooks/1805}{1805} & The Gentle Grafter & 84 & 0.1505 \\
\href{http://www.gutenberg.org/ebooks/10443}{10443} & The Rayner-Slade Amalgamation & 61 & 0.1486 \\
\href{http://www.gutenberg.org/ebooks/16339}{16339} & The Passenger from Calais & 125 & 0.1454 \\
\end{longtable}
\begin{longtable}{l | l | l | l}
\caption{Top 10 stories which are are represented by core emotional arc 10, sorted by downloads.}\\
ID & Title & DL's & max($W$[i,:])\\
\hline
\endhead
\href{http://www.gutenberg.org/ebooks/25305}{25305} & Memoirs Of Fanny Hill: A New and ... & 2,222 & 0.1296 \\
\href{http://www.gutenberg.org/ebooks/544}{544} & Anne's House of Dreams & 586 & 0.1676 \\
\href{http://www.gutenberg.org/ebooks/4540}{4540} & In His Steps & 230 & 0.0902 \\
\href{http://www.gutenberg.org/ebooks/38070}{38070} & The Norwegian Fairy Book & 149 & 0.1044 \\
\href{http://www.gutenberg.org/ebooks/16339}{16339} & The Passenger from Calais & 125 & 0.1454 \\
\href{http://www.gutenberg.org/ebooks/1535}{1535} & The Tragedy of Coriolanus & 116 & 0.1137 \\
\href{http://www.gutenberg.org/ebooks/35517}{35517} & The Three Impostors; or, The Tra... & 111 & 0.1148 \\
\href{http://www.gutenberg.org/ebooks/21374}{21374} & !Tention: A Story of Boy-Life du... & 92 & 0.2195 \\
\href{http://www.gutenberg.org/ebooks/24775}{24775} & Up the River; or, Yachting on th... & 88 & 0.1449 \\
\end{longtable}
\begin{longtable}{l | l | l | l}
\caption{Stories which are are represented by core emotional arc 11, sorted by the variance explained in their emotional arc by core emotional arc 11.}\\
ID & Title & DL's & max($W$[i,:])\\
\hline
\endhead
\href{http://www.gutenberg.org/ebooks/5343}{5343} & Rainbow Valley & 257 & 0.1533\\
\href{http://www.gutenberg.org/ebooks/1987}{1987} & The Outlet & 40 & 0.1444\\
\href{http://www.gutenberg.org/ebooks/13882}{13882} & John Thorndyke's Cases
: related ... & 217 & 0.1343\\
\href{http://www.gutenberg.org/ebooks/1167}{1167} & A Strange Disappearance & 270 & 0.1282\\
\href{http://www.gutenberg.org/ebooks/7477}{7477} & The Book of Wonder & 244 & 0.1276\\
\href{http://www.gutenberg.org/ebooks/6753}{6753} & Psmith in the City & 199 & 0.1245\\
\href{http://www.gutenberg.org/ebooks/5265}{5265} & The Ball and the Cross & 118 & 0.1229\\
\href{http://www.gutenberg.org/ebooks/974}{974} & The Secret Agent: A Simple Tale & 1,142 & 0.1224\\
\href{http://www.gutenberg.org/ebooks/27525}{27525} & Bones in London & 85 & 0.1157\\
\href{http://www.gutenberg.org/ebooks/24450}{24450} & Bones: Being Further Adventures i... & 121 & 0.1072\\
\href{http://www.gutenberg.org/ebooks/5758}{5758} & Many Cargoes & 82 & 0.1048\\
\href{http://www.gutenberg.org/ebooks/4090}{4090} & From Ritual to Romance & 90 & 0.1037\\
\href{http://www.gutenberg.org/ebooks/27198}{27198} & The Explorer & 62 & 0.1035\\
\href{http://www.gutenberg.org/ebooks/10832}{10832} & Carnacki, the Ghost Finder & 135 & 0.1032\\
\href{http://www.gutenberg.org/ebooks/13372}{13372} & The Gloved Hand & 70 & 0.0942\\
\href{http://www.gutenberg.org/ebooks/767}{767} & Agnes Grey & 287 & 0.0856\\
\href{http://www.gutenberg.org/ebooks/3055}{3055} & The Wood Beyond the World & 106 & 0.0799\\
\href{http://www.gutenberg.org/ebooks/2641}{2641} & A Room with a View & 1,354 & 0.0681\\
\end{longtable}
\begin{longtable}{l | l | l | l}
\caption{Top 10 stories which are are represented by core emotional arc 11, sorted by downloads.}\\
ID & Title & DL's & max(W[i,:])\\
\hline
\endhead
\href{http://www.gutenberg.org/ebooks/2641}{2641} & A Room with a View & 1,354 & 0.0681  \\
\href{http://www.gutenberg.org/ebooks/974}{974} & The Secret Agent: A Simple Tale & 1,142 & 0.1224  \\
\href{http://www.gutenberg.org/ebooks/767}{767} & Agnes Grey & 287 & 0.0856  \\
\href{http://www.gutenberg.org/ebooks/1167}{1167} & A Strange Disappearance & 270 & 0.1282  \\
\href{http://www.gutenberg.org/ebooks/5343}{5343} & Rainbow Valley & 257 & 0.1533  \\
\href{http://www.gutenberg.org/ebooks/7477}{7477} & The Book of Wonder & 244 & 0.1276  \\
\href{http://www.gutenberg.org/ebooks/13882}{13882} & John Thorndyke's Cases
: related ... & 217 & 0.1343  \\
\href{http://www.gutenberg.org/ebooks/6753}{6753} & Psmith in the City & 199 & 0.1245  \\
\href{http://www.gutenberg.org/ebooks/10832}{10832} & Carnacki, the Ghost Finder & 135 & 0.1032  \\
\end{longtable}
\begin{longtable}{l | l | l | l}
\caption{Stories which are are represented by core emotional arc 12, sorted by the variance explained in their emotional arc by core emotional arc 12.}\\
ID & Title & DL's & max($W$[i,:])\\
\hline
\endhead
\href{http://www.gutenberg.org/ebooks/318}{318} & John Barleycorn & 130 & 0.1909 \\
\href{http://www.gutenberg.org/ebooks/1595}{1595} & Whirligigs & 94 & 0.1563 \\
\href{http://www.gutenberg.org/ebooks/25728}{25728} & Desert Conquest; or, Precious Wa... & 58 & 0.1543 \\
\href{http://www.gutenberg.org/ebooks/557}{557} & Puck of Pook's Hill & 50 & 0.1482 \\
\href{http://www.gutenberg.org/ebooks/15976}{15976} & Puck of Pook's Hill & 40 & 0.1376 \\
\href{http://www.gutenberg.org/ebooks/26027}{26027} & Puck of Pook's Hill & 78 & 0.1357 \\
\href{http://www.gutenberg.org/ebooks/3326}{3326} & The Well-Beloved: A Sketch of a ... & 76 & 0.1226 \\
\href{http://www.gutenberg.org/ebooks/1611}{1611} & Seventeen
: A Tale of Youth and S... & 56 & 0.1145 \\
\href{http://www.gutenberg.org/ebooks/12803}{12803} & Headlong Hall & 124 & 0.1113 \\
\href{http://www.gutenberg.org/ebooks/610}{610} & Idylls of the King & 494 & 0.1075 \\
\href{http://www.gutenberg.org/ebooks/624}{624} & Looking Backward, 2000 to 1887 & 679 & 0.1060 \\
\href{http://www.gutenberg.org/ebooks/37532}{37532} & The Scottish Fairy Book & 98 & 0.1036 \\
\href{http://www.gutenberg.org/ebooks/10067}{10067} & The Mystery of the Boule Cabinet... & 54 & 0.0938 \\
\href{http://www.gutenberg.org/ebooks/316}{316} & The Golden Road & 151 & 0.0864 \\
\href{http://www.gutenberg.org/ebooks/6840}{6840} & Queen Lucia & 81 & 0.0801 \\
\end{longtable}
\begin{longtable}{l | l | l | l}
\caption{Top 10 stories which are are represented by core emotional arc 12, sorted by downloads.}\\
ID & Title & DL's & max($W$[i,:])\\
\hline
\endhead
\href{http://www.gutenberg.org/ebooks/624}{624} & Looking Backward, 2000 to 1887 & 679 & 0.1060 \\
\href{http://www.gutenberg.org/ebooks/610}{610} & Idylls of the King & 494 & 0.1075 \\
\href{http://www.gutenberg.org/ebooks/316}{316} & The Golden Road & 151 & 0.0864 \\
\href{http://www.gutenberg.org/ebooks/318}{318} & John Barleycorn & 130 & 0.1909 \\
\href{http://www.gutenberg.org/ebooks/12803}{12803} & Headlong Hall & 124 & 0.1113 \\
\href{http://www.gutenberg.org/ebooks/37532}{37532} & The Scottish Fairy Book & 98 & 0.1036 \\
\href{http://www.gutenberg.org/ebooks/1595}{1595} & Whirligigs & 94 & 0.1563 \\
\href{http://www.gutenberg.org/ebooks/6840}{6840} & Queen Lucia & 81 & 0.0801 \\
\href{http://www.gutenberg.org/ebooks/26027}{26027} & Puck of Pook's Hill & 78 & 0.1357 \\
\end{longtable}
\begin{longtable}{l | l | l | l}
\caption{Stories which are are represented by core emotional arc 13, sorted by the variance explained in their emotional arc by core emotional arc 13.}\\
ID & Title & DL's & max($W$[i,:])\\
\hline
\endhead
\href{http://www.gutenberg.org/ebooks/3815}{3815} & Rolling Stones & 107 & 0.1377\\
\href{http://www.gutenberg.org/ebooks/26732}{26732} & Free Air & 42 & 0.1295\\
\href{http://www.gutenberg.org/ebooks/30836}{30836} & Seven Keys to Baldpate & 57 & 0.1245\\
\href{http://www.gutenberg.org/ebooks/22342}{22342} & Supermind & 61 & 0.1185\\
\href{http://www.gutenberg.org/ebooks/9925}{9925} & Black Jack & 111 & 0.1174\\
\href{http://www.gutenberg.org/ebooks/21687}{21687} & The Youngest Girl in the Fifth: ... & 59 & 0.1097\\
\href{http://www.gutenberg.org/ebooks/19527}{19527} & The Yukon Trail: A Tale of the N... & 44 & 0.1085\\
\href{http://www.gutenberg.org/ebooks/6312}{6312} & Representative Men: Seven Lectures & 170 & 0.0986\\
\href{http://www.gutenberg.org/ebooks/33391}{33391} & Bill Nye's Cordwood & 248 & 0.0888\\
\end{longtable}
\begin{longtable}{l | l | l | l}
\caption{Top 10 stories which are are represented by core emotional arc 13, sorted by downloads.}\\
ID & Title & DL's & max(W[i,:])\\
\hline
\endhead
\href{http://www.gutenberg.org/ebooks/33391}{33391} & Bill Nye's Cordwood & 248 & 0.0888  \\
\href{http://www.gutenberg.org/ebooks/6312}{6312} & Representative Men: Seven Lectures & 170 & 0.0986  \\
\href{http://www.gutenberg.org/ebooks/9925}{9925} & Black Jack & 111 & 0.1174  \\
\href{http://www.gutenberg.org/ebooks/3815}{3815} & Rolling Stones & 107 & 0.1377  \\
\href{http://www.gutenberg.org/ebooks/22342}{22342} & Supermind & 61 & 0.1185  \\
\href{http://www.gutenberg.org/ebooks/21687}{21687} & The Youngest Girl in the Fifth: ... & 59 & 0.1097  \\
\href{http://www.gutenberg.org/ebooks/30836}{30836} & Seven Keys to Baldpate & 57 & 0.1245  \\
\href{http://www.gutenberg.org/ebooks/19527}{19527} & The Yukon Trail: A Tale of the N... & 44 & 0.1085  \\
\href{http://www.gutenberg.org/ebooks/26732}{26732} & Free Air & 42 & 0.1295  \\
\end{longtable}
\begin{longtable}{l | l | l | l}
\caption{Stories which are are represented by core emotional arc 14, sorted by the variance explained in their emotional arc by core emotional arc 14.}\\
ID & Title & DL's & max($W$[i,:])\\
\hline
\endhead
\href{http://www.gutenberg.org/ebooks/5077}{5077} & Marmion: A Tale of Flodden Field... & 125 & 0.1476 \\
\href{http://www.gutenberg.org/ebooks/20869}{20869} & The Skylark of Space & 246 & 0.1457 \\
\href{http://www.gutenberg.org/ebooks/15272}{15272} & Spenser's The Faerie Queene, Book I & 978 & 0.1302 \\
\href{http://www.gutenberg.org/ebooks/1027}{1027} & The Lone Star Ranger: A Romance ... & 253 & 0.1277 \\
\href{http://www.gutenberg.org/ebooks/2233}{2233} & A Damsel in Distress & 201 & 0.1256 \\
\href{http://www.gutenberg.org/ebooks/27690}{27690} & Nobody's Girl: (En Famille) & 70 & 0.1219 \\
\href{http://www.gutenberg.org/ebooks/32759}{32759} & Red Nails & 151 & 0.1158 \\
\href{http://www.gutenberg.org/ebooks/27063}{27063} & The Hero & 60 & 0.1155 \\
\href{http://www.gutenberg.org/ebooks/2804}{2804} & Rose in Bloom
: A Sequel to "Eigh... & 168 & 0.1083 \\
\href{http://www.gutenberg.org/ebooks/45658}{45658} & The Mystery of the Downs & 48 & 0.1030 \\
\href{http://www.gutenberg.org/ebooks/105}{105} & Persuasion & 2,535 & 0.0980 \\
\href{http://www.gutenberg.org/ebooks/619}{619} & The Warden & 215 & 0.0946 \\
\href{http://www.gutenberg.org/ebooks/34732}{34732} & Max Carrados & 92 & 0.0941 \\
\href{http://www.gutenberg.org/ebooks/291}{291} & The Golden Age & 42 & 0.0854 \\
\href{http://www.gutenberg.org/ebooks/15673}{15673} & The Day of the Beast & 98 & 0.0792 \\
\href{http://www.gutenberg.org/ebooks/1671}{1671} & When a Man Marries & 46 & 0.0651 \\
\end{longtable}
\begin{longtable}{l | l | l | l}
\caption{Top 10 stories which are are represented by core emotional arc 14, sorted by downloads.}\\
ID & Title & DL's & max($W$[i,:])\\
\hline
\endhead
\href{http://www.gutenberg.org/ebooks/105}{105} & Persuasion & 2,535 & 0.0980 \\
\href{http://www.gutenberg.org/ebooks/15272}{15272} & Spenser's The Faerie Queene, Book I & 978 & 0.1302 \\
\href{http://www.gutenberg.org/ebooks/1027}{1027} & The Lone Star Ranger: A Romance ... & 253 & 0.1277 \\
\href{http://www.gutenberg.org/ebooks/20869}{20869} & The Skylark of Space & 246 & 0.1457 \\
\href{http://www.gutenberg.org/ebooks/619}{619} & The Warden & 215 & 0.0946 \\
\href{http://www.gutenberg.org/ebooks/2233}{2233} & A Damsel in Distress & 201 & 0.1256 \\
\href{http://www.gutenberg.org/ebooks/2804}{2804} & Rose in Bloom
: A Sequel to "Eigh... & 168 & 0.1083 \\
\href{http://www.gutenberg.org/ebooks/32759}{32759} & Red Nails & 151 & 0.1158 \\
\href{http://www.gutenberg.org/ebooks/5077}{5077} & Marmion: A Tale of Flodden Field... & 125 & 0.1476 \\
\end{longtable}
\begin{longtable}{l | l | l | l}
\caption{Stories which are are represented by core emotional arc 15, sorted by the variance explained in their emotional arc by core emotional arc 15.}\\
ID & Title & DL's & max($W$[i,:])\\
\hline
\endhead
\href{http://www.gutenberg.org/ebooks/33066}{33066} & The Garden of Eden & 72 & 0.1165\\
\href{http://www.gutenberg.org/ebooks/40852}{40852} & Instigations: Together with An Es... & 79 & 0.0806\\
\end{longtable}
\begin{longtable}{l | l | l | l}
\caption{Top 10 stories which are are represented by core emotional arc 15, sorted by downloads.}\\
ID & Title & DL's & max(W[i,:])\\
\hline
\endhead
\href{http://www.gutenberg.org/ebooks/40852}{40852} & Instigations: Together with An Es... & 79 & 0.0806  \\
\href{http://www.gutenberg.org/ebooks/33066}{33066} & The Garden of Eden & 72 & 0.1165  \\
\end{longtable}
\begin{longtable}{l | l | l | l}
\caption{Stories which are are represented by core emotional arc 16, sorted by the variance explained in their emotional arc by core emotional arc 16.}\\
ID & Title & DL's & max($W$[i,:])\\
\hline
\endhead
\href{http://www.gutenberg.org/ebooks/872}{872} & Reprinted Pieces & 49 & 0.1103 \\
\href{http://www.gutenberg.org/ebooks/8673}{8673} & A Columbus of Space & 56 & 0.1077 \\
\href{http://www.gutenberg.org/ebooks/434}{434} & The Circular Staircase & 189 & 0.0978 \\
\href{http://www.gutenberg.org/ebooks/1263}{1263} & The Glimpses of the Moon & 58 & 0.0864 \\
\href{http://www.gutenberg.org/ebooks/3075}{3075} & The Return & 73 & 0.0759 \\
\end{longtable}
\begin{longtable}{l | l | l | l}
\caption{Top 10 stories which are are represented by core emotional arc 16, sorted by downloads.}\\
ID & Title & DL's & max($W$[i,:])\\
\hline
\endhead
\href{http://www.gutenberg.org/ebooks/434}{434} & The Circular Staircase & 189 & 0.0978 \\
\href{http://www.gutenberg.org/ebooks/3075}{3075} & The Return & 73 & 0.0759 \\
\href{http://www.gutenberg.org/ebooks/1263}{1263} & The Glimpses of the Moon & 58 & 0.0864 \\
\href{http://www.gutenberg.org/ebooks/8673}{8673} & A Columbus of Space & 56 & 0.1077 \\
\href{http://www.gutenberg.org/ebooks/872}{872} & Reprinted Pieces & 49 & 0.1103 \\
\end{longtable}
\begin{longtable}{l | l | l | l}
\caption{Stories which are are represented by core emotional arc 18, sorted by the variance explained in their emotional arc by core emotional arc 18.}\\
ID & Title & DL's & max($W$[i,:])\\
\hline
\endhead
\href{http://www.gutenberg.org/ebooks/5776}{5776} & 100\%: the Story of a Patriot & 64 & 0.1731 \\
\href{http://www.gutenberg.org/ebooks/3188}{3188} & Mark Twain's Speeches & 500 & 0.0778 \\
\end{longtable}
\begin{longtable}{l | l | l | l}
\caption{Top 10 stories which are are represented by core emotional arc 18, sorted by downloads.}\\
ID & Title & DL's & max($W$[i,:])\\
\hline
\endhead
\href{http://www.gutenberg.org/ebooks/3188}{3188} & Mark Twain's Speeches & 500 & 0.0778 \\
\href{http://www.gutenberg.org/ebooks/5776}{5776} & 100\%: the Story of a Patriot & 64 & 0.1731 \\
\end{longtable}
\begin{longtable}{l | l | l | l}
\caption{Stories which are are represented by core emotional arc 20, sorted by the variance explained in their emotional arc by core emotional arc 20.}\\
ID & Title & DL's & max($W$[i,:])\\
\hline
\endhead
\href{http://www.gutenberg.org/ebooks/15119}{15119} & Handy Dictionary of Poetical Quo... & 46 & 0.0728 \\
\end{longtable}
\begin{longtable}{l | l | l | l}
\caption{Top 10 stories which are are represented by core emotional arc 20, sorted by downloads.}\\
ID & Title & DL's & max($W$[i,:])\\
\hline
\endhead
\href{http://www.gutenberg.org/ebooks/15119}{15119} & Handy Dictionary of Poetical Quo... & 46 & 0.0728 \\
\end{longtable}

\clearpage
\pagebreak
\section{Additional Hierarchical Clustering Figures}
\label{sec:clustering-supp}

In the section, we include additional results from the hierarchical clustering analysis.

\begin{figure*}[!htb]
  \centering
  \includegraphics[width=0.98\textwidth]{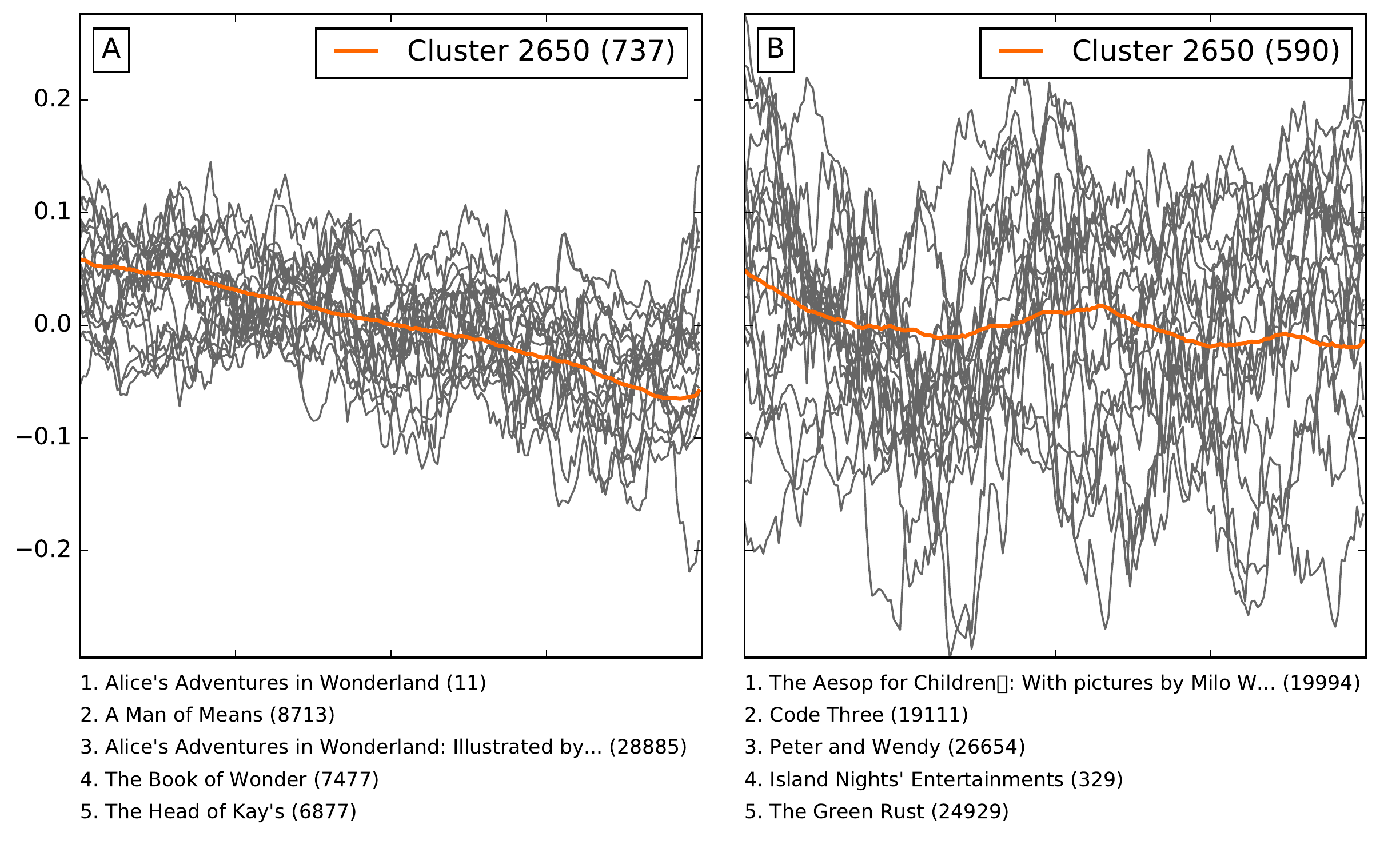}
  \caption[]{
    The 2 clusters identified by Agglomerative Clustering using Ward's method.
  }
  \label{fig:ward-cluster-2}
\end{figure*}

\begin{figure*}[!htb]
  \centering
  \includegraphics[width=0.98\textwidth]{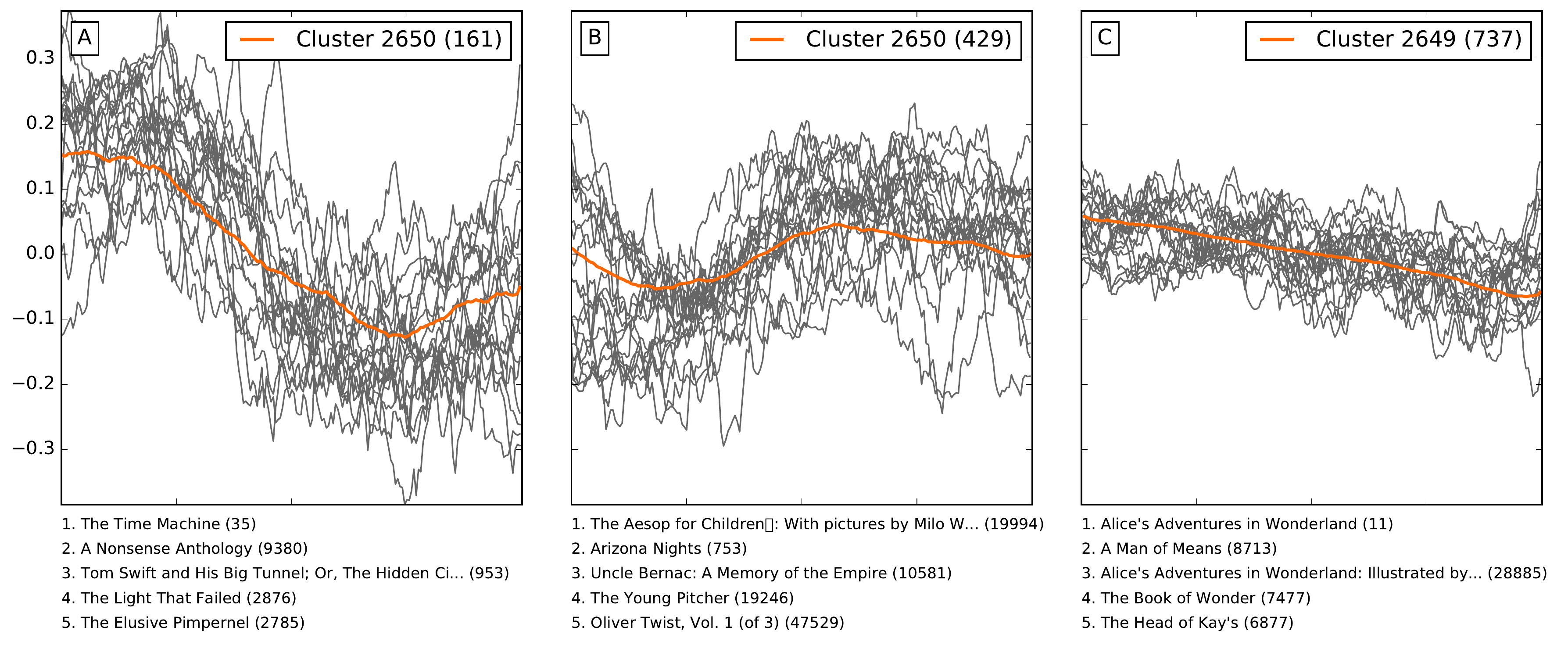}
  \caption[]{
    The 3 clusters identified by Agglomerative Clustering using Ward's method.
  }
  \label{fig:ward-cluster-3}
\end{figure*}

\begin{figure*}[!htb]
  \centering
  \includegraphics[width=0.98\textwidth]{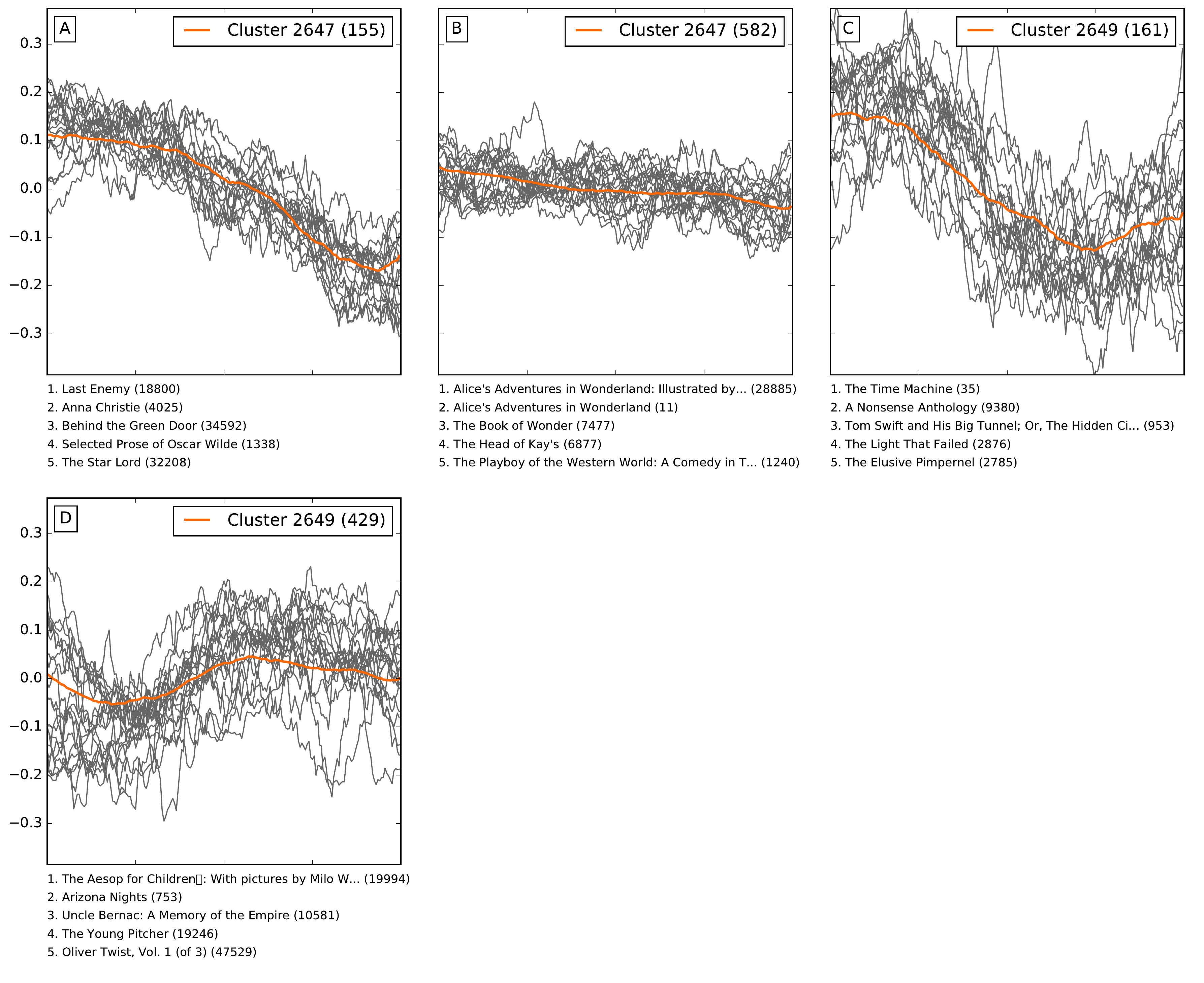}
  \caption[]{
    The 4 clusters identified by Agglomerative Clustering using Ward's method.
  }
    \label{fig:ward-cluster-4}
\end{figure*}

\begin{figure*}[!htb]
  \centering
  \includegraphics[width=0.98\textwidth]{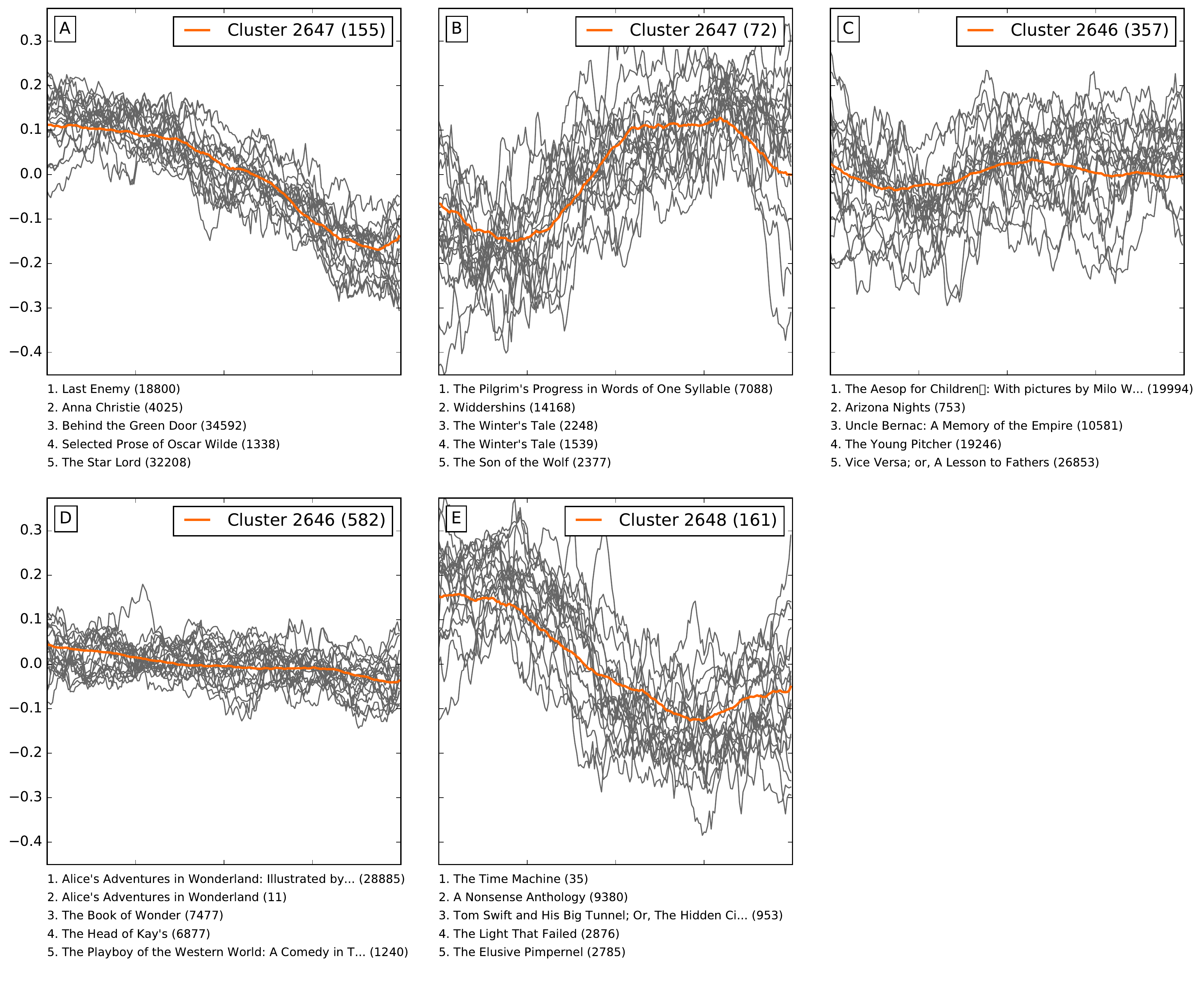}
  \caption[]{
    The 5 clusters identified by Agglomerative Clustering using Ward's method.
  }
    \label{fig:ward-cluster-5}
\end{figure*}

\begin{figure*}[!htb]
  \centering
  \includegraphics[width=0.98\textwidth]{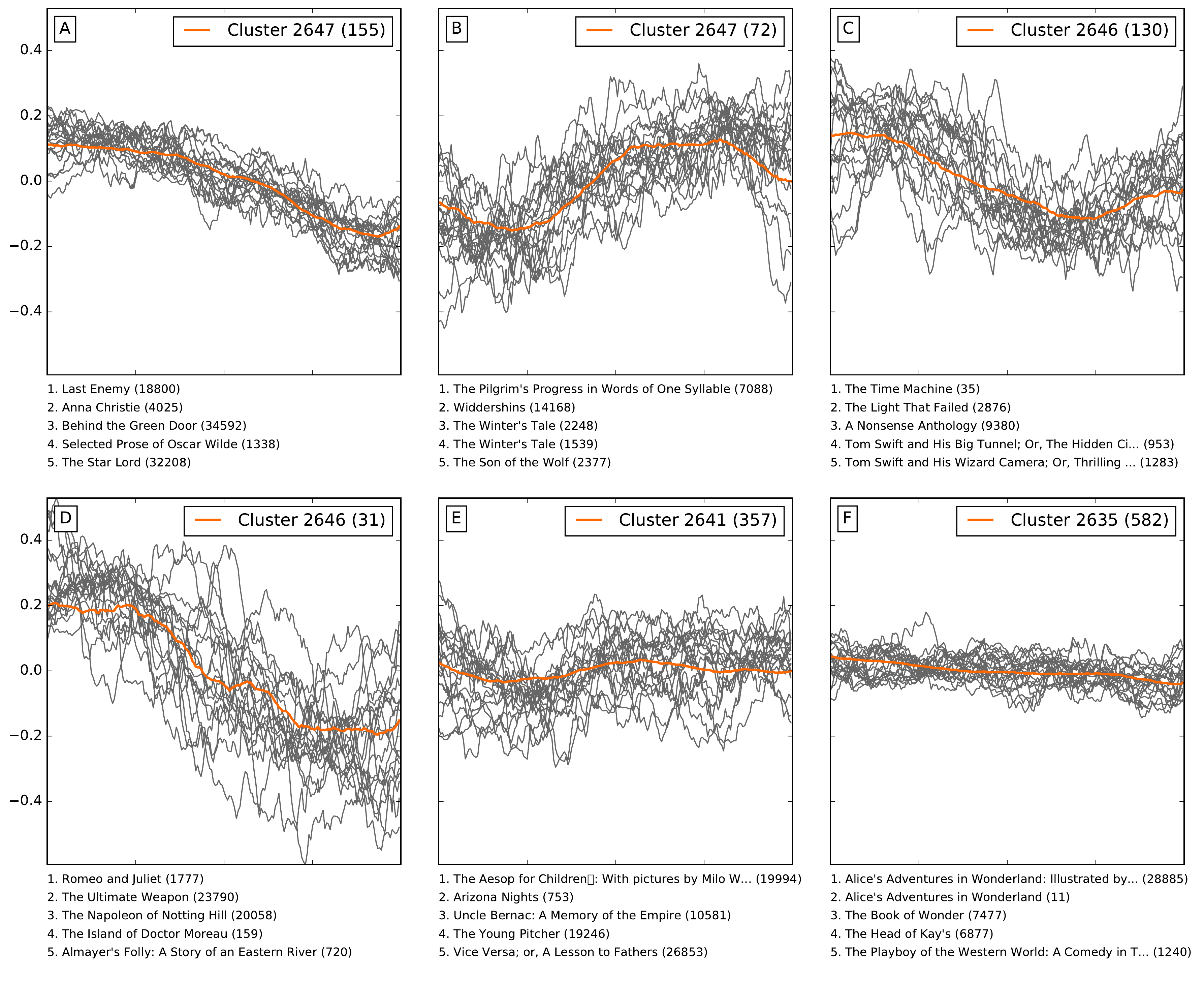}
  \caption[]{
    The 6 clusters identified by Agglomerative Clustering using Ward's method.
  }
    \label{fig:ward-cluster-6}
\end{figure*}

\begin{figure*}[!htb]
  \centering
  \includegraphics[width=0.98\textwidth]{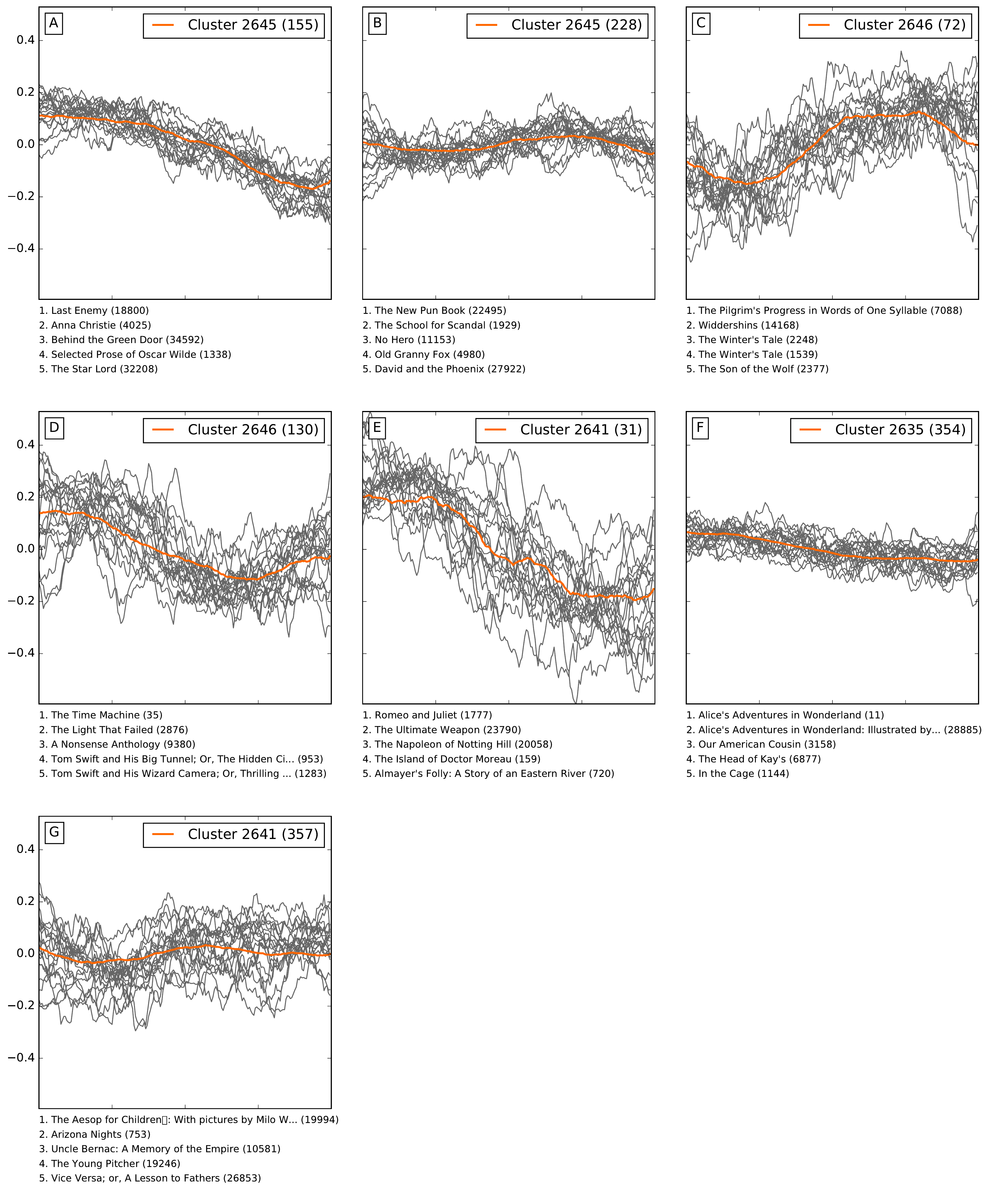}
  \caption[]{
    The 7 clusters identified by Agglomerative Clustering using Ward's method.
  }
    \label{fig:ward-cluster-7}
\end{figure*}

\begin{figure*}[!htb]
  \centering
  \includegraphics[width=0.98\textwidth]{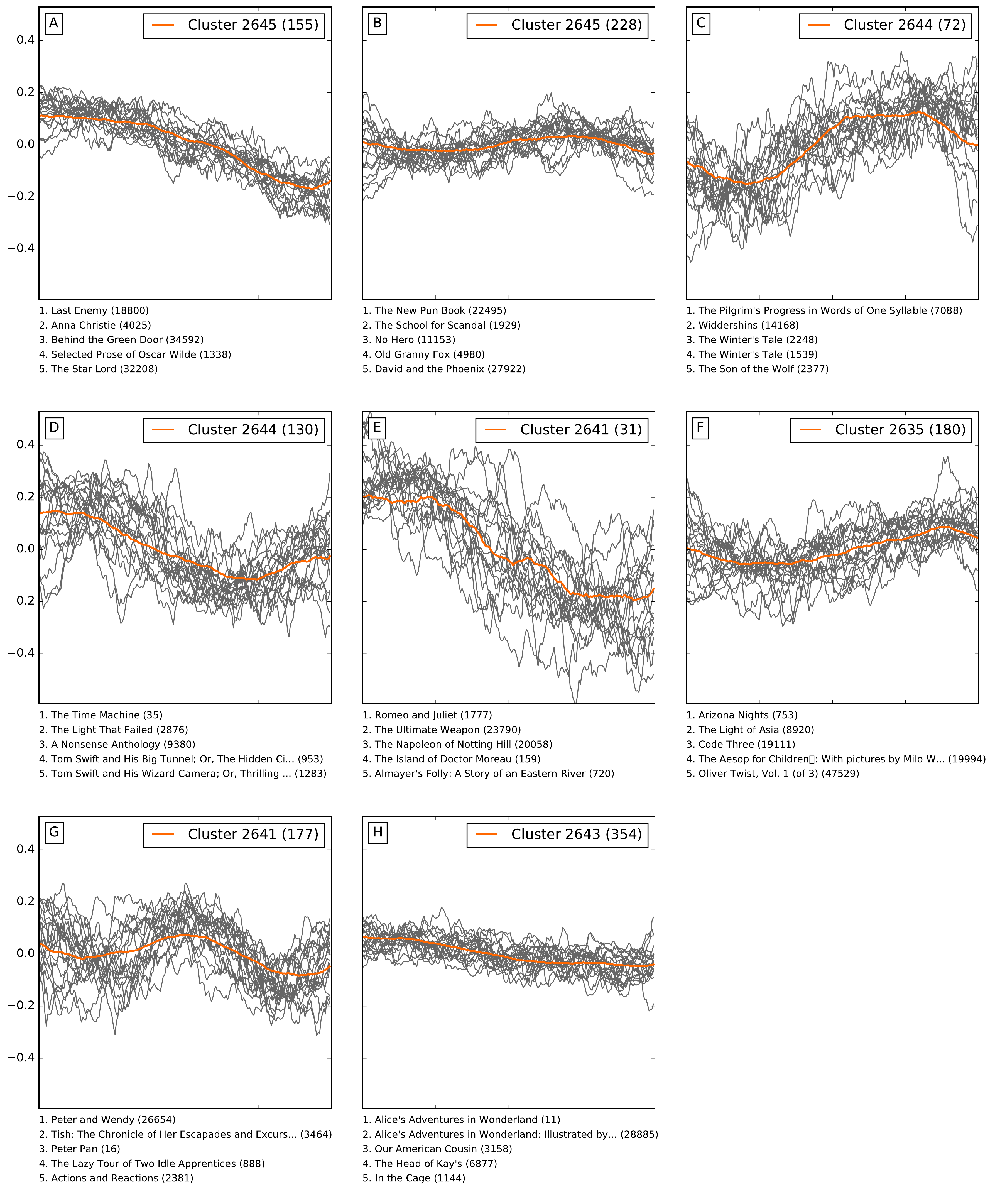}
  \caption[]{
    The 8 clusters identified by Agglomerative Clustering using Ward's method.
  }
    \label{fig:ward-cluster-8}
\end{figure*}

\begin{figure*}[!htb]
  \centering
  \includegraphics[width=0.98\textwidth]{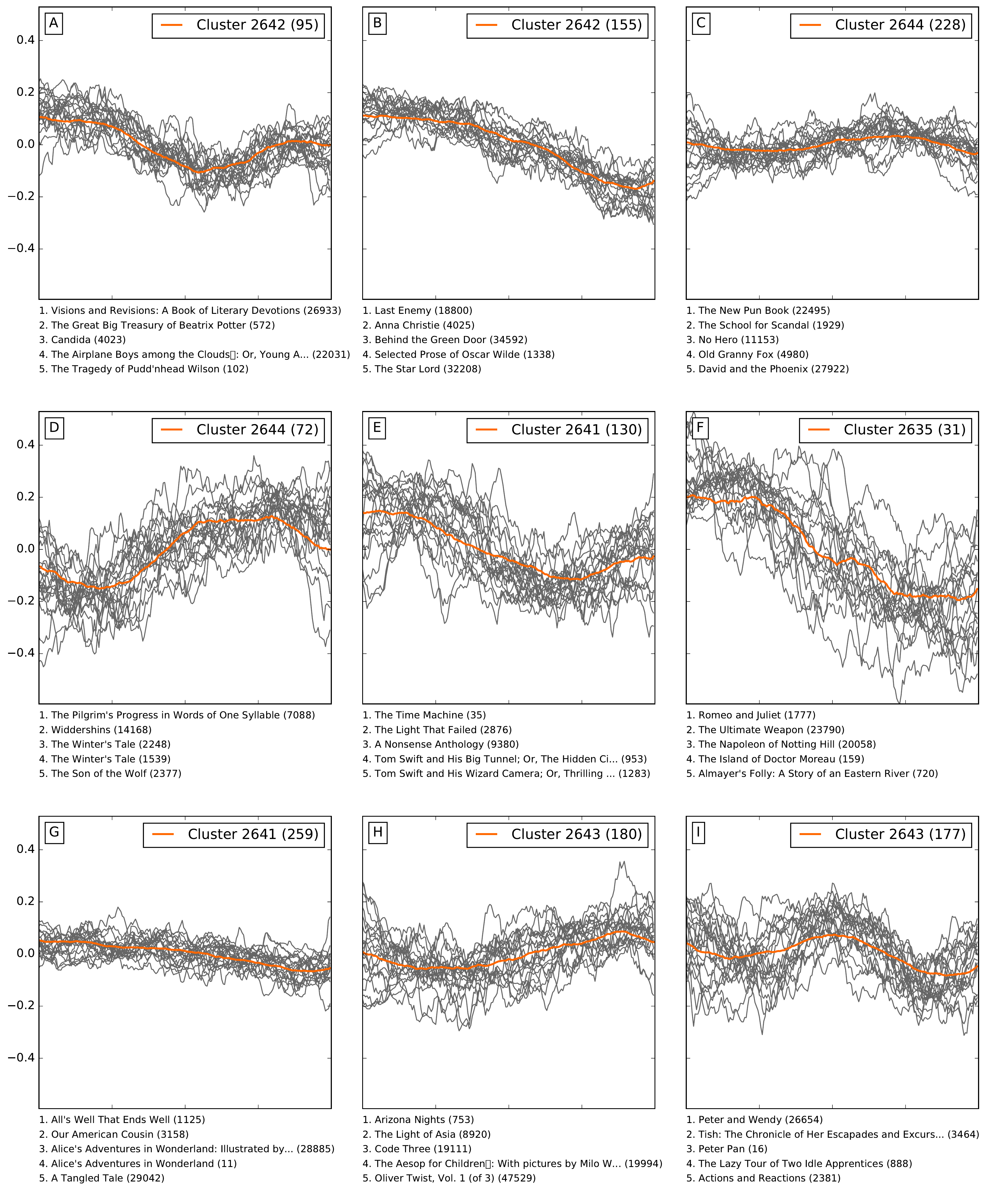}
  \caption[]{
    The 9 clusters identified by Agglomerative Clustering using Ward's method.
  }
    \label{fig:ward-cluster-9}
\end{figure*}

\begin{figure*}[!htb]
  \centering
  \includegraphics[width=0.9\textwidth]{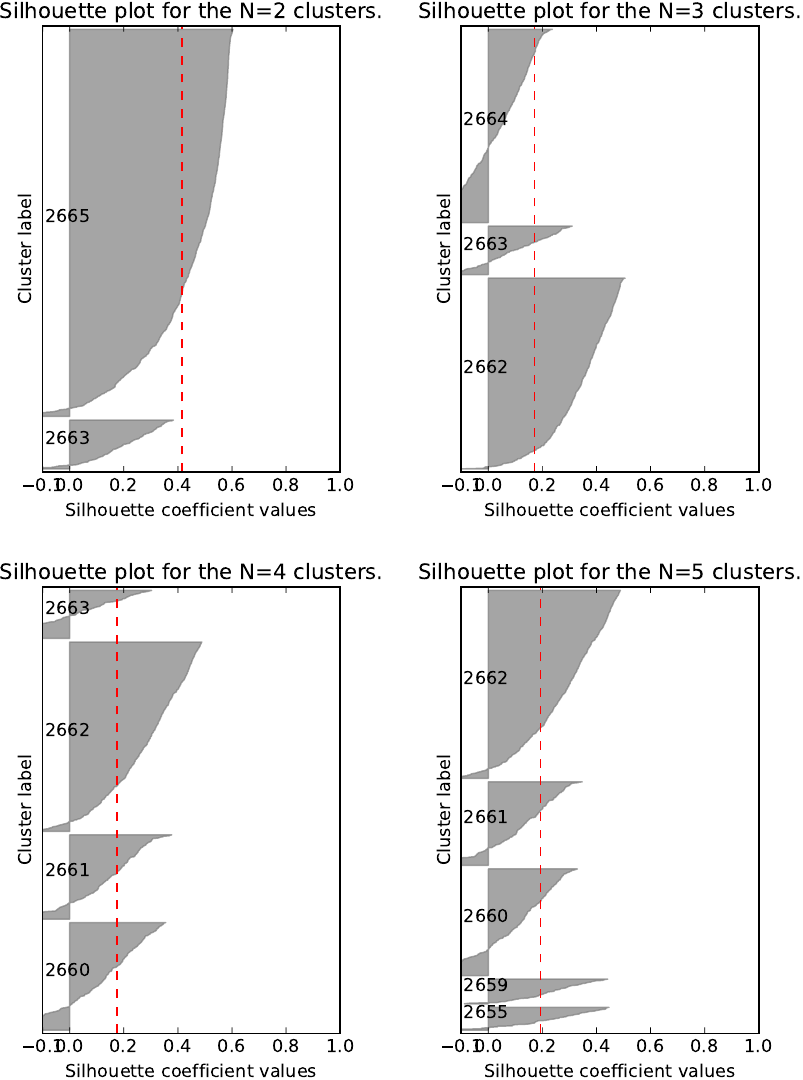}
  \caption[]{
    The silhouette plots for 2--5 clusters identified by Agglomerative Clustering using Ward's method.
  }
  \label{fig:clustering-2-5-clusters}
\end{figure*}

\begin{figure*}[!htb]
  \centering
  \includegraphics[width=0.9\textwidth]{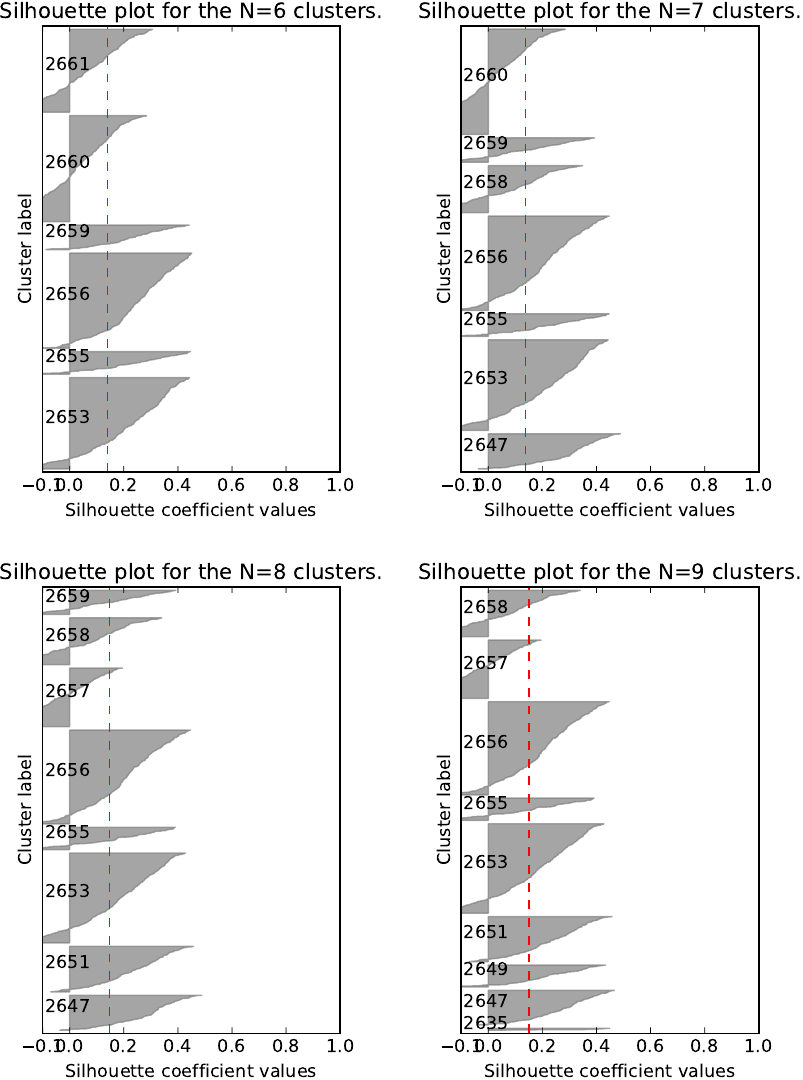}
  \caption[]{
    The silhouette plots for 6--9 clusters identified by Agglomerative Clustering using Ward's method.
  }
  \label{fig:clustering-6-9-clusters}
\end{figure*}

\clearpage
\pagebreak
\section{Additional SOM Figures}
\label{sec:SOM-supp}

\begin{figure}[ht]
  \centering
  \includegraphics[width=0.78\textwidth]{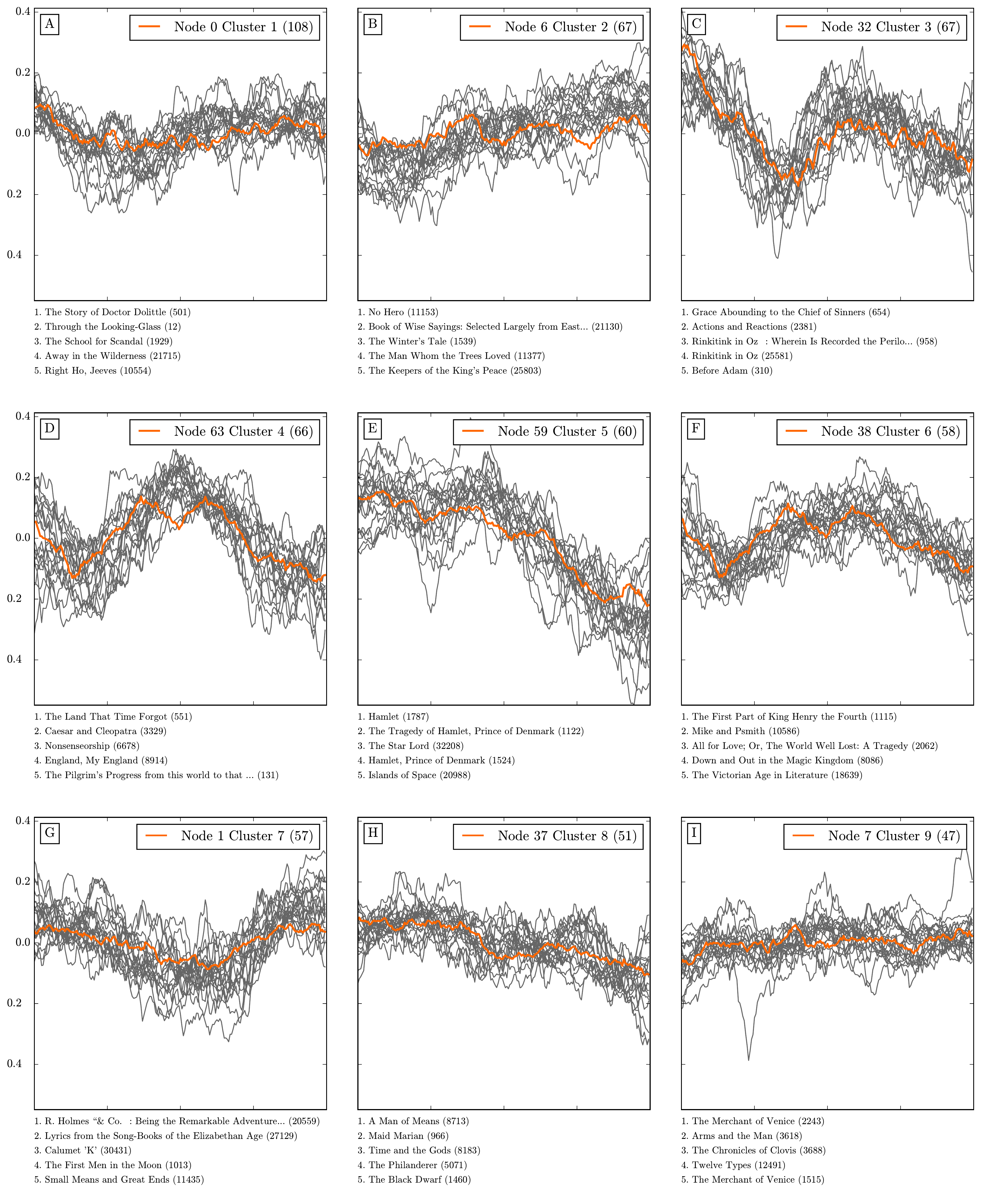}
  \caption[]{
    The vector for each of the top 9 SOM nodes, accompanied with those sentiment time series which are closest to that node.
    The core stories which we have found with other methods are readily visible.
  }
  \label{fig:SOM-stories}
\end{figure}

In Fig.~\ref{fig:SOM-stories} we show the emotional arcs that are closest to each of 9 most frequently winning nodes in the winner-take-all implementation the Self Organizing Map.

\clearpage
\pagebreak
\section{Null comparison details}
\label{sec:shuffled}

\begin{figure}[!htb]
  \centering
  \includegraphics[width=0.58\textwidth]{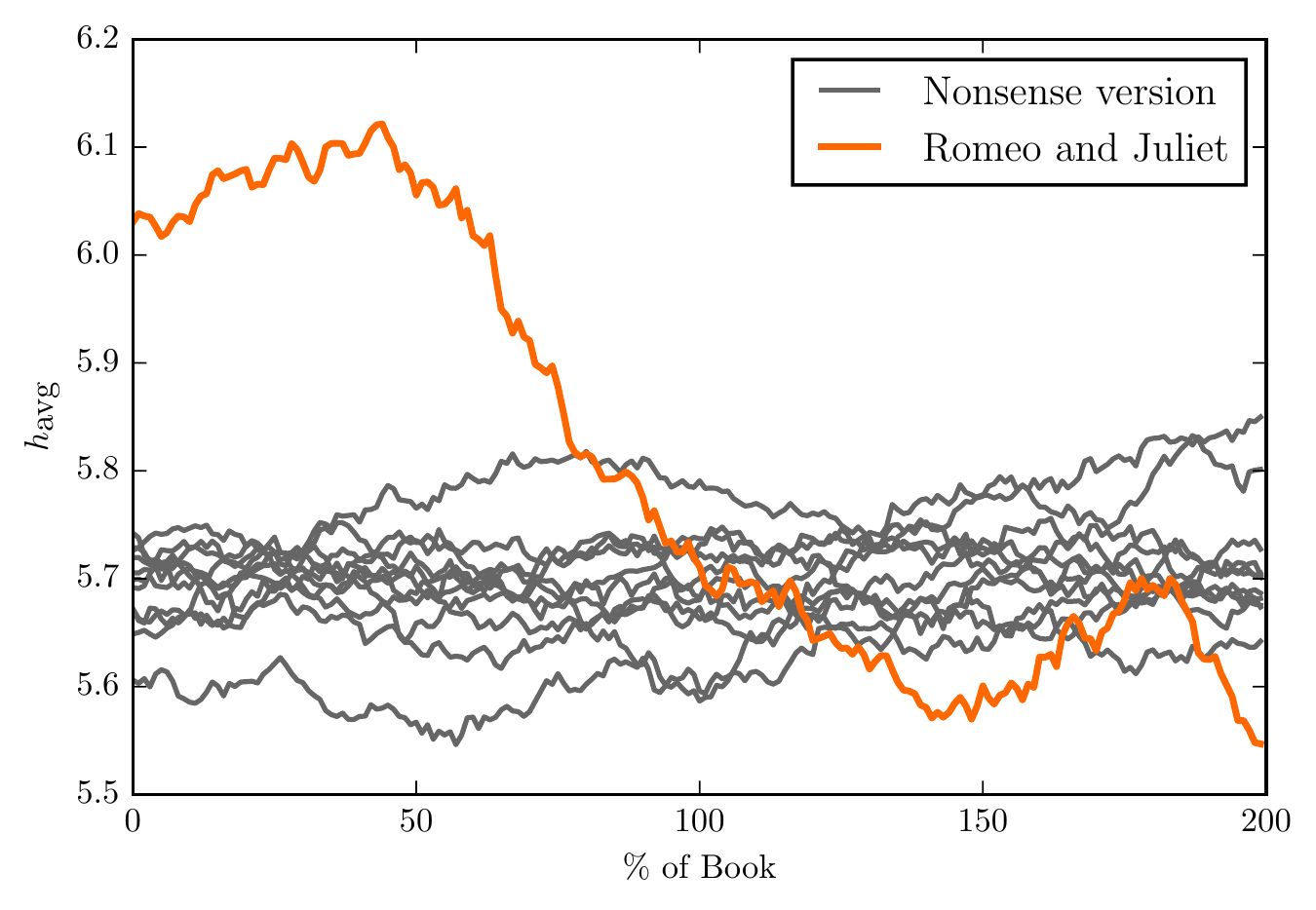}
  \caption{The emotional arc of \textit{Romeo And Juliet} by William Shakespeare (Gutenberg ID 1777), along with 11 ``nonsense'' versions, as produced by a 2-gram Markov model.
    We see that the emotional arc from the true version has more structure than the nonsense versions.}
  \label{fig:salad-1}
\end{figure}

\begin{figure}[!htb]
  \centering
  \includegraphics[width=0.58\textwidth]{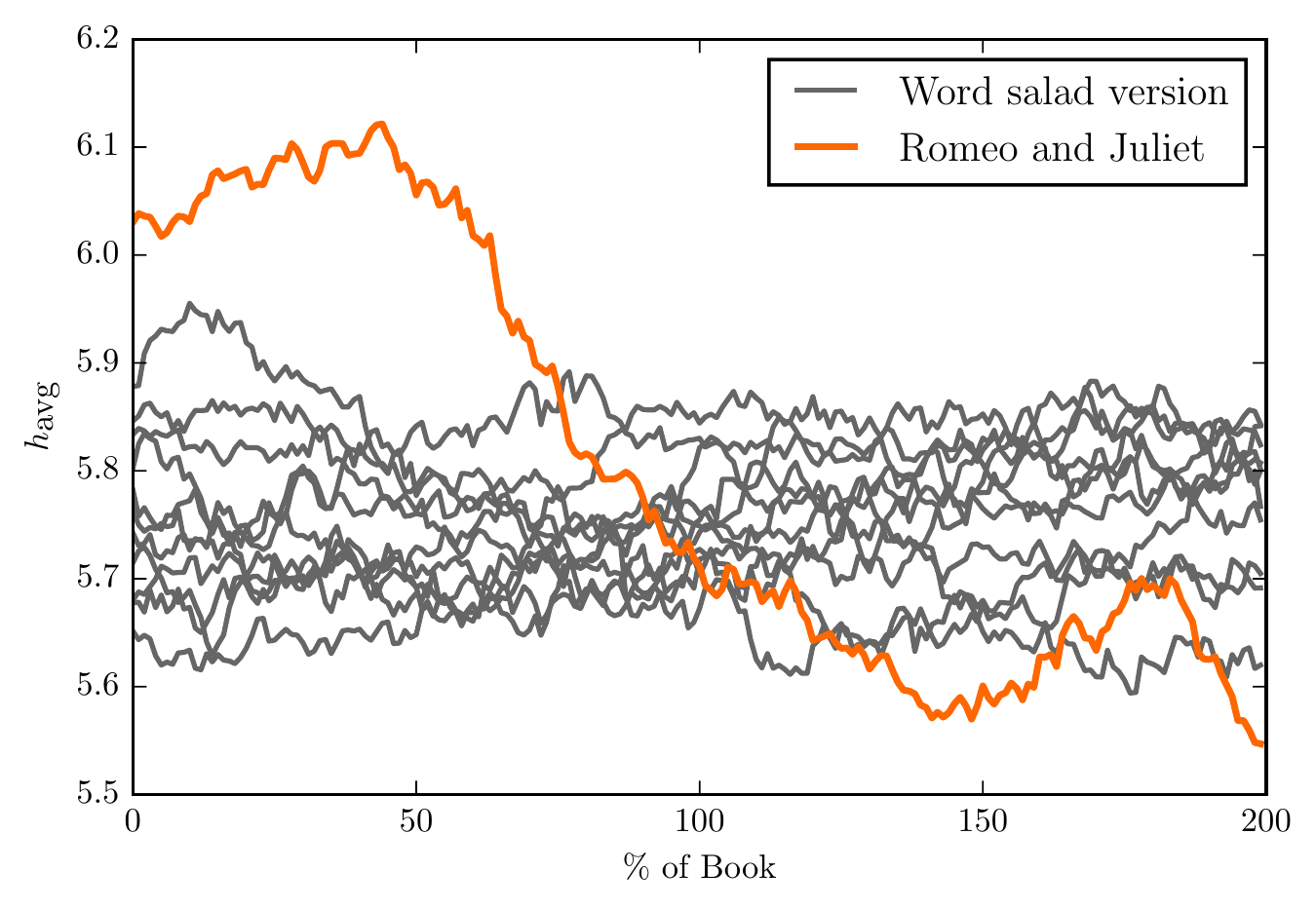}
  \caption{The emotional arc of \textit{Romeo And Juliet} by William Shakespeare, along with 11 ``word salad'' versions, as produced by randomly shuffling the words in the book.
    We see that the emotional arc from the true version has more structure than the word salad versions as well.}
  \label{fig:salad-2}
\end{figure}

An example of the ``nonsense'' and ``word salad'' text is presented first in Appendix \ref{sec:construction}.
First, we examine the resulting timeseries for an example book in Figs. \ref{fig:salad-1} and \ref{fig:salad-2}.
We then go on to present the full result of the SVD, agglomerative clustering, and SOM to ``nonsense'' English fiction books with more than 40 downloads.

\clearpage
\pagebreak
\subsection{Null SVD}

SVD modes from the emotional arcs of word salad books.
We observe higher frequency modes appearing more quickly, and a more even spread of mode coefficients.

\begin{figure*}[!htb]
  \centering
  \includegraphics[width=0.9\textwidth]{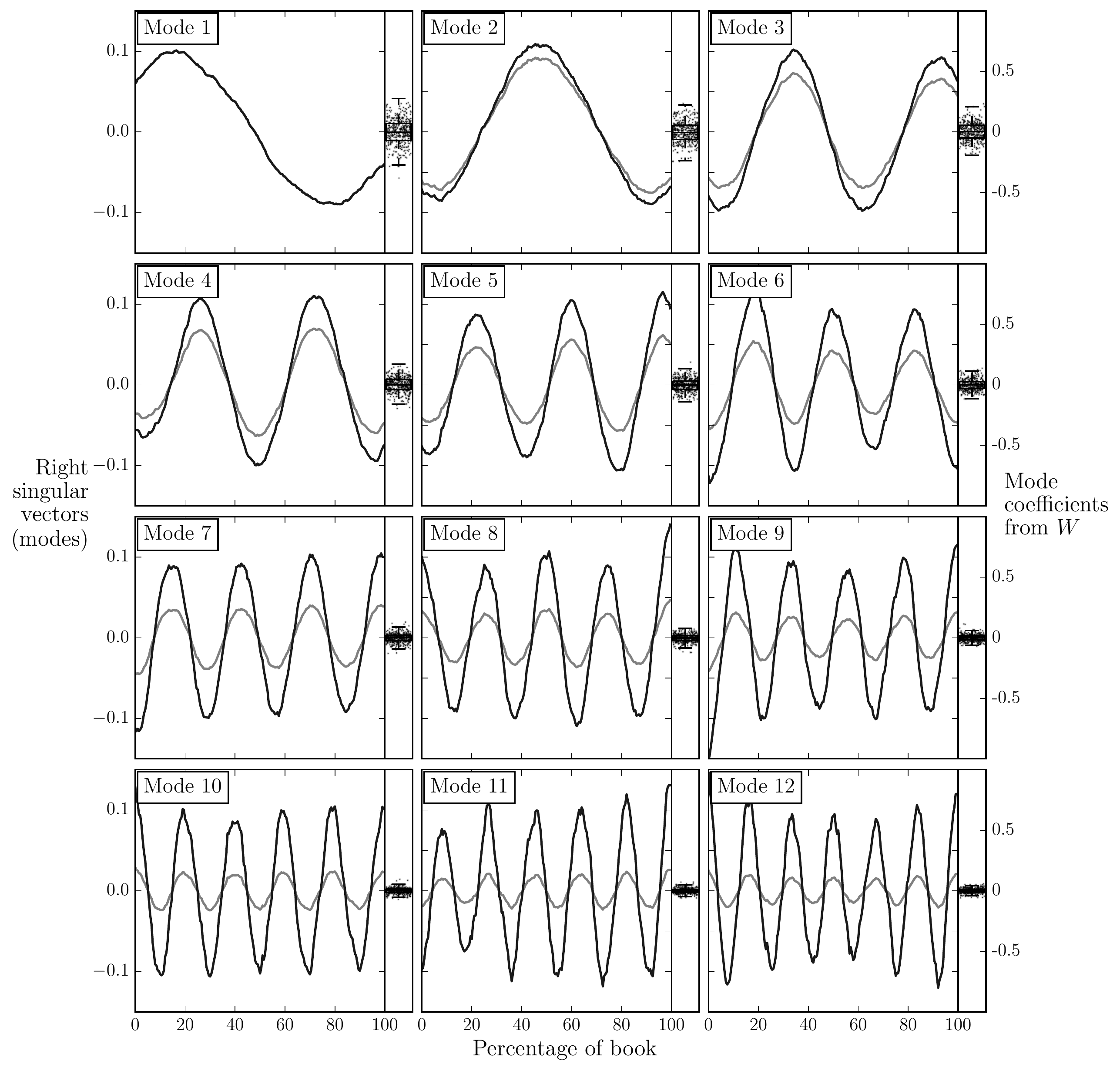}
  \caption[]{
    Top 12 modes from the Singular Value Decomposition of \nbooks~nonsense Project Gutenberg books.
    We show in a lighter color modes weighted by their corresponding singular value, where we have scaled the matrix $\Sigma$ such that the first entry is 1 for comparison.
    The mode coefficients normalized for each book are shown in the right panel accompanying each mode, in the range -1 to 1, with the ``Tukey'' box plot.
  }
  \label{fig:SVD-12-comp-salad}
\end{figure*}

\begin{figure*}[!htb]
  \centering
  \includegraphics[width=.98\textwidth]{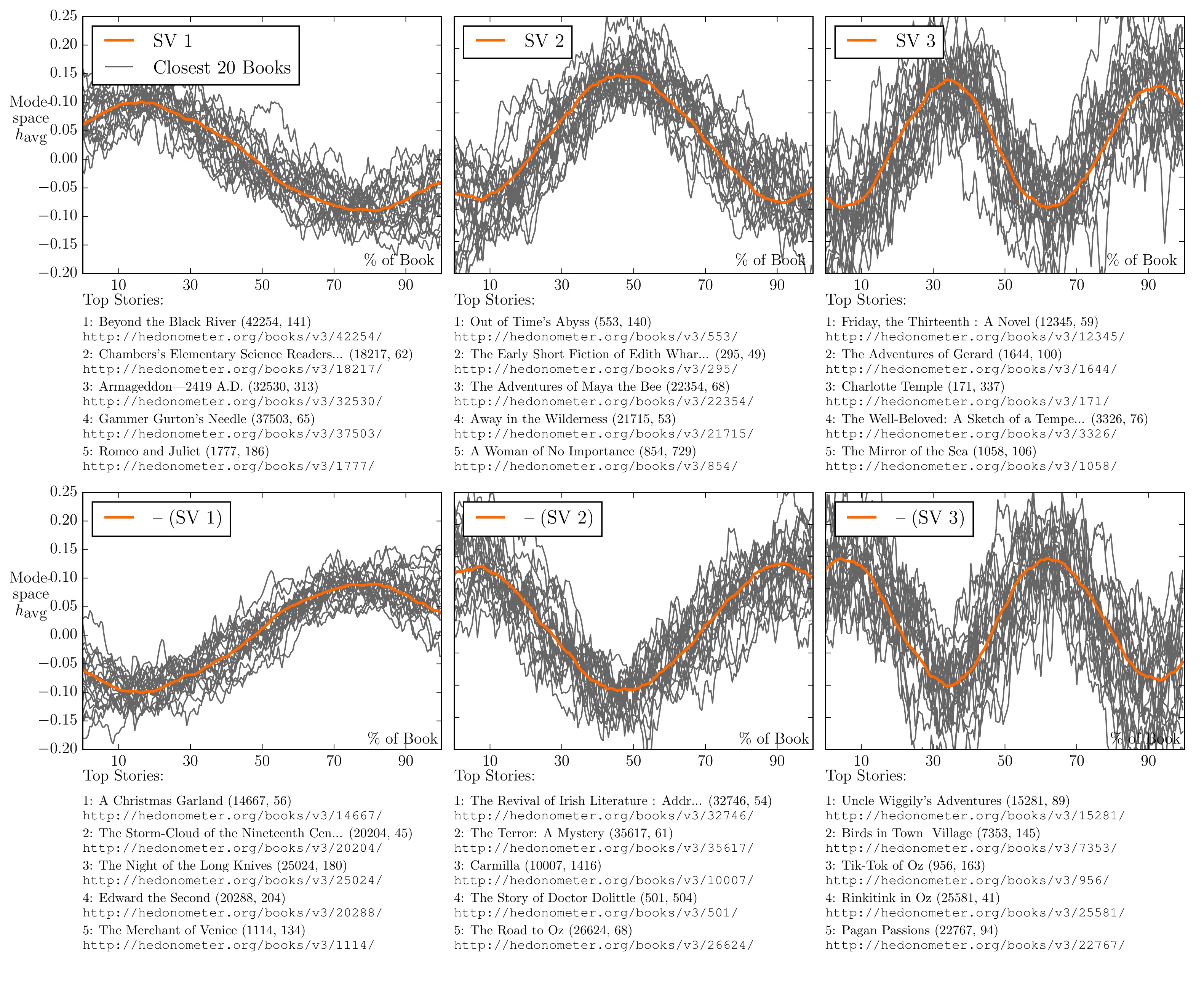}
  \caption[]{
    First 3 SVD modes from nonsense books and their negation with the closest stories to each.
    Links below each story point to an interactive visualization on \url{http://hedonometer.org} which enables detailed exploration of the emotional arc for the story.
  }
  \label{fig:SVD-1-3-labelled-salad}
\end{figure*}

\begin{figure*}[!htb]
  \centering
  \includegraphics[width=.98\textwidth]{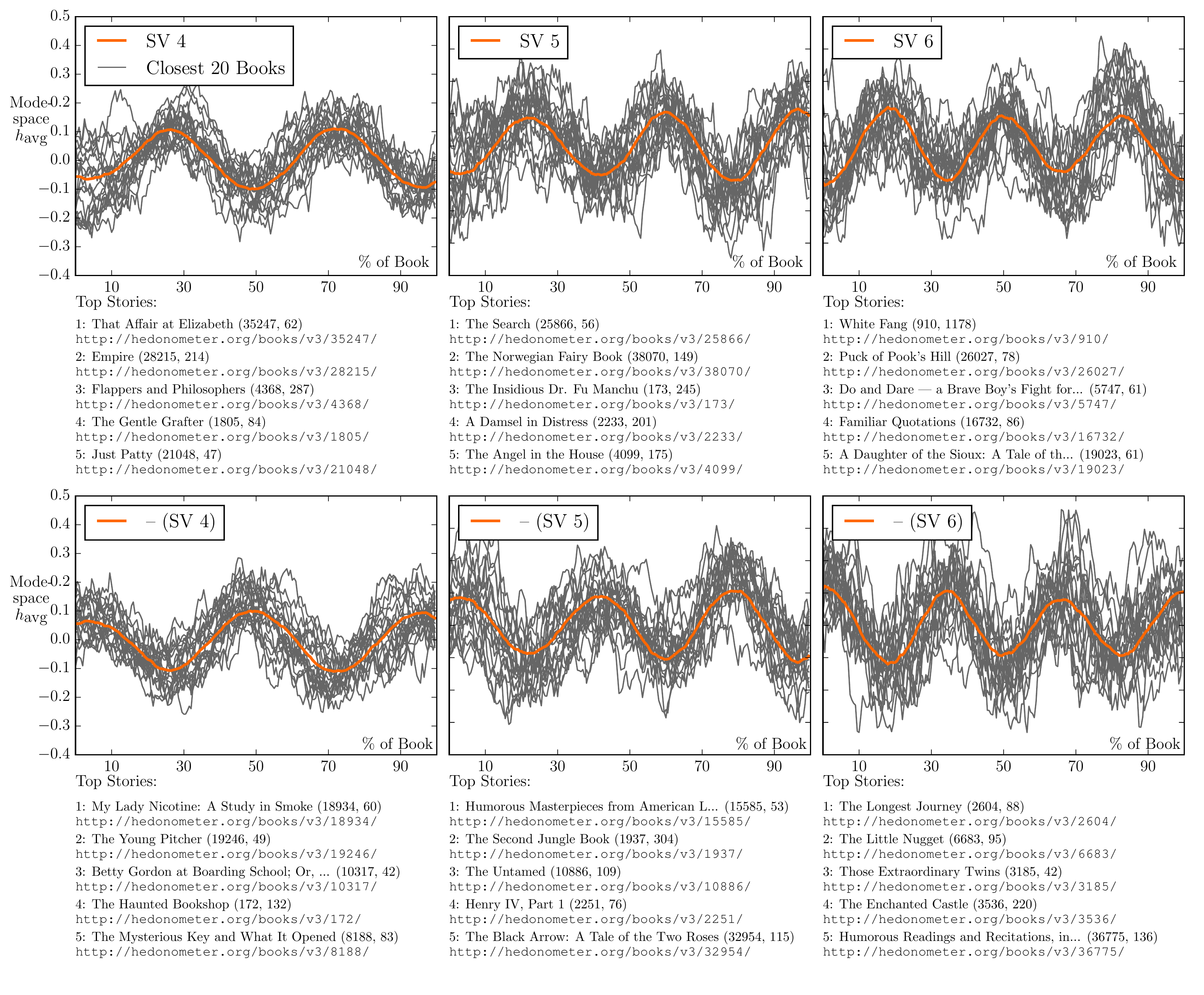}
  \caption[]{
    Modes 4--6 from the SVD analysis of nonsense books and their negation with the closest stories to each.
    Links below each story point to an interactive visualization on \url{http://hedonometer.org} which enables detailed exploration of the emotional arc for the story.
  }
  \label{fig:SVD-4-6-labelled-salad}
\end{figure*}

\begin{figure*}[!htb]
  \centering
  \includegraphics[width=.6\textwidth]{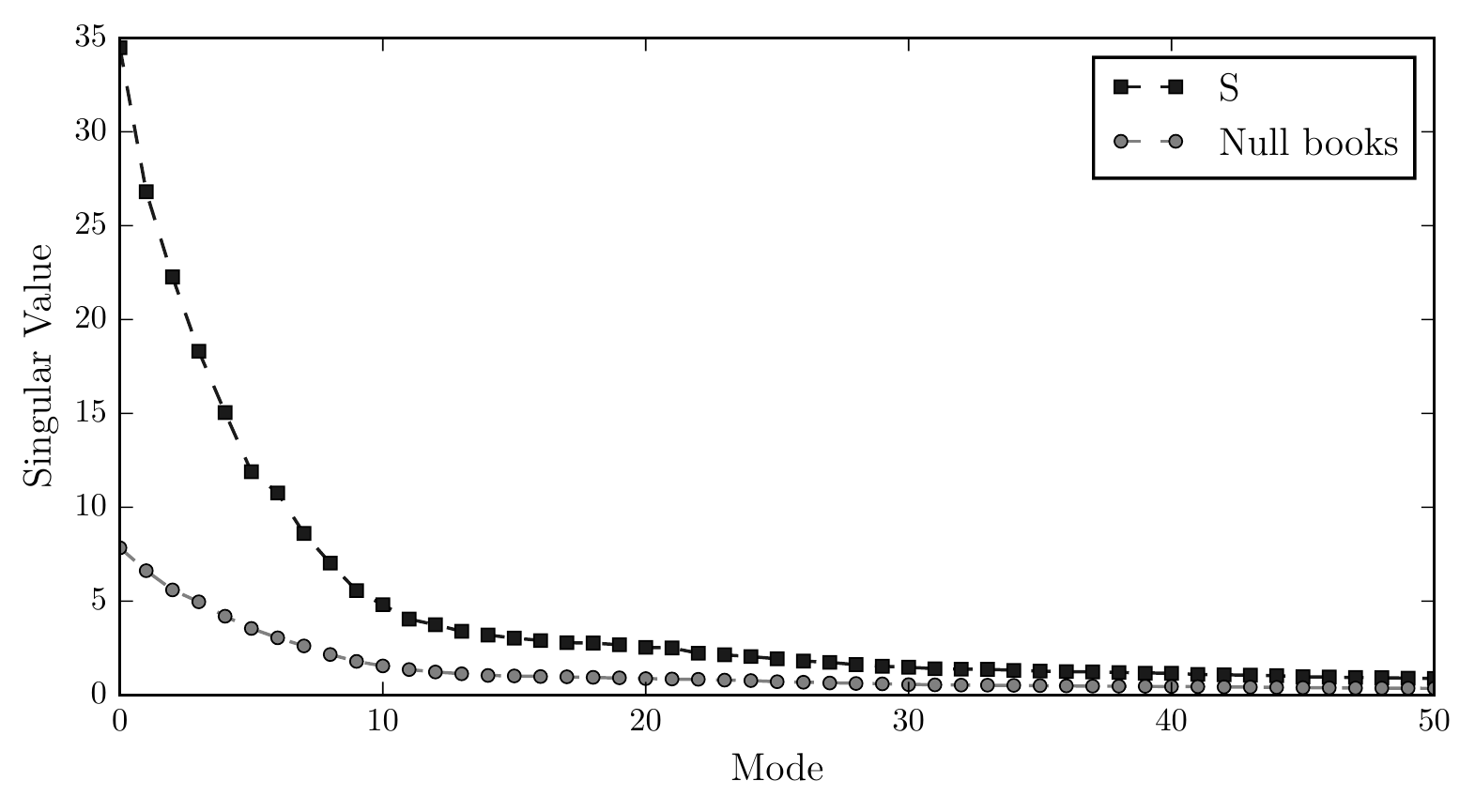}
  \caption[]{
    Comparison of the singular value spectra from the emotional arcs of nonsense books and the emotional arcs of individual Project Gutenberg books.
    The spectra from the nonsense books is muted, indicating both lower total variance explained and less important ordering of the singular vectors.}
  \label{fig:SVD-spectrum-comparison}
\end{figure*}

\clearpage
\pagebreak
\subsection{Null Hierarchical Clustering}

Dendrogram of clustering using Ward's method on the emotional arcs of word salad books.
We observe comparatively low linkage cost for these emotional arcs, indicating the absence of distinct clusters.

\begin{figure*}[!htb]
  \centering
  \includegraphics[width=0.98\textwidth]{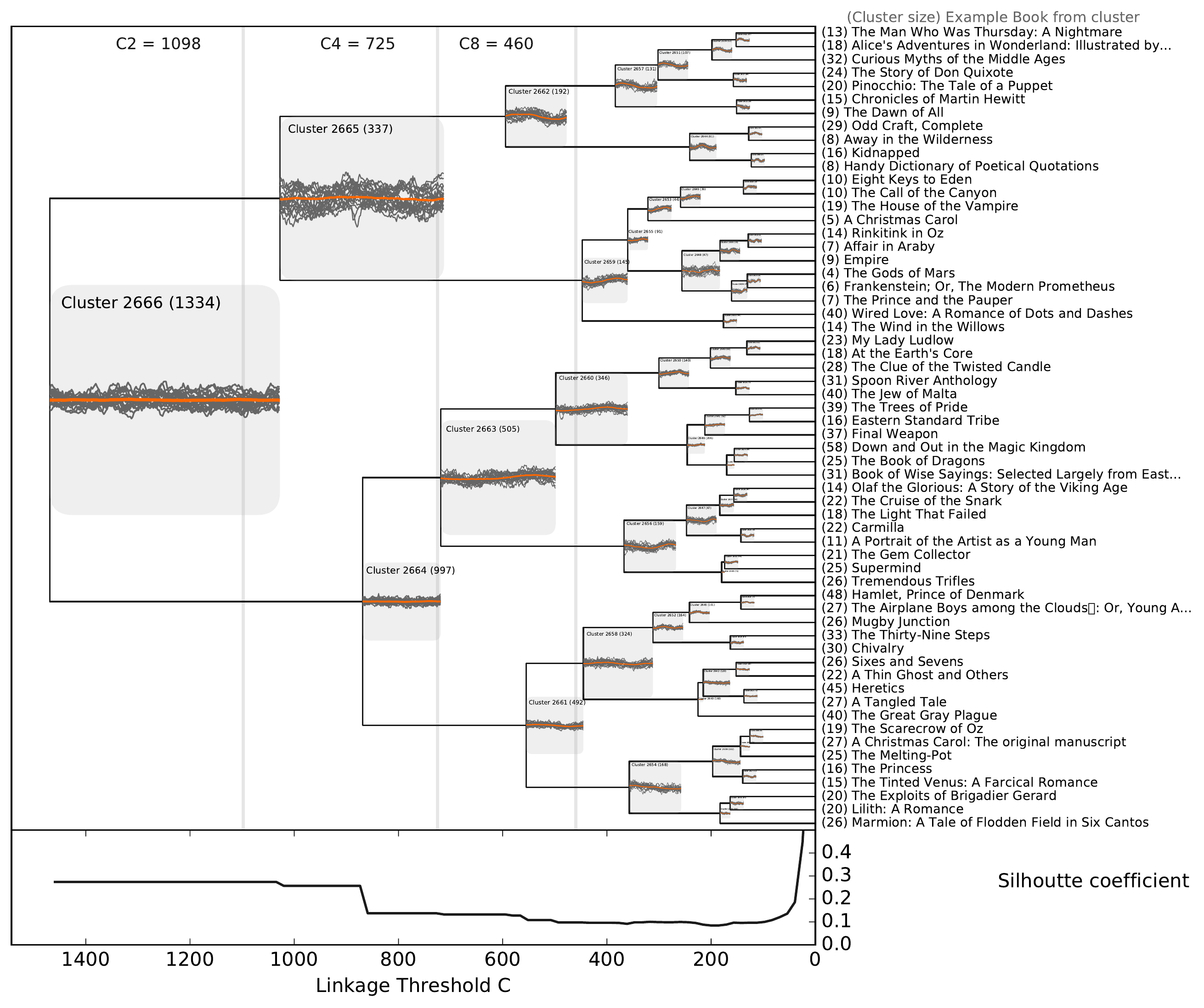}
  \caption[]{
    Dendrogram from the agglomerative clustering procedure using Ward's minimum variance method on nonsense books.
    For each cluster, a selection of the 20 most central books to a fully-connected network of books are shown along with the average of the emotional arc for all books in the cluster, along with the cluster ID and number of books in each cluster (shown in parenthesis).
    At the bottom, we show the average Silhouette value for all books, with higher value representing a more appropriate number of clusters.
    For each of the 60 leaf nodes (right side) we show the number of books within the cluster and the most central book to that cluster's book network.
  }
  \label{fig:ward-small-salad}
\end{figure*}

\begin{figure*}[!htb]
  \centering
  \includegraphics[width=0.98\textwidth]{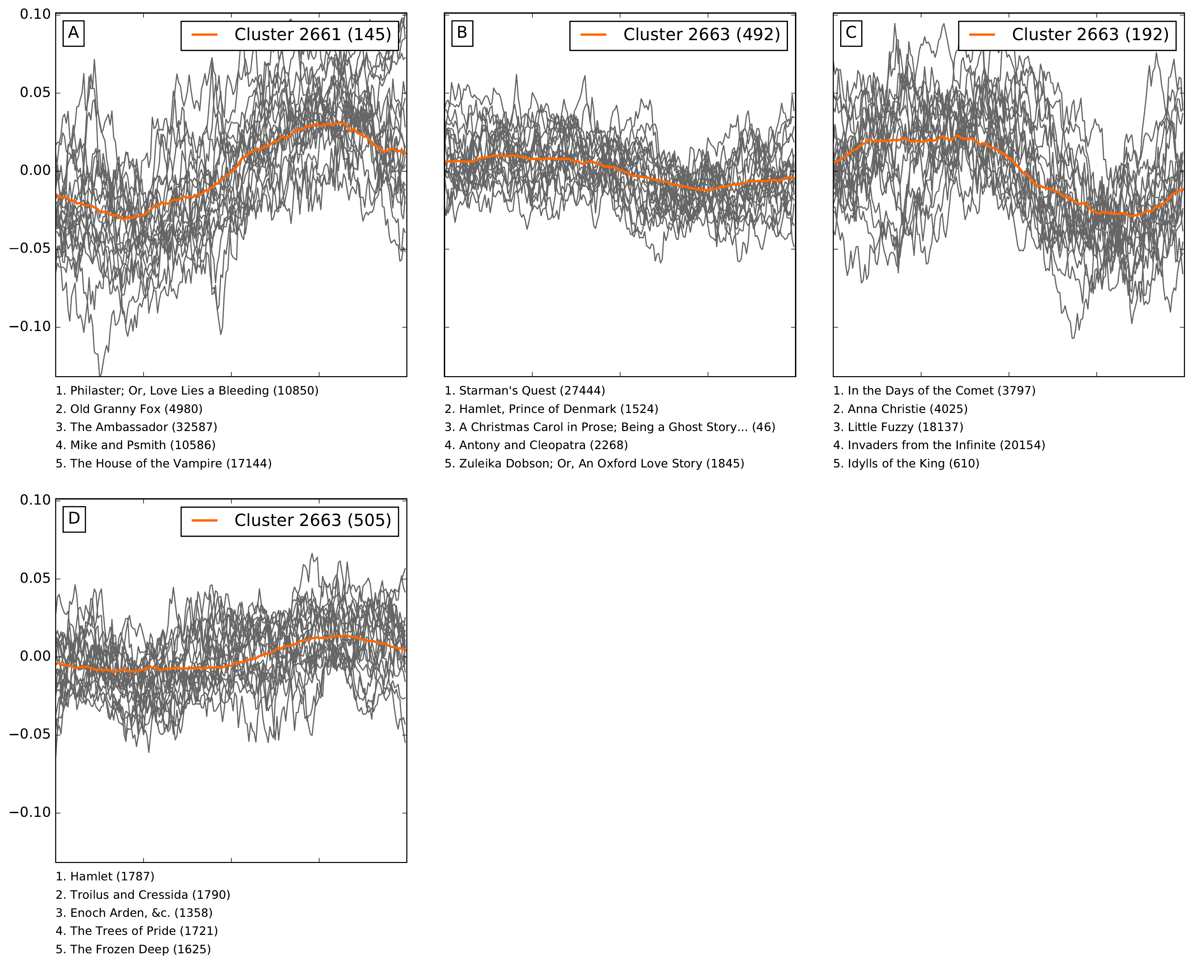}
  \caption[]{
    Four clusters (linkage threshold 850) from the hierarchical clustering of word salad books.
    We observe that the cluster mean emotional arc and the most central emotional arcs have high variance, without a visible signal.}
  \label{fig:ward-4-salad}
\end{figure*}

\clearpage
\pagebreak
\subsection{Null Self Organizing Map (SOM)}

\begin{figure*}[!htb]
  \centering
  \includegraphics[width=0.98\textwidth]{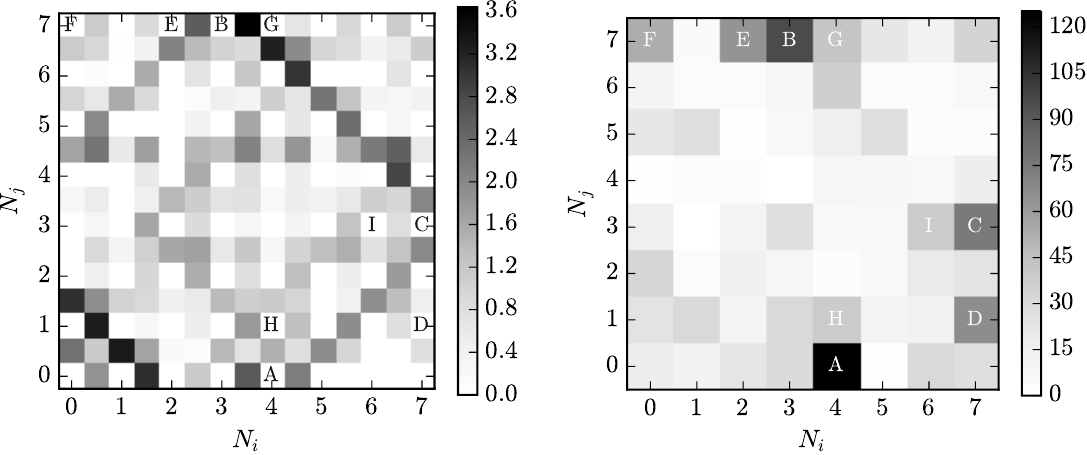}
  \caption[]{
    Results of the SOM applied to nonsense versions of Project Gutenberg books.
    Left panel: Nodes on the 2D SOM grid are shaded by the number of stories for which they are the winner.
    Right panel: The B-Matrix shows that there are clear clusters of stories in the 2D space imposed by the SOM network.
  }
  \label{fig:SOM-matrices-salad}
\end{figure*}

\begin{figure}[!htb]
  \centering
  \includegraphics[width=0.78\textwidth]{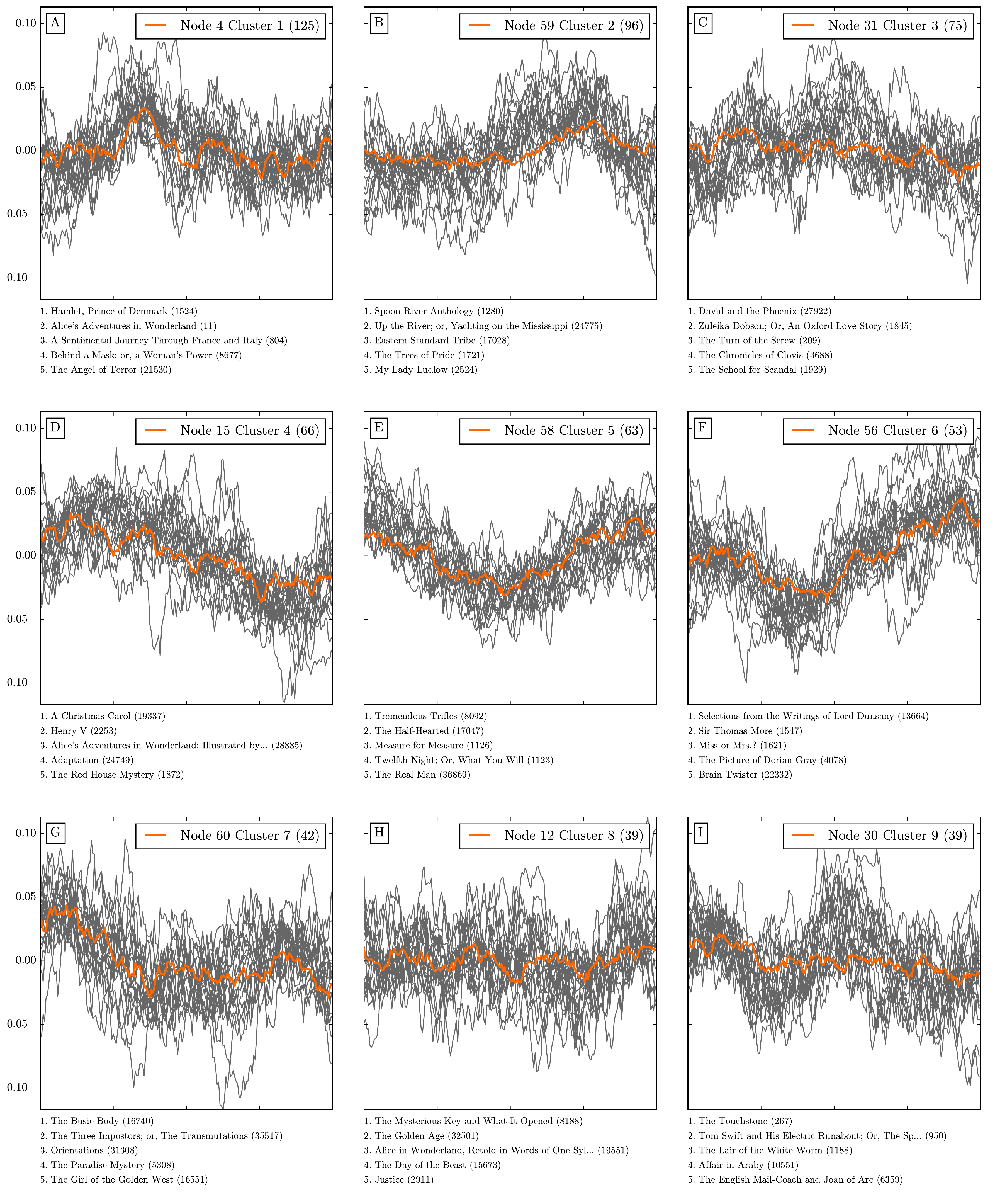}
  \caption[]{
    The vector for each of the top 9 SOM nodes for null emotional arcs, accompanied with those sentiment time series which are closest to that node.
    Panels D and E show what appear to be similar arcs to the six we identified in real books, but overall see that the emotional arcs from null arcs show little coherent structure, especially considering the y-range here being 0.1 compared to the 0.4 of the real books (had we used the same y-range, very little of the variation would be visible at all).
  }
  \label{fig:SOM-stories-salad}
\end{figure}

\end{document}